\documentclass[sigconf,screen]{acmart}

\AtBeginDocument{%
  \providecommand\BibTeX{{%
    \normalfont B\kern-0.5em{\scshape i\kern-0.25em b}\kern-0.8em\TeX}}}

\copyrightyear{2023}
\acmYear{2023}
\setcopyright{rightsretained}
\acmConference[AIES '23]{AAAI/ACM Conference on AI, Ethics, and
Society}{August 8--10, 2023}{Montréal, QC, Canada}
\acmBooktitle{AAAI/ACM Conference on AI, Ethics, and Society (AIES '23), August 8--10, 2023, Montréal, QC, Canada}
\acmDOI{10.1145/3600211.3604659}
\acmISBN{979-8-4007-0231-0/23/08}

\usepackage{subcaption}

\usepackage{xspace}

\makeatletter
\DeclareRobustCommand\onedot{\futurelet\@let@token\@onedot}
\def\@onedot{\ifx\@let@token.\else.\null\fi\xspace}

\def\eg{\emph{e.g}\onedot}

\makeatother

\usepackage{nicematrix}
\usepackage{booktabs}
\usepackage{multirow}
\usepackage{siunitx}
\usepackage{makecell}
\usepackage{enumitem}
\NiceMatrixOptions{notes = { para , style = \alph{#1} } }

\usepackage{booktabs, multirow, tabularx}
\newcolumntype{C}{>{\centering\arraybackslash}X}
\NewExpandableDocumentCommand\mcc{O{1}m}{\multicolumn{#1}{c}{#2}}
\usepackage{setspace}
\usepackage[nameinlink]{cleveref}
\usepackage{soul}

\begin{document}



\title{Flickr Africa: Examining Geo-Diversity in Large-Scale, Human-Centric Visual Data}


\author{Keziah Naggita}
\authornote{Authors contributed equally to this research.\\ This work was done when Keziah was an intern at Sony AI.}
\email{knaggita@ttic.edu}
\affiliation{%
  \institution{TTI-Chicago}
  \country{USA}
}

\author{Julienne LaChance}
 \authornotemark[1]
\email{julienne.lachance@sony.com}
\affiliation{%
  \institution{SONY AI America}
   \country{USA}
  }

\author{Alice Xiang}
\email{alice.xiang@sony.com}
\affiliation{%
  \institution{SONY AI America}
   \country{USA}
}

\renewcommand{\shortauthors}{Naggita, LaChance and Xiang.}

\begin{abstract}
Biases in large-scale image datasets are known to influence the performance of computer vision models as a function of geographic context. To investigate the limitations of standard Internet data collection methods in low- and middle-income countries, we analyze human-centric image geo-diversity on a massive scale using geotagged Flickr images associated with each nation in Africa. We report the quantity and content of available data with comparisons to population-matched nations in Europe as well as the distribution of data according to fine-grained intra-national wealth estimates. Temporal analyses are performed at two-year intervals to expose emerging data trends. Furthermore, we present findings for an ``othering'' phenomenon as evidenced by a substantial number of images from Africa being taken by non-local photographers. The results of our study suggest that further work is required to capture image data representative of African people and their environments and, ultimately, to improve the applicability of computer vision models in a global context. 
\end{abstract}



\maketitle

\section{Introduction}\label{sec:introduction}

\begin{figure*}[ht!]
\centering
\subfloat[Africa: Image data distribution colored by RWI.]{\includegraphics[width=0.44\textwidth]{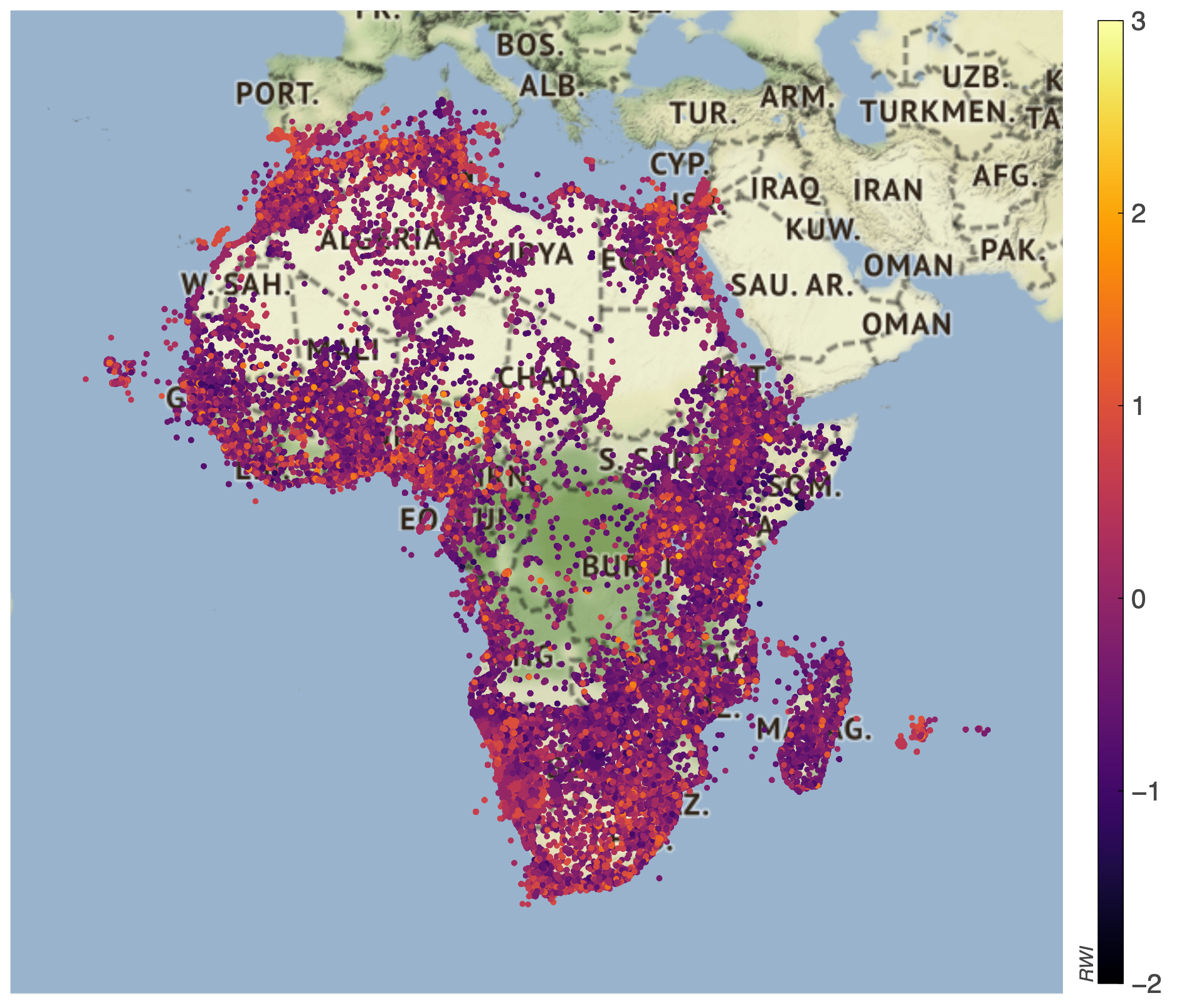}\label{fig:image_map}}
\subfloat[Africa: Total number of geotagged images.]{\includegraphics[width=0.42\textwidth]{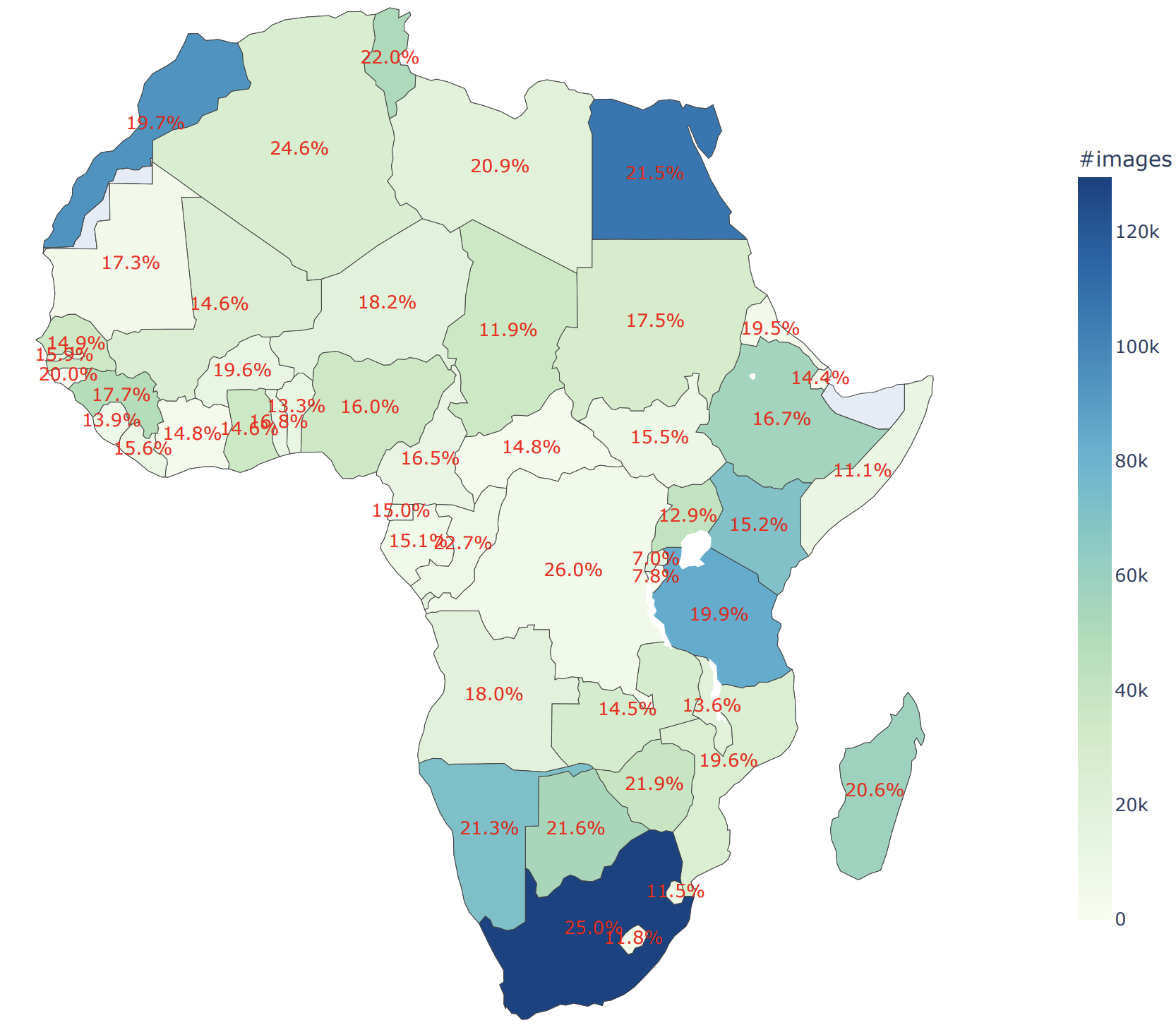} \label{fig:name_geo}}
\hfill
\subfloat[Madagascar: RWI group overlaid with image \\ count
per region.]{\includegraphics[width=0.44\textwidth]{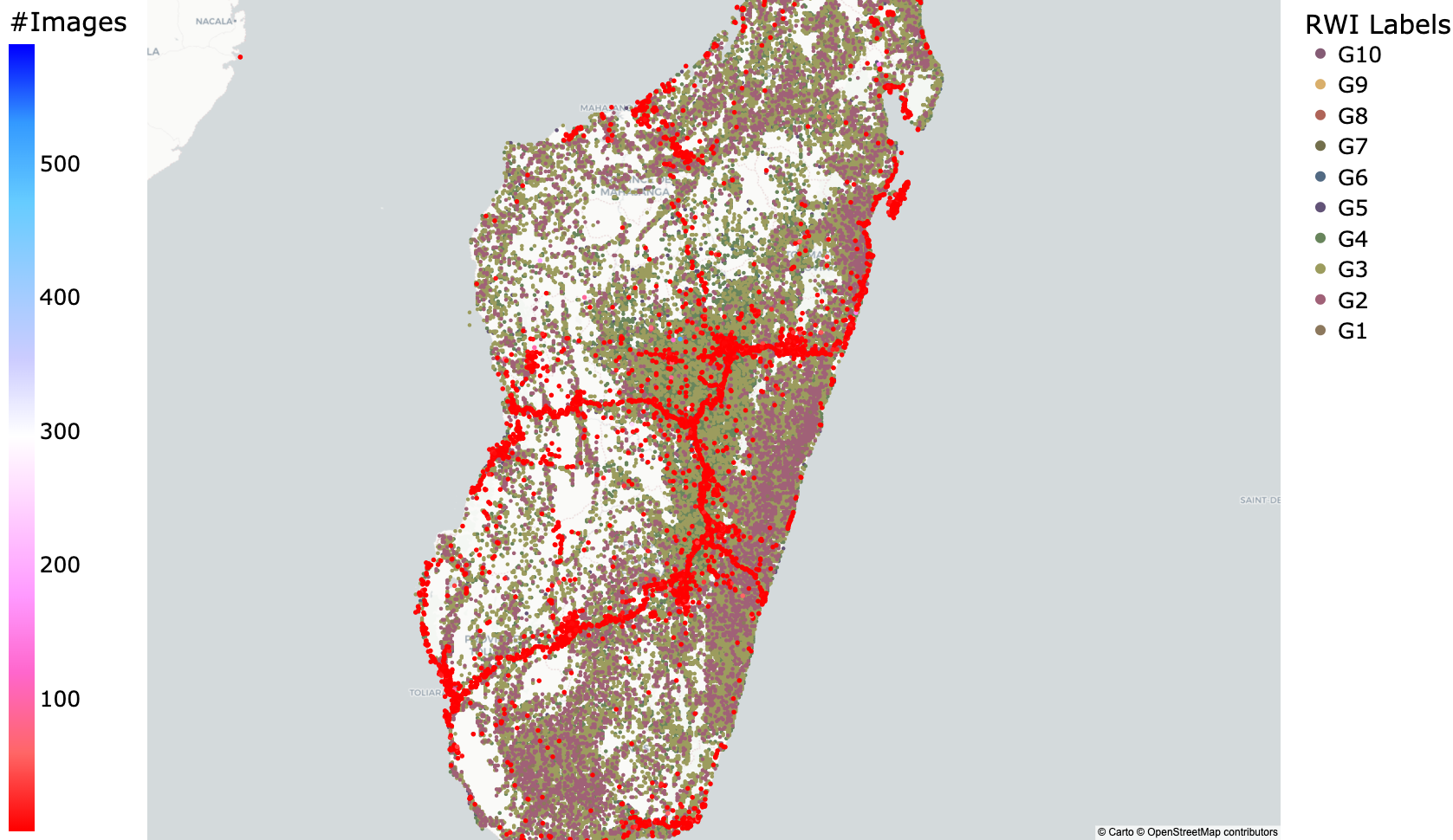} \label{fig:rwi_geo}}
\subfloat[Algeria: RWI group overlaid with image \\ count per region.]{\includegraphics[width=0.44\textwidth]{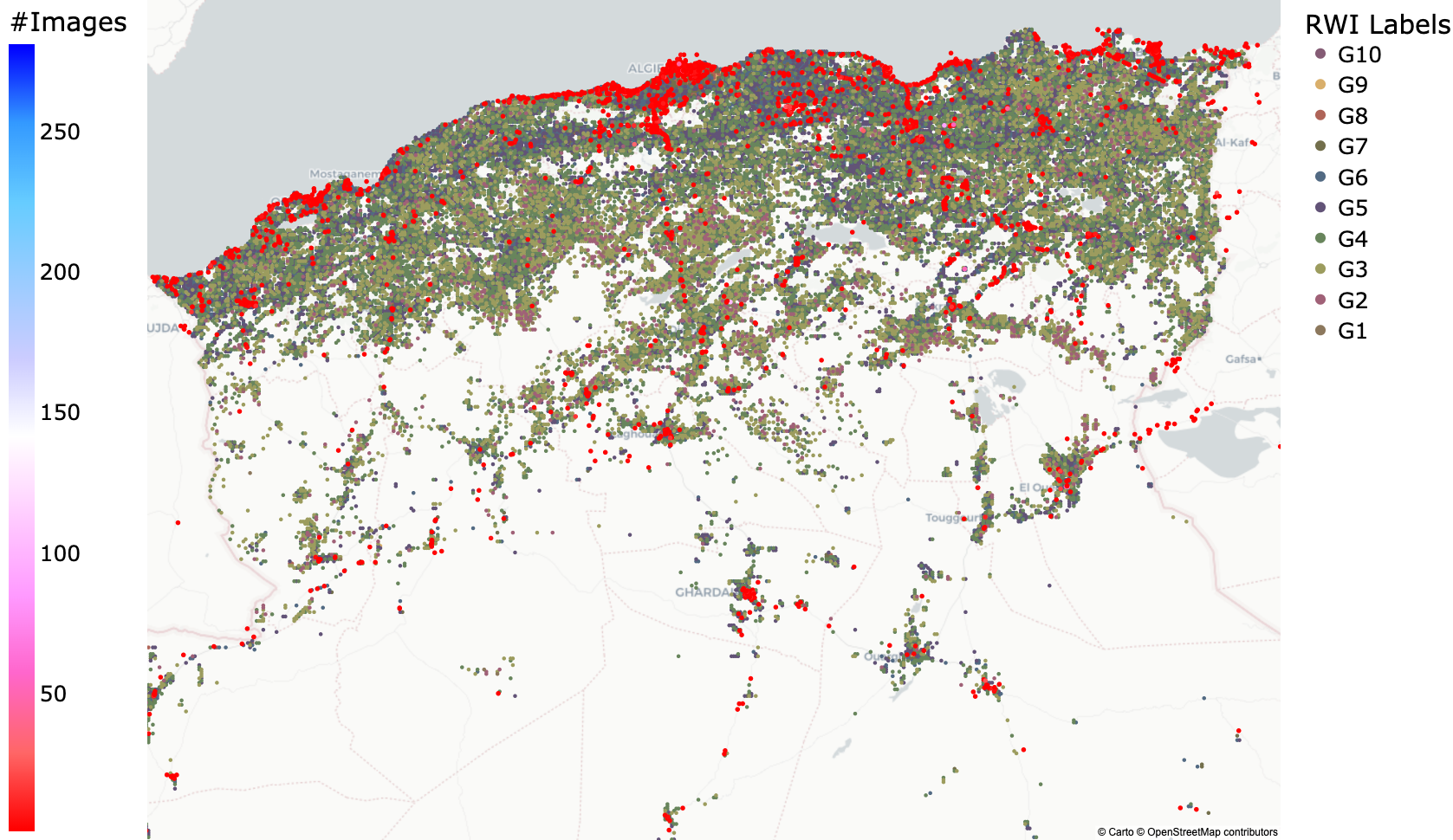} \label{fig:algeria_data__map}}
\caption{A collection of maps displaying relative wealth index (RWI) and geolocation of Flickr Africa images via country name query. Tolerance distances from geotag to nearest RWI-labeled point are: ((\textbf{a, b}) dist: $\leq 300km$; (\textbf{c, d}) $\leq 10km$). 
 (\textbf{b}) Nations are colored according to total number of geotagged images and the percentages (rounded to one decimal place) indicate the percentage of geotagged images out of the total image count per nation. South Africa had the highest number of geotagged images and Sao Tome and Principe had the smallest number of geotagged images while Cape Verde had the highest percentage of geotagged images and Rwanda had the lowest percentage of geotagged images.}
\label{fig:rwi-labeled data}
\end{figure*}

Data collection and processing are crucial to the machine learning (ML) pipeline and are the source of many biases in AI systems, which have been shown to largely stem from a lack of diverse representation in training datasets \cite{gender_shades}. Currently, most large-scale computer vision datasets are collected via webscraping and subsequent data cleaning. For example, the ImageNet database (\cite{imageNet}; $42607$ citations per Google Scholar, accessed Sept. 14, 2022) is comprised of images sourced from search engines like Google and Flickr, while the COCO dataset (\cite{coco-ds}; $26751$ citations per Google Scholar, accessed Sept. 14, 2022) is comprised of images sourced entirely from Flickr. Thus, biases inherent to Flickr influence the performance of models for visual tasks as diverse as object classification, pose estimation, instance segmentation, image captioning, and beyond. Some of these dataset biases have been explored in detail: for ImageNet and the Flickr-sourced Open Images dataset \cite{openimages} it has been shown that data from India, China, and African and South-East Asian countries is vastly underrepresented despite their large populations \cite{obj-notfor-everone}; while for COCO, data has been shown to be heavily skewed towards lighter-skinned and male individuals \cite{Zhao_2021_ICCV}. In particular, such biases impact the applicability of models in a global context. For instance, \citet{obj-notfor-everone} manually sourced image data from 264 globally-distributed households and demonstrated how object recognition model performance drops when applied in lower-income nations. Motivated by the popularity of datasets sourced using Flickr data, we here analyze $\mathbf{1.5}$ \textbf{million} geotagged images in the Flickr database to deeply explore its representation of African people and settings (see \Cref{fig:rwi-labeled data}).

\raggedbottom

In this paper, we aim to highlight the limitations of webscraping generic and human-centric\footnote{That is: involving people, their interactions with each other, and/or their activities in the environments in which they live.} image data from Africa for ML training purposes. We analyze image data for every African nation with direct comparisons to population-matched higher-GDP European nations and show that there is far less data available from Africa. We report the distribution of African geotagged image data as a function of fine-grained, intra-national wealth estimates \cite{RWI-estimates} and assess data with respect to license restrictions, population size, nominal GDP, Internet usage, and official languages. Additionally, we collect crowdsourced annotations to explore image content, and provide evidence for an ``othering'' phenomenon as the majority of African geotagged images we analyzed were taken by foreigners, while the opposite trend is shown for select European nations. Such results highlight the importance of considering geodiversity metrics beyond ancestry/ethnicity of individuals within images and, moreover, how the mechanisms by which images are obtained can quantitatively and qualitatively affect how the image corpus represents the world (\eg imposing a ``Western gaze''). Overall, we find that Flickr provides a very limited and skewed representation of African countries which likely contributes to many of the biases in models trained on popular, large-scale image datasets. 

\subsection{Related work}\label{sec:background_and_related_work}

While prior works have explored diversity beyond Western nations \cite{diversity-india, survey-of-fair-diverse}, studies of access to, and applicability of, AI systems in Africa remain limited \cite{data-sharing-africa, data-sharing-resources}. Scholars such as \citet{data-sharing-africa} highlight the challenges of data sharing practices in Africa, such as those concerning trust, awareness, and infrastructure, and note that ``The continent’s plural and at times divergent norms, practices, and traditions furthermore complicate the African data access and sharing ecosystem.'' Computer vision researchers have produced diverse datasets in an effort to reduce model biases and assess fairness outcomes (\eg \cite{karkkainenfairface}), with some centered specifically on geographic and contextual diversity \cite{dollar_street, obj-notfor-everone, GD-VCR, GeoImageNet, fMoW, crowsource-diverse}. Such data collection involves trade-offs, however \cite{seen_or_misseen}; while manual data collection enables desired contextual diversity specifications to be met, it is expensive and frequently limits access to low-income regions (see \eg \cite{obj-notfor-everone}). Thus, researchers have explored more automated methods of scraping diverse data from web platforms and public media, producing datasets such as the Geo-Diverse Visual Commonsense Reasoning dataset (GD-VCR) \cite{GD-VCR}, GeoImageNet \cite{GeoImageNet} , Functional Map of the World (fMoW) \cite{fMoW}, YFCC100M \cite{thomee2016yfcc100m}, and Open Images Extended \cite{kuznetsova2020open}, among others. \citet{revise} construct an ImageNet-style image data hierarchy across languages and cultures beyond English for visually grounded reasoning. While valuable, these initiatives have not deeply explored intra-national diversity, such as according to regional wealth estimates. Likewise, those datasets which utilize geolocation alone may result in a stereotypical portrayal of people in developing nations. 

Geodiversity has been studied from various angles beyond dataset production. Scholars have proposed methods for measuring geodiversity in image datasets \cite{overload-imagery, class-geodiverse} or performing geography-aware learning \cite{Ayush_2021_ICCV}. \citet{Zhao_2021_ICCV} expose the propogation of racial and cultural biases into model predictions, while \citet{ds-diversity-measure-mit} study geographical bias in image search and retrieval. \citet{crowdworksheets} highlight the importance of annotators' lived experiences on their annotation results. 

Additionally, \citet{mapping_world} and \citet{localness}, among other researchers, have studied volunteered-geographic information (VGI) and its relation to localness in Flickr user-generated content. 
At metropolitan-area and individual landmark spatial scales, \citet{mapping_world} use textual and visual image data to develop a classification technique which automatically exposes the relation between location and content in six months of Flickr-scraped images. 
\citet{localness} define four localness metrics: n-days, plurality, and location-field, to investigate the localness of user-generated content on Flickr, Twitter, and Swarm. In particular, their work assessed the Flickr-scraped YFCC100M dataset, containing images from thousands of users in the contiguous United States, whereas we focus on Africa and a few population-matched European countries. Notably, the authors found that with $31.1\%$ recall accuracy, only $40.7\%$ of Flickr images inspected with the ``location field'' localness metric (photographer self-reported location information) were local.

\section{Methodology}

\subsection{Data Collection}

\subsubsection{Flickr Africa} \textit{For each nation in Africa, we utilized Flickr queries to construct a dataset of images and associated metadata.} Using the FlickrAPI, we scraped images and associated metadata from Flickr between dates $2004$-$02$-$10$ and $2022$-$02$-$10$ ($18$ years) by querying by country name (\eg ``\emph{Togo}'') and the country name + people (\eg ``\emph{Togo people}''), with the latter querying choice motivated by construction methods of related large image datasets (\eg COCO, which utilizes the Flickr query ``person''). We scraped Flickr data for $54$ African countries: \{\emph{Algeria, Angola, Benin, Botswana, Burkina Faso, Burundi, Cameroon, Cape Verde, Central African Republic (CAF), Chad, Comoros, Ivory Coast, The Democratic Republic of the Congo (DRC), Djibouti, Egypt, Equatorial Guinea, Ethiopia, Eritrea, Gabon, Gambia, Ghana, Guinea, Guinea Bissau, Kenya, Lesotho, Liberia, Libya, Madagascar, Malawi, Mali, Mauritania, Mauritius, Morocco, Mozambique, Namibia, Niger, Nigeria, Republic of Congo, Rwanda, Sao Tome and Principe, Senegal, Seychelles, Sierra Leone, Somalia, South Africa, South Sudan, Sudan, Swaziland, Tanzania, Togo, Tunisia, Uganda, Zambia,} and \textit{Zimbabwe}\}. Utilizing Flickr metadata associated with each image, we generated $108$ \emph{csv} images data files ($2$ per country, associated with each query) with values for the following variables: \{\emph{``license'',	``title'', ``datetaken'', ``image\_url'',	``country'', ``city'',	``tags'', ``latitude'', ``longitude'', ``rwi of nearest point'', ``distance to nearest rwi labelled point (km)'', ``latitude of nearest point'', and ``longitude of nearest point''}\}. City and country information were determined by reverse geo-locating the longitude-latitude values provided in the image metadata using open-source reverse geocode (\cite{geocode-py}; accuracy analyses in \cite{kounadi2013accuracy}). 
All data is available on \href{https://doi.org/10.5281/zenodo.7133542}{zenodo.org}. The RWI data is described below. Total image counts were recorded and images without valid geotags were excluded.

\subsubsection{Population-matched European countries} \textit{The data collection process was repeated for four European nations.} In the interest of comparing data availability and content to higher-GDP European nations, we chose the following countries as a function of similar population size (\cite{nominal_gdp, european_countries_popn, african_countries_popn}): Switzerland and Sierra Leone (GDP: $841.97$k vs. $4.27$k); Cyprus and Djibouti (GDP: $27.73$k vs. $3.84$k); Finland and Central African Republic (CAF) (GDP: $297.62$k vs. $2.65$k); and Slovenia and Lesotho (GDP: $63.65$k vs. $2.56$k). For all $58$ countries we collected data pertaining to percentage of internet users \cite{internet_users}, nominal GDP \cite{nominal_gdp}, population size \cite{european_countries_popn,african_countries_popn} and official languages \cite{official_languages}.

\subsubsection{Relative Wealth Estimates} \textit{Fine-grained relative wealth estimates were associated with each geotagged image.} To assess the image distribution according to local wealth estimates, we utilize the relative wealth index (RWI) data collected from Low and Middle-Income Countries (LMICs) by Facebook's Data for Good project \cite{RWI-estimates}. RWI scores are normalized by nation, so the data should only be utilized for intra-national wealth analyses. The RWI dataset contains relative wealth distribution for $49$ African countries, such that the following countries are excluded from our original list of nations: \{\emph{Somalia, Seychelles, Sao Tome and Principe, Sudan,} and \emph{South Sudan}\}.
Therefore, when analyzing the relationship of RWI to geotagged images, these four countries are excluded. 
RWI data is provided in the form of 3-lettered iso-codes and the following variables are provided:  \emph{``quadkey'', ``latitude'', ``longitude'', ``rwi'', and ``error''}; Nominatim  API \cite{nominatim} was utilized to assign and add variables \emph{``country'', ``city''} to the data files. Using k-nearest neighbour, we computed the nearest RWI-labeled geographic location of each image.
\Cref{fig:image_map} shows the distribution of RWI-labeled geotagged images with a ~$300km$ maximum tolerance limit between the image geotag and the nearest RWI-labeled location.

\subsubsection{Manual Content Annotation} \textit{Crowdsourced annotations were collected for six additional image features.} 

\begin{figure*}[!thbp]
\centering
\includegraphics[width=0.86\textwidth]{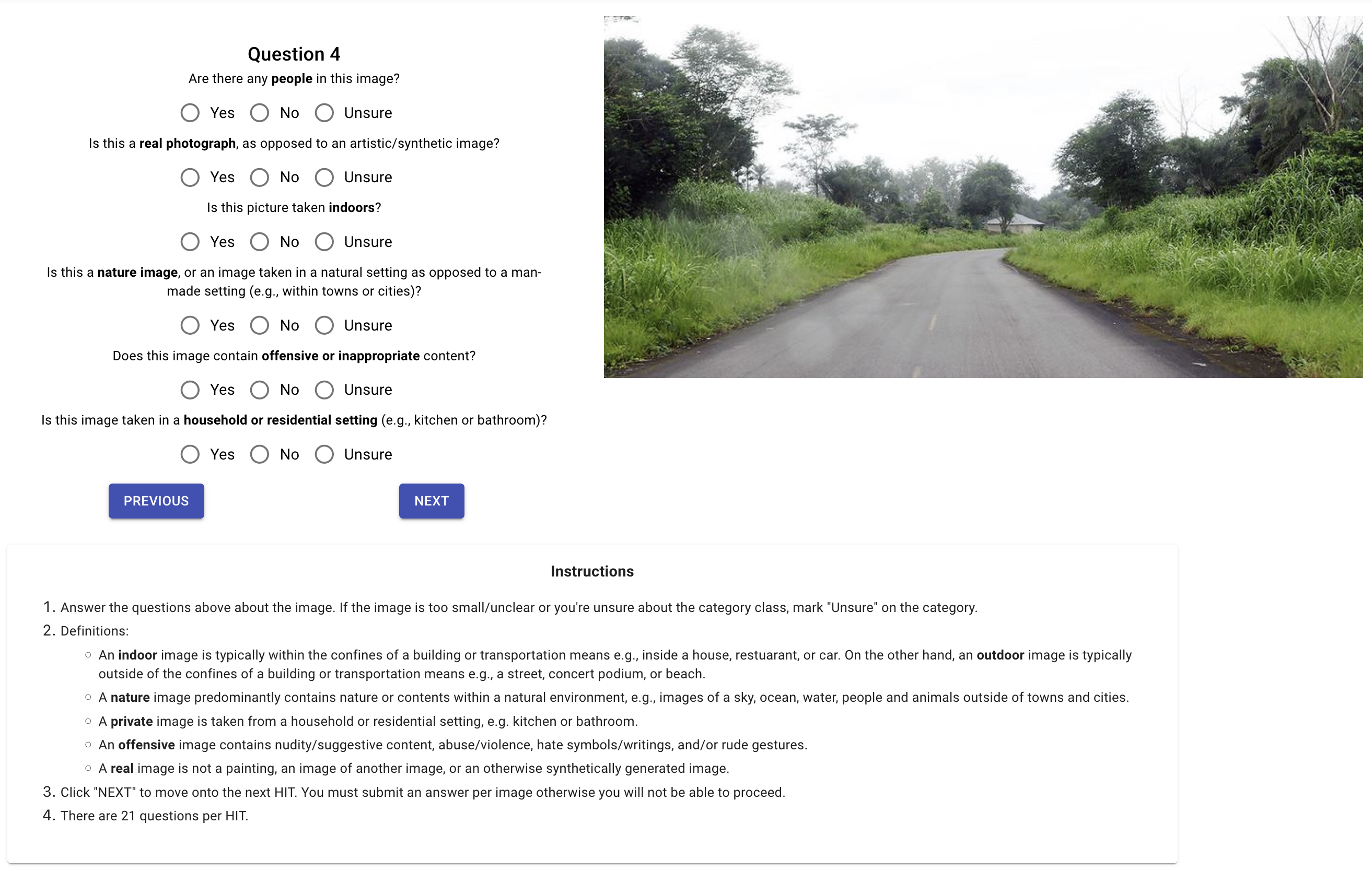}
\caption{Sample AMT task page which annotators utilized to label binary attributes pertaining to image content. Each HIT involved interaction with an introductory instructional page followed by 21 task pages similar to the one shown above.}
\label{fig:amt_task_pages}
\end{figure*}

We used Amazon Mechanical Turk (AMT) to collect annotations describing image contents.
Each Human Intelligence Task (HIT) involved $21$ images, with six binary questions per  image as shown in \Cref{fig:amt_task_pages}. The binary questions required the annotator label the image according to: indoor vs. outdoor setting, public vs. private setting, nature vs. manmade setting, the presence of people, real vs. synthetic image type, and offensive vs. inoffensive content. Below were our definitions of the terms or labels;
\begin{itemize}
    \item An indoor image is typically within the confines of a building or transportation means, e.g., inside a house, restaurant, or car. 
    \item A  private image is taken from a household or residential setting, e.g., kitchen or bathroom.
    \item A nature image predominantly contains nature or contents within a natural environment, e.g., images of a sky, ocean, water, people and animals outside of towns and cities. 
    \item A  real image is not a painting, an image of another image, or an otherwise synthetically generated image. 
    \item An offensive image contains abuse or violence, nudity or suggestive content, hate symbols or writings, and or rude gestures. 
\end{itemize}

\raggedbottom

We compensated workers at a rate of $\$15$ USD/hour. We sourced each annotation from three different annotators and chose the majority consensus value, excluding those images marked ``Unsure''. To ensure high annotation quality, we recruited workers with at least $95\%$ acceptance rating and completion of $1000$+ prior tasks. We randomly inserted a gold standard image within each set of $20$ standard images to assess annotator performance; if the worker failed on this test image, we discarded the annotations but still paid the worker for their contribution.

\subsection{Limitations of Our Approach}

We acknowledge four notable limitations of our method. First, we recognize that geolocation data (longitude, latitude) is inherently unreliable. Values may be modified or removed by the Flickr user or otherwise not reflect the location of capture, while reverse geolocation methods are computationally expensive and often fail, particularly with geographic locations close to region borders. This motivates our use of both geotags and country name tags for cross-validation of location, though this restricts us to fewer data samples overall. 
Secondly, to determine location of photographers to assess localness, we relied on photographers volunteered information of their location from their profile metadata. This doesn't take into account confounding factors like an immigrant visiting their home nation. 
Additionally, some forms of geodiversity are difficult or impossible to determine from visual inspection alone, such as an individual's gender, ethnicity, or religion. Finally, we were limited to obtaining data using only two queries, namely, by country name or country name + ``people''. We anticipate future work exploring a wider variety of query terms, both in English and local languages; here, no correlation was determined between dominant national languages and geotagged image availability.

\subsection{Ethical Considerations}
We note that although the Flickr images analyzed here are all publicly viewable, we show that most have the Flickr default license of ``All Rights Reserved''. Thus, we have opted to provide image URLs in lieu of images for direct download to avoid duplication of protected content, particularly in the event that a Flickr user chooses to remove or modify the permissions of an image. We acknowledge the weaknesses of this method in terms of consent, as public Flickr images are typically not taken by those in the images (as pointed out by \citet{Birhane_2021_WACV}); likewise, Flickr users may wish to avoid the utilization of their images for research purposes. Given that our objective is to critique large-scale image dataset curation strategies which do not respect image licenses (\eg the methodology for generating the COCO dataset), we deemed it justifiable to perform basic analyses on protected images and to build awareness regarding widespread license violations in standard AI training pipelines.

\begin{figure*}[!thbp]
\centering
\subfloat[]{\includegraphics[width=0.5\textwidth]{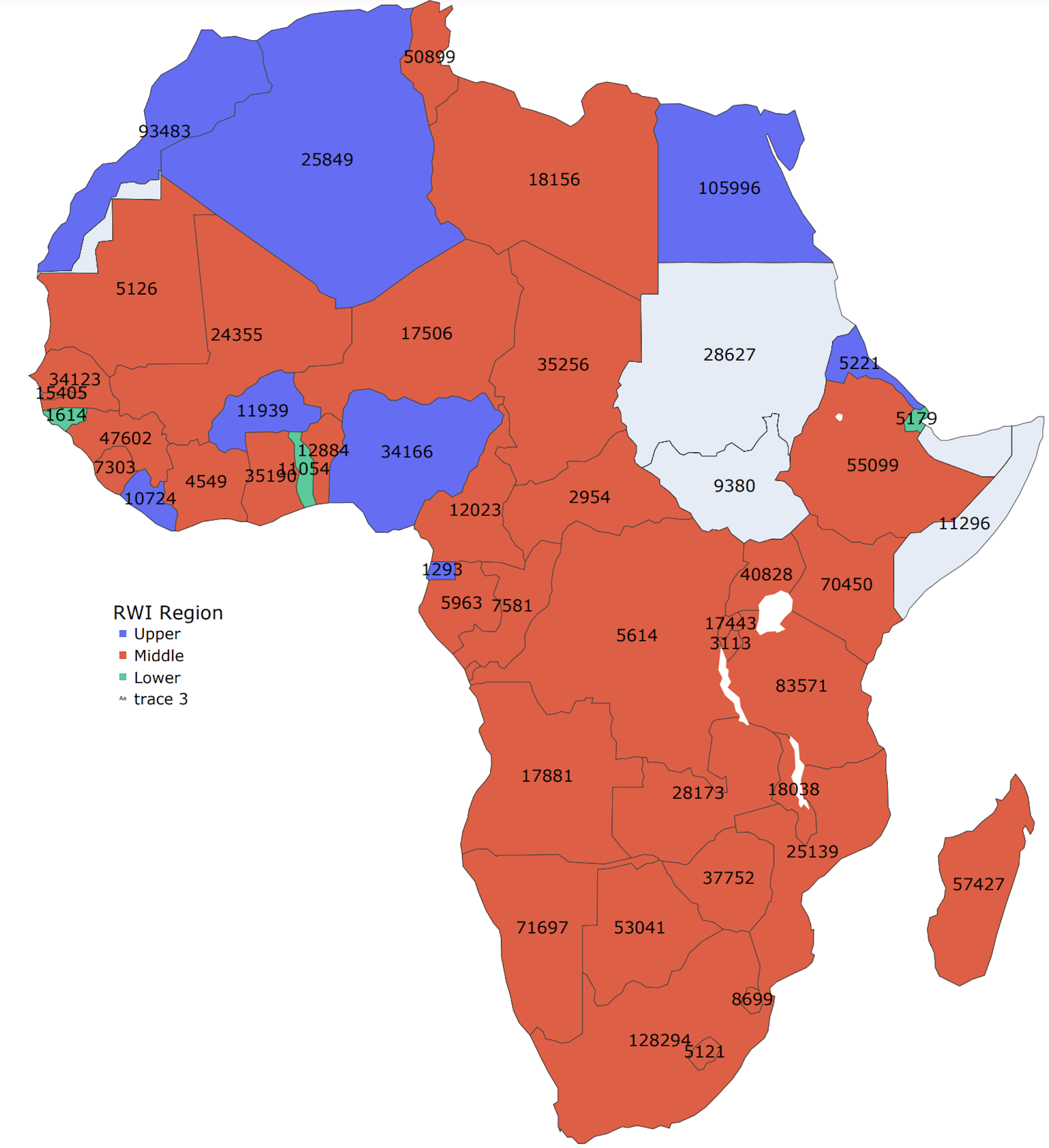} \label{fig:rwi_geo_name}}
\subfloat[]{\includegraphics[width=0.496\textwidth]{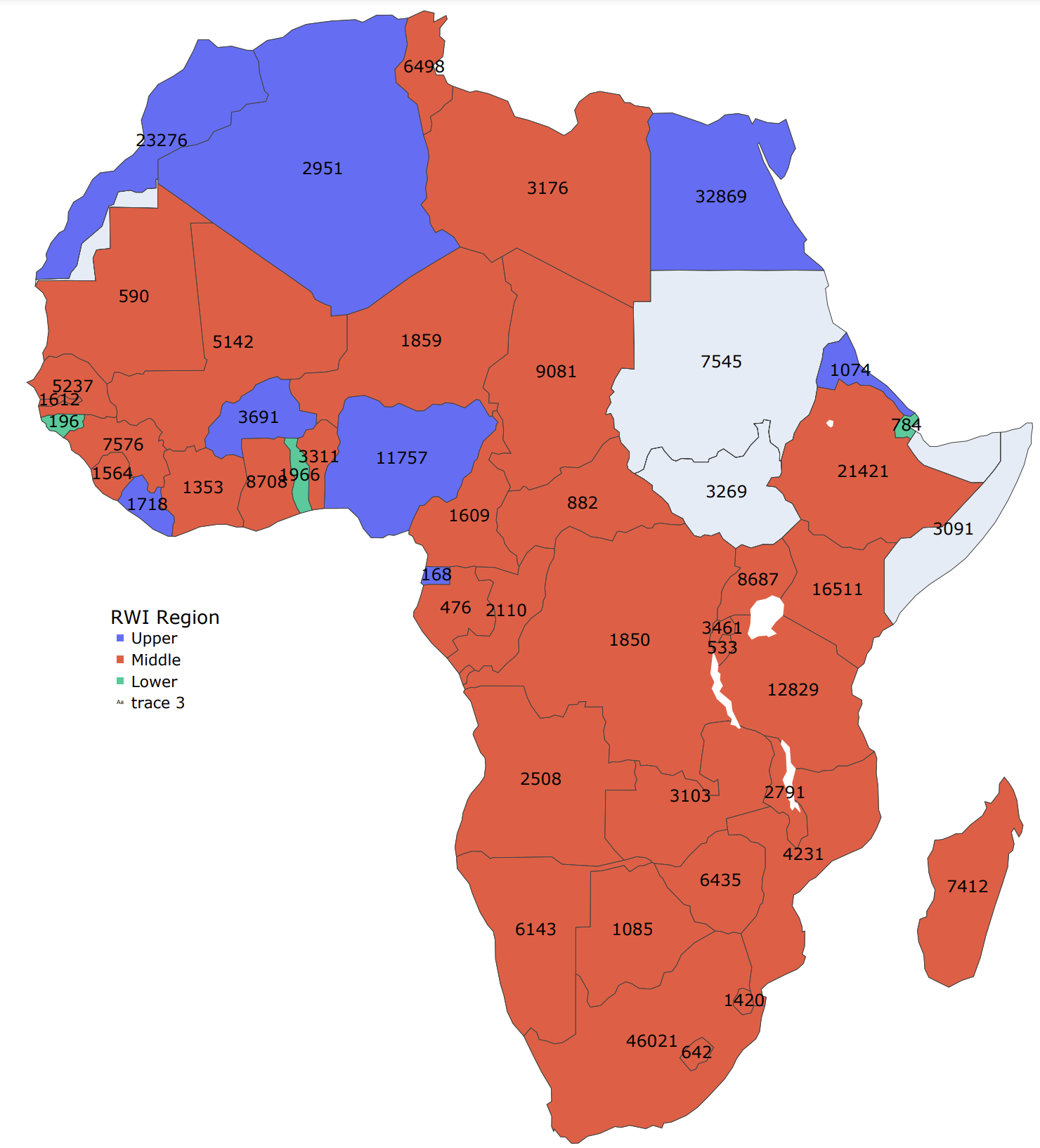} \label{fig:rwi_geo_people}}
\caption{Geotagged image counts by nation and respective RWI regions for (\textbf{a}) ``query-by-country name'' and (\textbf{b}) ``query-by-country name+people''. Nations are colored according to dominant RWI group (upper, middle, or lower wealth group) from which most images were sourced. Images mainly came from middle RWI groups ($G4$, $G5, G6$ and $G7$). The numbers denote the number of geotagged images. Countries that didn't have RWI data are in grey.}
\label{fig:name_people_map}
\end{figure*}
\raggedbottom

\begin{figure*}[!thbp]
\centering
\subfloat[]{\includegraphics[width=0.48\textwidth]{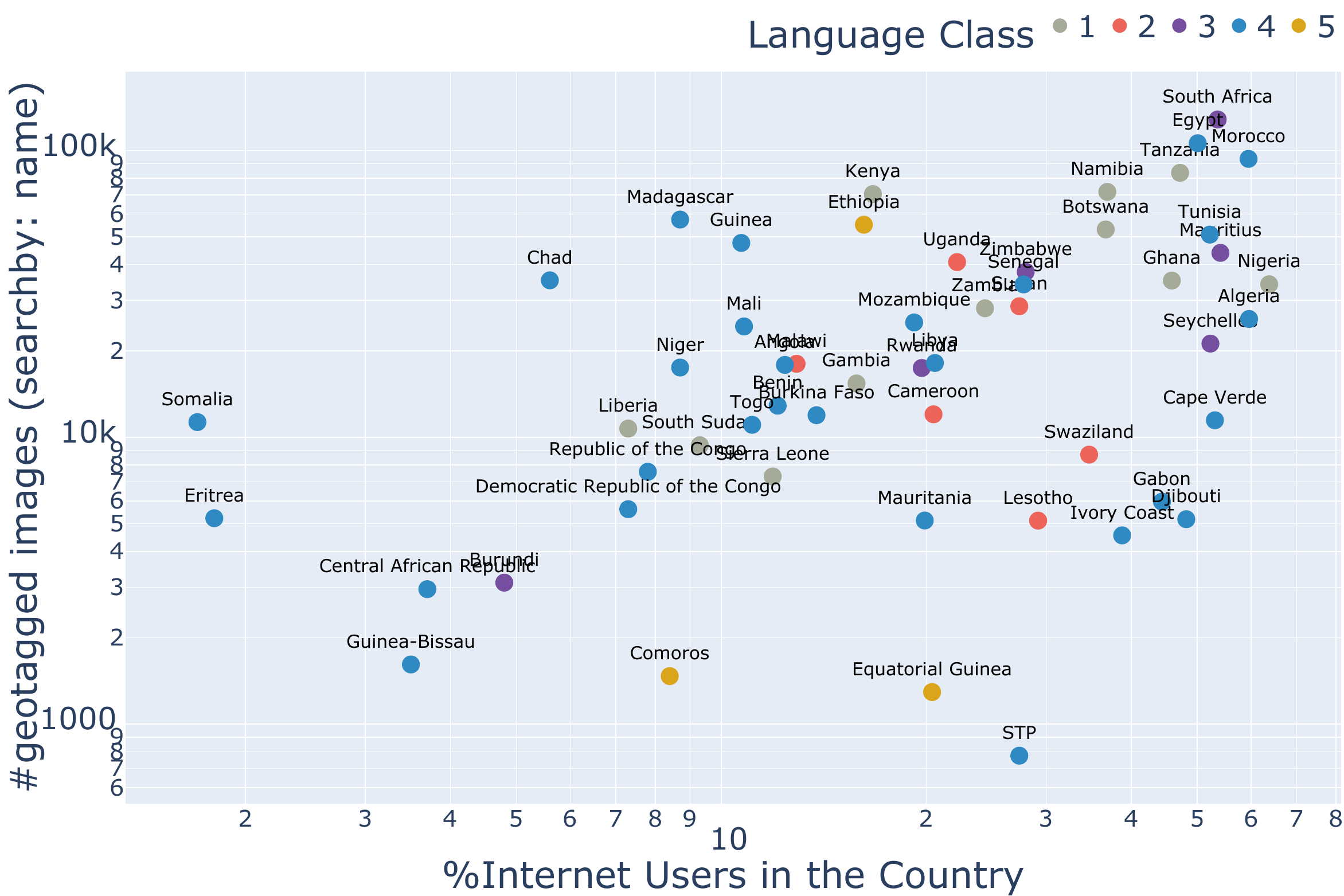}}
\hfill
\subfloat[]{\includegraphics[width=0.48\textwidth]{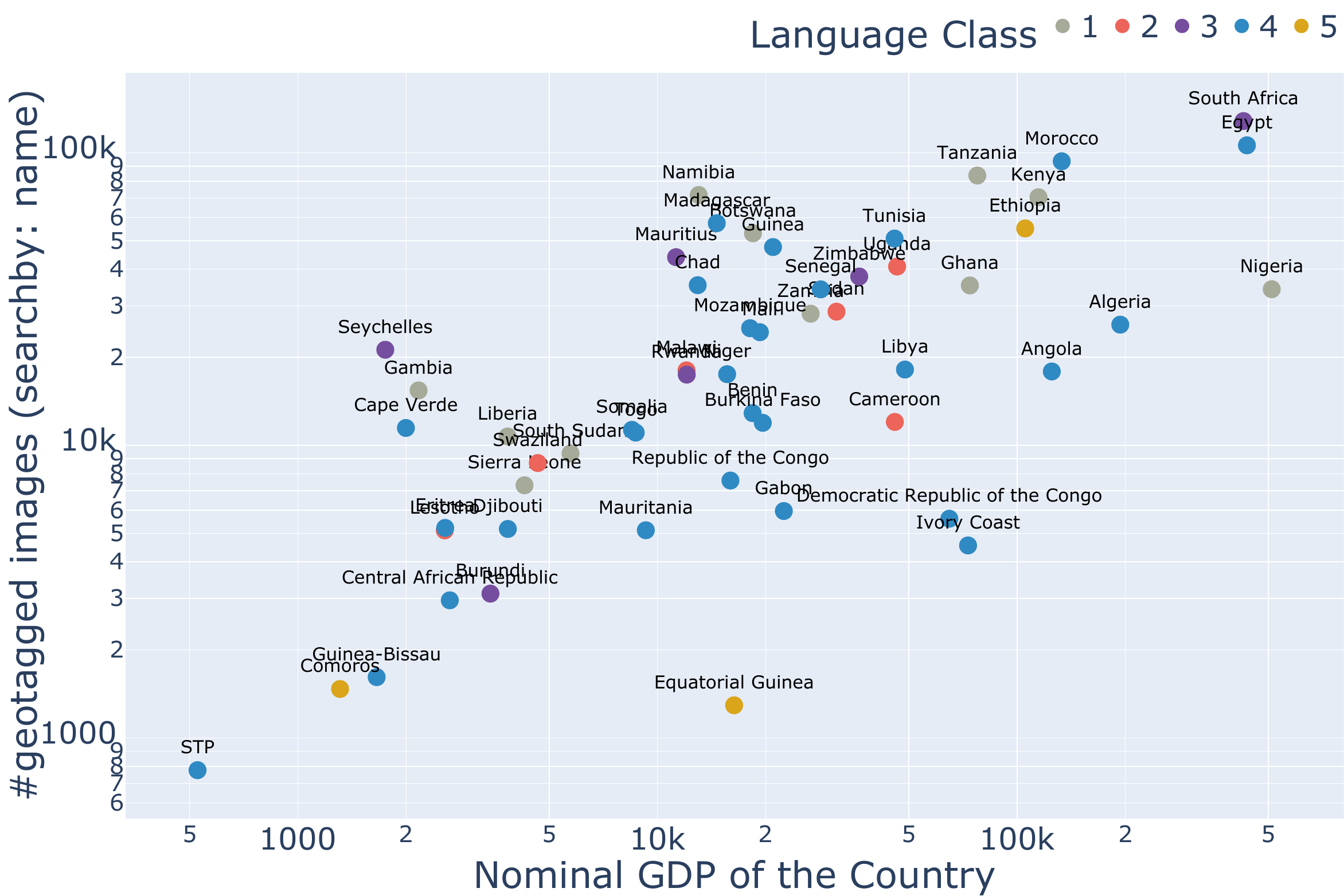}}
\caption{Plots showing the number of geotagged images as a function of (\textbf{a}) the percentage of Internet users in the country, and (\textbf{b}) the nominal GDP of the country. Data points are colored according to national dominant language class (see \Cref{sssec:dataavail}). In general, the number of geotagged images increased with increase internet usage and GDP, with no observable trend in language class; ``STP'' denotes the country Sao Tome and Principe.}
\label{fig:gdp_internet-usage}
\end{figure*}
\raggedbottom

\setlength{\tabcolsep}{0.34em}
\begin{table*}[ht!]
\footnotesize
\caption{Number of geotagged and total images, and percentage (rounded to 2 decimal places) of geotagged images for the five countries with the highest percentage of geotagged images according to query type.} 
\begin{tabularx}{\textwidth}{@{} *{2}{C} @{}}
    \label{table:top_geo_images}
     \begin{tabular}{@{} llll @{}}
        \toprule
        \multirow{2.3}{*}{}
        & \mcc[3]{query-by-name}          \\
         \midrule
                 Country &  \#geotagged &  \#all images &  \%geotagged \\
        \midrule
          Cape Verde &        11,465 &        43,283 &       26.49 \\
                 DRC &         5,614 &        21,560 &       26.04 \\
        South Africa &       128,294 &       513,082 &       25.00 \\
             Algeria &        25,849 &       105,254 &       24.56 \\
   Republic of Congo &         7,581 &        33,438 &       22.67 \\
        \bottomrule
     \end{tabular}    &  \label{table:top_geo_images_people}
          \begin{tabular}{@{} llll @{}}
        \toprule
        \multirow{2.3}{*}{}
        & \mcc[3]{query-by-name+people}          \\
         \midrule
                 Country &  \#geotagged &  \#all images &  \%geotagged \\
        \midrule
            Burkina Faso &         3,691 &        12,990 &       28.41 \\
                     DRC &         1,850 &         7,028 &       26.32 \\
       Republic of Congo &         2,110 &         8,249 &       25.58 \\
              Cape Verde &         1,333 &         5,390 &       24.73 \\
                Botswana &         1,085 &         4,664 &       23.26 \\
        \bottomrule
     \end{tabular}
\end{tabularx}
\label{table:most_geotagged_image_countries}
\end{table*}
\raggedbottom

\setlength{\tabcolsep}{0.34em}
\begin{table*}[ht!]
\footnotesize
\caption{Number of geotagged and total images, and percentage (rounded to 2 decimal places) of geotagged images for the five countries with the lowest percentage of geotagged images according to query type.} 
\begin{tabularx}{\textwidth}{@{} *{2}{C} @{}}
    \label{table:least_geo_images}
     \begin{tabular}{@{} llll @{}}
        \toprule
        \multirow{2.3}{*}{}
        & \mcc[3]{query-by-name}          \\
         \midrule
                 Country &  \#geotagged &  \#all images &  \%geotagged \\
        \midrule
         Lesotho &         5,121 &        43,282 &       11.83 \\
       Swaziland &         8,699 &        75,631 &       11.50 \\
         Somalia &        11,296 &       101,989 &       11.08 \\
         Burundi &         3,113 &        39,888 &        7.80 \\
          Rwanda &        17,443 &       250,469 &        6.96 \\  

        \bottomrule
     \end{tabular}    &  \label{table:least_geo_images_people}
          \begin{tabular}{@{} llll @{}}
        \toprule
        \multirow{2.3}{*}{}
        & \mcc[3]{query-by-name+people}          \\
         \midrule
                 Country &  \#geotagged &  \#all images &  \%geotagged \\
        \midrule
           Zambia &         3,103 &        33,063 &        9.39 \\
Equatorial Guinea &          168 &         2,081 &        8.07 \\
            Gabon &          476 &         7,286 &        6.53 \\
          Burundi &          533 &         9,862 &        5.40 \\
           Rwanda &         3,461 &        76,521 &        4.52 \\
        \bottomrule
     \end{tabular}
\end{tabularx}
\label{table:lowest_geotagged_image_countries}
\end{table*}
\raggedbottom

\section{Results and Discussion}

\subsection{Data Availability and Geographic Distribution} \label{sssec:dataavail}

There were very few geotagged images from Africa, as shown in \Cref{fig:rwi_geo_name,fig:rwi_geo_people} and in Appendix \Cref{fig:name_geo_cts,fig:people_geo_cts}, and \Cref{table:geo_images,table:geo_images_people}. In terms of total geotagged image counts with query-by-name and query-by-name+people from African nations, South Africa ($128,294$ \& $46,021$) and Egypt ($105,996$ \& $ 32,869$) had the highest counts, and Equatorial Guinea ($1,293$ \& $168$) and Sao Tome and Principe ($776$ \& $116$) had the lowest counts.  Cape Verde had the highest percentage of geotagged images ($26.49\%$) from query-by-name and Burkina Faso ($28.41\%$) had the highest from query-by-name+people. By contrast, Rwanda had the lowest percentage of geotagged images ($6.96\%$, $4.52\%$) from both query-by-name and query-by-name+people. African nations with the highest and lowest percentages of geotagged images are summarized in  \Cref{table:most_geotagged_image_countries} and \Cref{table:lowest_geotagged_image_countries}.\\
\textbf{Thus}, the low number of African geotagged images indicates the ineffectiveness of Flickr scraping as a data collection methodology in this region and, therefore, a need to explore alternative geodiverse data collection methods, \eg utilizing manual data collection.

The population-matched European countries had higher numbers and percentages of geotagged images than the corresponding African countries, as is further emphasized in \Cref{table:europe_africa_geotag_counts}. For example, with query-by-name, despite relatively similar population sizes, the percentage change of the number of geotagged images from Sierra Leone to Switzerland is $1673.48\%$, that is, $18\times$ as many total geotagged images as Sierra Leone as shown in \Cref{table:europe_africa_geotag_counts}. \\
\textbf{Thus}, African countries had far fewer images (both geotagged and non-geotagged) than the corresponding European countries of similar population size.  We recommend that computer vision experts be cognizant of this discrepancy in Flickr scraped datasets and to consider the corresponding potential for bias when training computer vision models. 

We analyzed the statistical effect of factors that might potentially affect taking, uploading and tagging images on Flickr;  population-size, internet usage, official language, and countries' GDP.\\
In general, the number of geotagged images increased with population size (correlation: $0.412$ \& $0.538$, query-by-name and query-by-name+people respectively), internet usage ($0.474$ \& $0.385$), and GDP ($0.599$ \& $0.748$); the latter two are shown in \Cref{fig:gdp_internet-usage}. 
An investigation of the effect of these variables on the number of geotagged images was found to be statistically significant:  (population size: \textit{p-value} $=0.0019$ \& \textit{p-value} $=0.000119$, query-by-name and query-by-name+people respectively), (internet usage: \textit{p-value} $=0.00029$ \& \textit{p-value} $=0.003999$), and (GDP: \textit{p-value} $=0.160$ \& \textit{p-value} $=0.059$). By contrast, official language was not found to have a meaningful correlation to the number of geotagged images (\textit{p-value} $=0.2021$ \& \textit{p-value} $=0.846$). 
Because image dataset queries are typically done in English, to assess the impact of dominant national languages relative to English on geotagged data availability, we coded each of the countries' official languages (\cite{official_languages}) according to five categories for analysis: $1$- (English is the only official language), $2$- (English is among the two official languages), $3$- (English among atleast three official languages), $4$- (English not among atmost three official languages), and $5$- (English not among atleast three official languages). No correlation was determined for any language category. \\
\textbf{Thus}, when data collection is required in regions with lower population size, internet usage, and/or GDP, we recommend the use of local, manual data collection techniques in lieu of webscraping whenever feasible. Additionally, RWI information may be useful when assessing diverse areas for data collection.

\subsection{Tags and Licenses: Query-Based and Applicability Limitations} \label{sssec:tags_and_limits}

\setlength{\tabcolsep}{0.34em}
\begin{table*}[!thp]
\footnotesize
\caption{Number of geotagged and total images, and percentage (rounded to 2 decimal places) of geotagged images for population-matched African and European nations (query-by-name) side-by side, e.g Switzerland and Sierra Leone on line 1.} 
\begin{tabularx}{\textwidth}{@{} *{2}{C} @{}}
    \label{table:europe}
     \begin{tabular}{@{} lllll @{}}
        \toprule
        \multirow{2.3}{*}{}
        & \mcc[3]{European Countries}          \\
         \midrule
        Country &  population &  \#geotagged &  \#all images & \%geotagged \\
        \midrule
        Switzerland &           8.75M &      129,518 &    535,843&     24.17\\
            Finland &           5.55M &      119,901 &     522,637&     22.94 \\
           Slovenia &          2.11M &       86,630 &       371,584&     23.31 \\
             Cyprus &           918.10k &    77,826 &  371,504&     20.95 \\
        \bottomrule
     \end{tabular}    &  \label{table:africa}
          \begin{tabular}{@{} lllll @{}}
        \toprule
        \multirow{2.3}{*}{}
        & \mcc[3]{African Countries}          \\
         \midrule
        Country &  population & \#geotagged &  \#all images & \%geotagged \\
        \midrule
     Sierra Leone &   8.30M &     7,303 &    52,530&     13.90\\
              CAF &   5.60M &   2,954 &      19,901&     14.84 \\
          Lesotho &   2.10M &    5,121 &      43,282&     11.83 \\
         Djibouti &   976.11k  &   5,179 &     36,029&     14.37 \\
        \bottomrule
     \end{tabular}
\end{tabularx}
\label{table:europe_africa_geotag_counts}
\end{table*}
\raggedbottom
\begin{figure}[!htp]
\captionsetup{singlelinecheck=off}
\begin{subfigure}{0.48\textwidth}
    \includegraphics[width=1.05\textwidth]{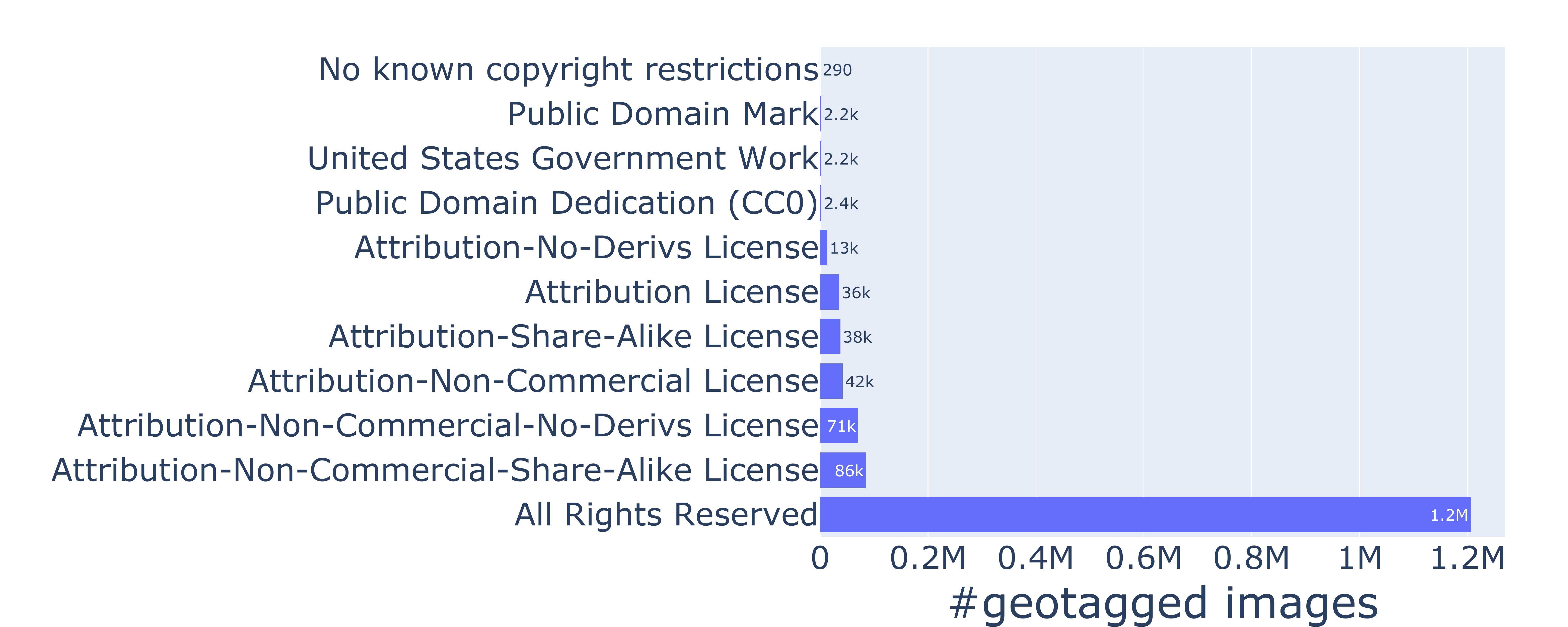}
    \caption{License information for query-by-name}
    \label{fig:license_bigdata_name}
\end{subfigure}
\hfill
\begin{subfigure}{0.48\textwidth}
    \includegraphics[width=1.05\textwidth]{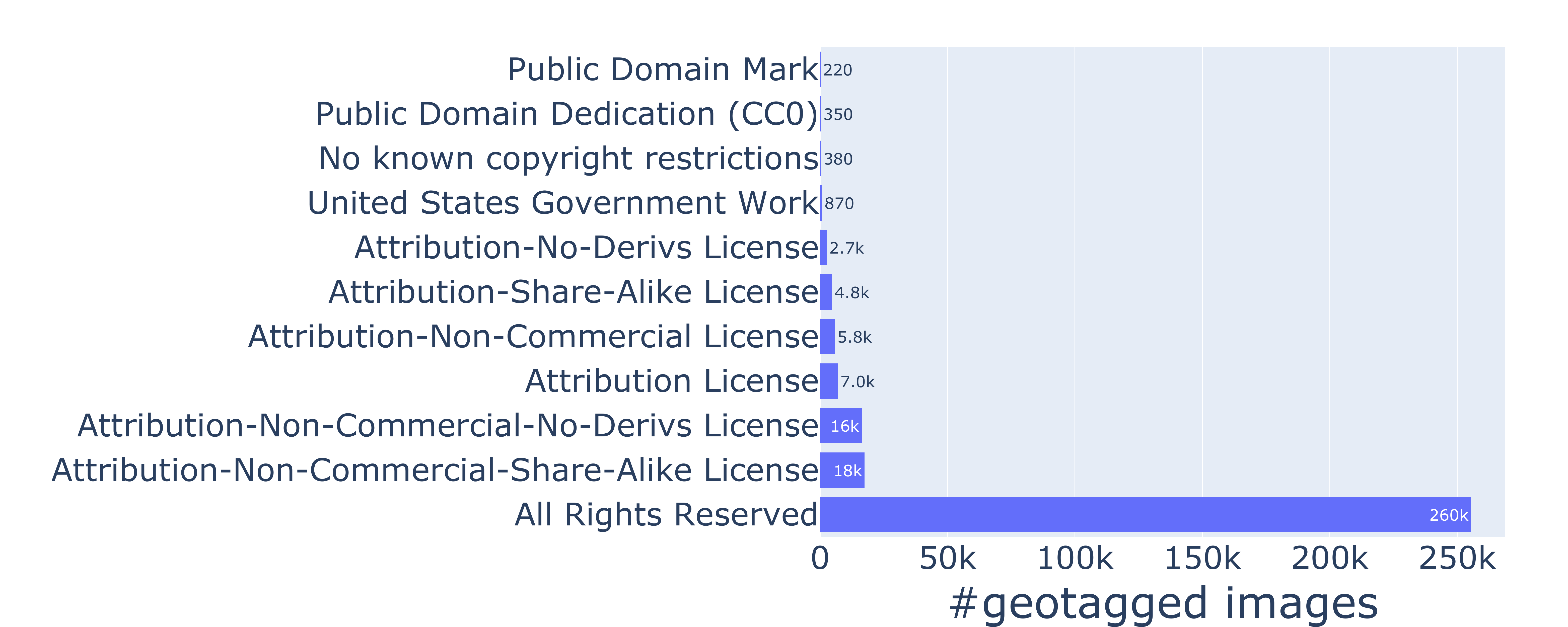}
    \caption{License information for query-by-name+people}
    \label{fig:license_bigdata_people}
\end{subfigure}
\caption[foo bar]{Bar charts showing the total count of each Flickr license type for the entire image datasets as a function of (\textbf{a}) ``query-by-name'' and (\textbf{b}) ``query-by-name+people''. The most restrictive license type is by far the most common, ``All Rights Reserved'', likely because it is the Flickr default option.}. 
\label{fig:license_bigdata}
\end{figure}
\raggedbottom

The use of both country name tags and geotags (latitude/longitude) was found to be necessary to ensure data was accurately sourced from the country of interest.
The na\"{\i}ve country name querying method is particularly limiting when applied to certain nations, such as \emph{Chad}, \emph{Guinea}, and \emph{Republic of Congo}. Images from query by \emph{Chad} were predominantly geotagged from \emph{United States} ($54.63\%$), \emph{United Kingdom} ($16.58\%$), and \emph{Canada} ($9.30\%$), with only $5.14\%$ of the images coming from \emph{Chad} according to geotag location results. In total images from query by \emph{Chad} were geotagged from $129$ countries. Likewise, images from query by \emph{Guinea} predominantly came from \emph{Papua New Guinea} ($29.97\%$) and \emph{United States} ($10.84\%$), with geotags from $190$ countries. Finally, image geotags from  query by \emph{Republic of Congo} mainly reflected the following countries; \emph{Congo, The Democratic Republic of the} ($42.14\%$), \emph{United Kingdom} ($11.34\%$), and \emph{United States} ($7.55\%$). \\
\textbf{Thus}, we conclude that reliance upon country name queries is insufficient for constructing a geodiverse dataset in the absence of more robust geolocation data. We recommend that data collectors consider using RWI data to source more geographically diverse visual data.

We furthermore report the most frequent tags as the name of the place where the image was taken, for example ``Africa'' and the country name, in addition to image contents. The least frequent tags were usually those in foreign languages and whose meanings were hard to decipher because of multiple concatenated words. \\
\textbf{Thus}, in an African context, the utilization of image tags alone to generate datasets with specific image content may be less reliable due to the variable nature of selected tags; we believe this warrants future exploration. 

Additionally, the vast majority of images with query-by-name  and query-by-name+people respectively are licensed as ``All Rights Reserved'' ($80.46\%$, $81.99\%$), indicating the Flickr default setting when images are uploaded to the platform (see \Cref{fig:license_bigdata}). \\
\textbf{Thus}, those constructing datasets using Flickr Africa data must be aware that most images are unavailable for model training and evaluation without copyright violations, thereby further limiting ethical access to geographically diverse data.

\subsection{Geodiversity by RWI}

To assess the impact of wealth on the availability of geotagged image data, we examine image counts by RWI values binned into $10$ percentile groups, $G1$-$G10$. For most nations, the majority of image data comes from the middle RWI regions ($G4$, $G5, G6$ and $G7$) and the least from low RWI regions ($G1$, $G2$ and $G3$). However, this is not always the case, \eg Madagascar and Algeria from which data is sourced from low-income areas (along main roads close to national parks) or high-income areas (in major cities), respectively. \\
\textbf{Thus}, RWI has potential as a mechanism for constructing geo-diverse datasets in future work.

\begin{figure*}[thbp]
\centering
\subfloat[]{\includegraphics[width=0.81\textwidth]{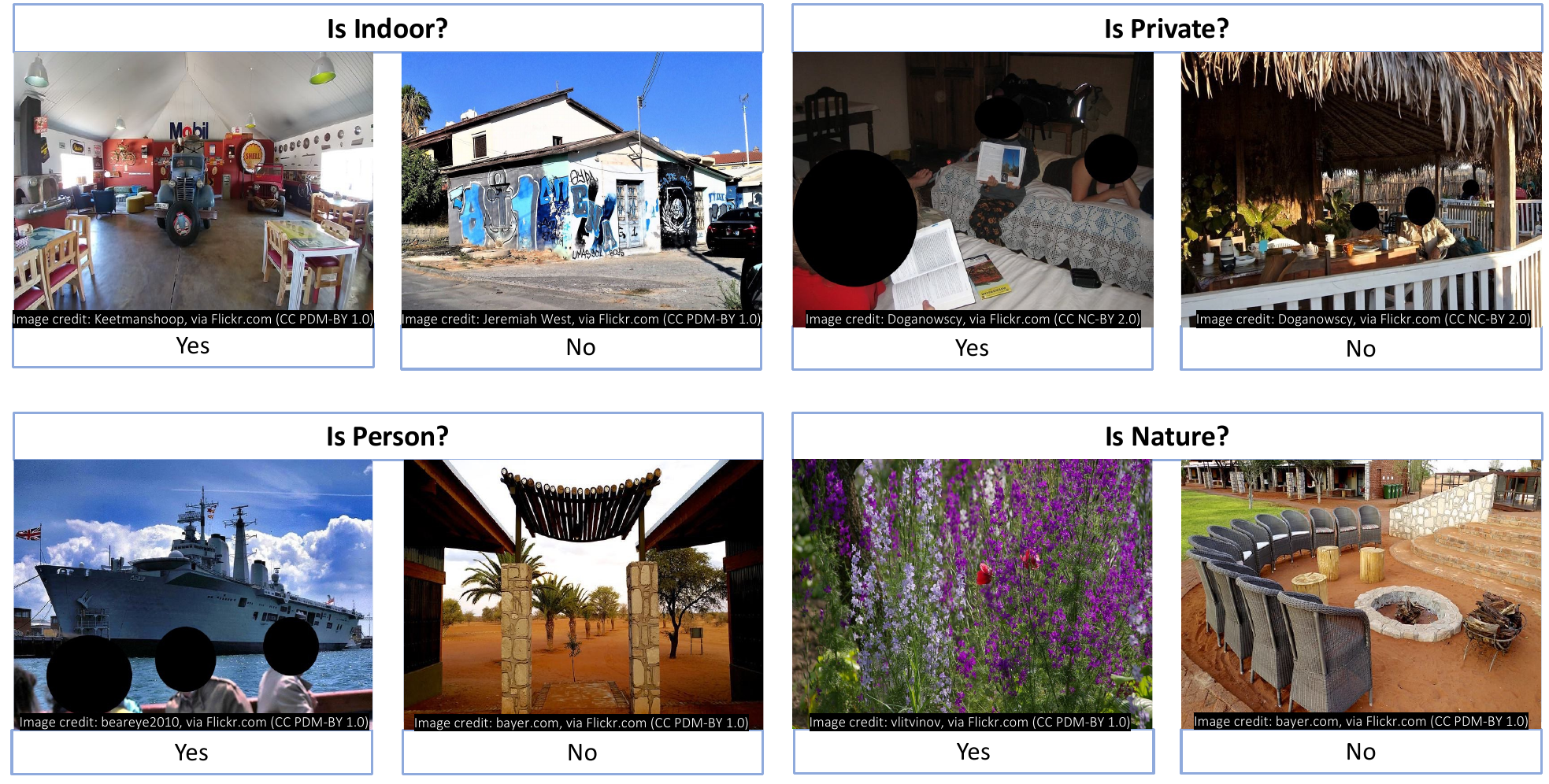}
\label{fig:sample-images}}
\hfill
\subfloat[]{\includegraphics[width=0.81\textwidth]{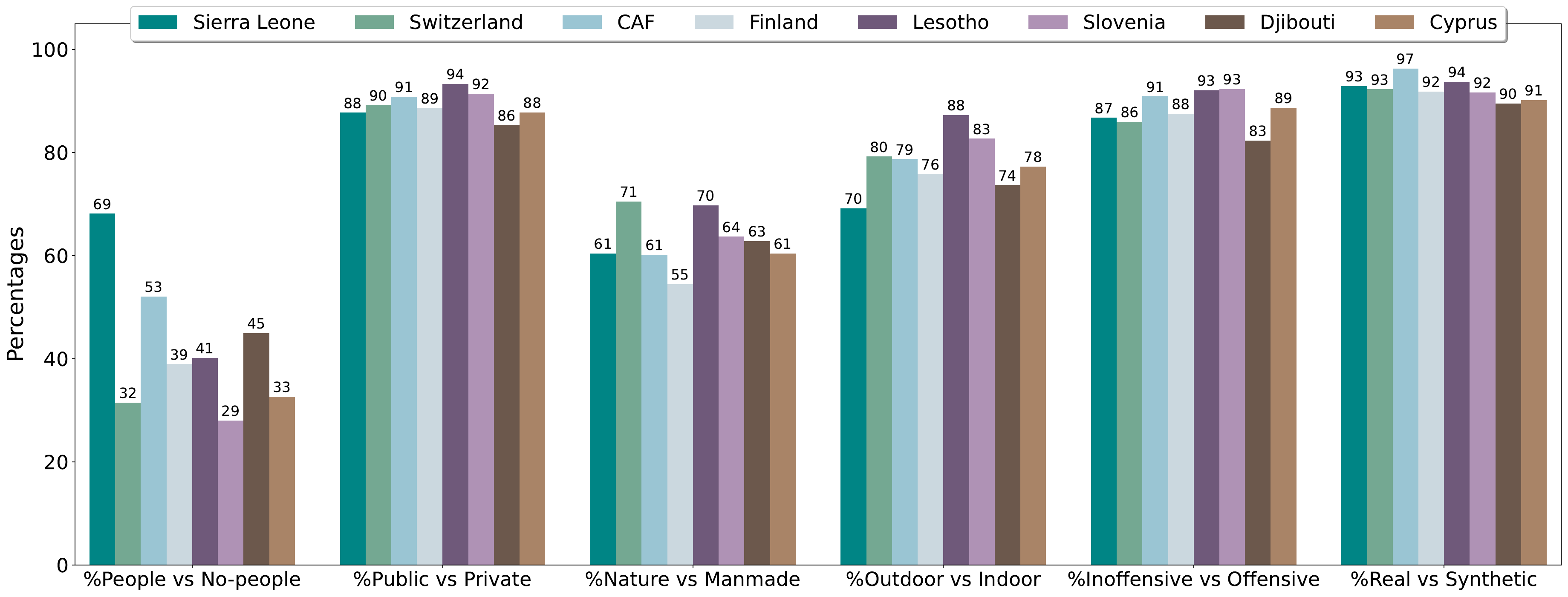}\label{fig:all_vars_bars}}
\caption{(\textbf{a}) Sample images sourced from African nations, and (\textbf{b}) image content percentages for select African and European nations, according to AMT workers reporting across six binary attributes. Workers were asked to report if the images contained people, public settings, nature content, outdoor settings, and inoffensive content, and if they appeared to be real images. Image content was found to be similar percentage-wise across different nations, although far fewer images overall were captured in the population-matched African nations in comparison to corresponding European nations. Images displayed here were selected among those with permissible licenses with face obfuscation for display purposes only. Full images and license information can be found in Appendix  \Cref{fig:v1,fig:v4,fig:v2,fig:v3,fig:v5,fig:v6,fig:v7,fig:v8}.}
\label{fig:all_amt_variables}
\end{figure*}
\raggedbottom
\begin{figure*}[thbp]
\centering
\includegraphics[width=0.81\textwidth]{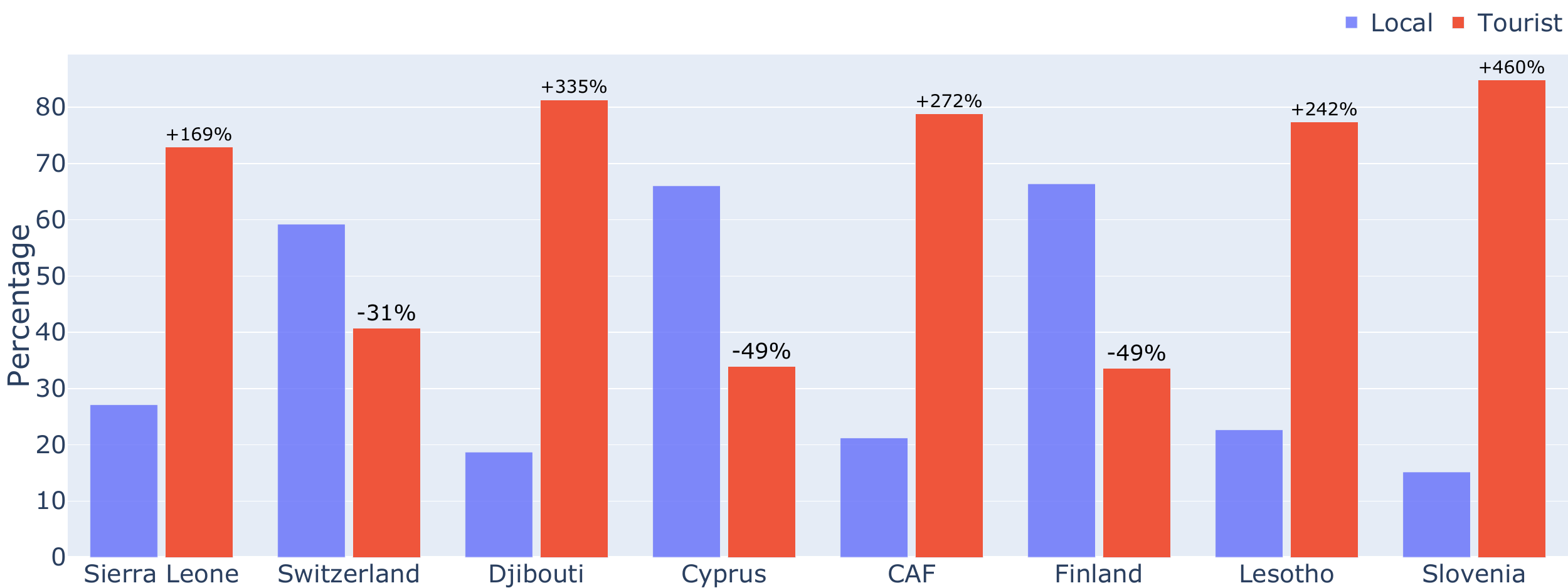}
\caption{Bar chart showing the percentage out of 2,000 images from each population-matched African/European nation pair via query-by-name as taken by locals (blue) and tourists (red), according to Flickr users' reported locations. The percent change from local to tourist percentage value is additionally indicated for each nation. Images from African countries were predominantly taken by foreigners whereas those from higher nominal GDP European countries were predominantly taken by locals. ``CAF'' indicates the country Central African Republic. }
\label{fig:locals_vs_non}
\end{figure*}
\raggedbottom

\subsection{Image Content}

By utilizing crowdsourced annotations, we examine 16,000 images' content data across 2,000 images from each population-matched African and European nation pair (identical to the image subset in \Cref{ssec:nonlocal}). Sample images by attribute and results for matched African/European nation pairs are shown in \Cref{fig:sample-images} for each binary attribute with the exception of ``offensive'' vs. ``inoffensive'' content and with manually obscured human faces. We collected information about these six attributes to gauge the applicability of African-sourced image datasets for various computer vision tasks: \eg the presence of people for human-centric tasks such as pose estimation, body part segmentation or face detection; the prevalence of indoor/private settings for specific object recognition tasks; or real/appropriate image content for training dataset viability. Likewise, we originally hypothesized that African images were more likely to be taken by foreigners (which was found to be supported by the data; see  \Cref{ssec:nonlocal}); this motivated the count of nature-centric images.

The AMT results revealed that query-by-name images from both African and European countries were predominantly  ``real'' ($93.47$\% and $91.94$\%), ``inoffensive'' ($88.51$\% and $89.41$\%), ``outdoor'' ($77.89$\% and $79.28$\%), ``public'' ($90.27$\% and $90.19$\%), and ``nature'' ($63.68$\% and $62.96$\%) images. There were negligible variations across nations for the percentage of ``real', ``outdoor'', ``public'', and ``nature'' images. However, as shown in \Cref{fig:all_vars_bars}, nations varied in percentage of people in images, and in general most nations' images did not contain people.  For example, $69\%$ of Sierra Leone's $2000$ sampled geotagged images contained people, while only $33\%$ of Djibouti's $2000$ sampled geotagged images contained people. \\
\textbf{Thus}, although no major differences between African and European image content were observed according to the six attributes considered, we believe these findings are important in the context of data regarding data quantity. Given that image content was fairly similar across most attributes annotated, and there exist far fewer geotagged images from Africa (see \Cref{table:europe_africa_geotag_counts}), we anticipate insufficient African data availability for certain computer vision tasks. For example, the lower prevalence of images captured in ``private'' and ``indoor'' settings indicates \eg household object image data inaccessibility, which thereby impacts downstream object recognition system models consistent with the findings of \citet{obj-notfor-everone}.

\begin{figure*}[!thbp]
\centering
\includegraphics[width=0.85\textwidth]{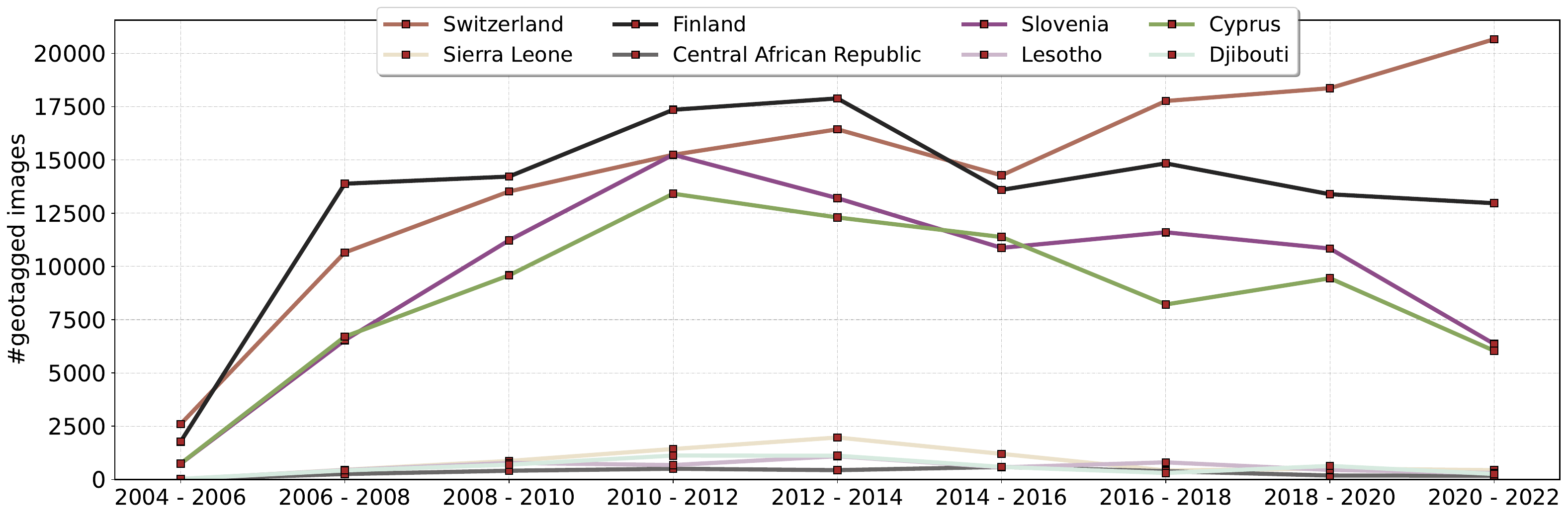}
\caption{A comparison of the number of geotagged images from select African and European nations in approximately 2-year time ranges, as queried by country name. In general, a high number of images were uploaded in the dates in range 2012-2014, with the exception of Switzerland which had high upload volume in 2020-2022.}
\label{fig:africa_europe_images_temp}
\end{figure*}
\raggedbottom

\subsection{Local vs. Non-Local Representation} 
\label{ssec:nonlocal}

Beyond analyzing the image content of the $2,000$ randomly-sampled images from each of the $8$ nations, we examined the local vs. non-local status of those Flickr users who captured and uploaded the geotagged images. An assessment of the residence or origin of the Flickr users revealed that for the African geotagged nations, images were far more likely to be taken by foreigners than locals whereas the opposite trend was observed for higher-GDP European nations, according to comparisons between geotags and Flickr users' reported locations. For Sierra Leone, $+169\%$ of images were captured by foreigners compared to locals, while for Switzerland it was -$31\%$. The same trend applies to Djibouti and Cyprus (+$335\%$ and -$49\%$) and CAF and Finland (+$272\%$ and -$49\%$); results are reported in \Cref{fig:locals_vs_non}. 
In general, more images across all nations were taken by non-locals compared to the smaller $2,000$-image datasets (\eg see Appendix \Cref{fig:alldata-Djibouti-Temporal,fig:all-data-tours}). However, the prior trends held in the sense that when African countries were considered, far more geotagged images were taken by non-locals than in comparable European nations, e.g., +$329\%$ for Sierra Leone versus +$39\%$ for Switzerland. 
Additionally, a random inspection of the Flickr map\footnote{When we inspected the map (\url{https://www.flickr.com/map/}) on 06-16-2023, 2/2 of the geotagged images were taken by France and Spanish photographers} also further shows that images geotagged in Africa are less likely to be taken by the locals. \\
\textbf{Thus}, the prevalence of non-local representation may explain the image content results described in the previous section, as Flickr users from similar backgrounds may contribute image data from both Africa and Europe. AI practitioners should be wary of stereotyped representations of African life within such datasets given that these images are typically taken by foreigners in public, outdoor locations. 
Additionally, current methodologies for image dataset collection are unlikely to capture visual data pertaining to the private, daily life of African people nor visual information the locals of each country consider to be  important, resulting in biases propagated by AI systems trained on such data.

\subsection{Temporal Analysis}
We performed a temporal analysis to investigate and contextualize the data in time according to data quantity, relative wealth index (RWI) at location of capture, license type and Flickr user origin. 
We studied the geotagged images distribution in the in approximately 2-year spans time ranges.

\paragraph{Number of geotagged images} 
In general, there were relatively fewer geotagged images in the years 2004-2006 and 2020-2022, as shown in 
\Cref{fig:africa_europe_images_temp} for population-matched African/European nations and in the Appendix  \Cref{table:temporal_name_counts,table:temporal_people_counts} for all African nations. The image distribution could be the result of factors including less internet penetration and popularity of Flickr from 2004-2006 reducing image uploads, and the COVID-19 pandemic limiting outdoor activities from 2020-2022. This trend held in all analyzed countries with the exception of Switzerland; there, the highest number of images was uploaded in the date range 2020-2022, as shown in \Cref{fig:africa_europe_images_temp}.
The highest number of uploaded and subsequently downloaded geotagged images for most nations came from 2010-2014, potentially explained by the growth of internet usage and exposure to Flickr in different countries within this time span.

\paragraph{RWI regions of image uploads} We explored trends in dominant RWI groups per nation over time, in order to determine if there were observable shifts towards images sourced from higher or lower RWI regions. Over the time range of 2004 to 2022, query-by-name images from Botswana, Libya, Namibia, South Africa, Tunisia, Swaziland, Uganda, Zambia, and Zimbabwe all came from the middle RWI regions.  Images from Morocco consistently came from the upper RWI regions.  On the other hand, query-by-name+people images from Rwanda and Swaziland all came from the middle RWI regions and those from Morocco all came from upper RWI regions.
Lower and middle RWI regions countries had their data distributions varying between lower and middle RWI regions over the years. 
Countries whose images were from predominantly upper RWI regions had their data distributions varying between middle and upper RWI regions over the years.

\paragraph{Licenses of the uploaded images} We analyzed the quantity of images with various Flickr license options.
Images were found to have predominantly the ``All Rights Reserved'' license type across all time ranges analyzed; as noted in \Cref{sssec:tags_and_limits}, this substantially limits data usage. There were almost no images licensed under the ``Public Domain Dedication (CC0) CC'' and ``Public Domain Mark CC'' among those uploaded to Flickr from $2004$ to $2022$.

\paragraph{Local vs. non-local representation} 
We performed a temporal analysis of the geotagged images to investigate the local vs. non-local status of Flickr users. For the $2,000$ randomly sampled images from the $8$ countries analyzed for image content, we observed differences in sampling dates: that is, Cyprus, Slovenia, and Finland images were mainly sampled from $2004$ to $2008$; Switzerland images were mainly sampled from $2004$ to $2006$; and the African nation images were were mainly sampled from $2004$ to $2012$. Following these results, we repeated the temporal analysis across all images sourced from each of the $8$ nations. In general, more images across all nations were taken by non-locals compared to the smaller $2,000$-image datasets. However, the prior trends held in the sense that when African countries were considered, far more geotagged images were taken by non-locals than in comparable European nations, e.g., +$329\%$ for Sierra Leone versus +$39\%$ for Switzerland. \Cref{ssec:nonlocal} describes implications of non-local representation in image data from Africa; namely, the risk of an ``othering'' phenomenon and its impact on downstream bias in AI systems.

\section{Conclusion and Future Work}\label{sec:conclusion}

Geographical context shapes data, and data shapes the performance of models trained using such data. The key findings from our Flickr Africa data analysis (1) expose the limitations of current large-scale image data collection methodologies, and (2) expose unique data challenges to Africa, including the lack of data crucial to specific domains (\eg a researcher cannot source sufficient, representative household object data if very few images are taken within indoor/private scenes). Notably, we reported on the extreme lack of data availability when compared to wealthy European nations; for instance when querying by country name, Switzerland had 18x the geotagged image data as Sierra Leone, an African nation of similar population size (8.75M vs. 8.30M, respectively), while Sao Tome and Principle only had (776, 116) geotagged images in total (depending on query). Moreover, data may be even less accessible according to use case, given that most of the Flickr Africa data has a restrictive use license, and certain image content attributes were found to appear less frequently (\eg private and indoor settings). Nationally, higher quantities of geotagged image data was found to positively correlate with population size, GDP, and Internet usage, but no significant correlation was discovered based on dominant national languages. Additionally, we interrogate where African image data comes from: generally from middle-wealth regions as measured intra-nationally by RWI, though this differs by nation; and with images mainly taken by foreigners, though the opposite trend is identified in wealthier European nations. We discussed how AI systems may propagate biases in accordance with the stereotyped representation of African life by outsiders. Temporal analyses were performed and demonstrated that certain trends, such as dominant RWI region, prevalence of restrictive license type, and non-local representation of African nations in geotagged images held over time. 

Looking forward, we encourage new scholarship centering novel methods for sourcing geodiverse datasets and measuring new forms of geodiversity specific to Africa, such as analyses of tribal diversity as opposed to the more commonly studied diversity by race/ethnicity. We openly provide our large-scale dataset to enable future researchers to utilize and augment Flickr Africa for model evaluations across a wide domain of computer vision tasks; likewise, more rigorous bias identification methods (\eg \cite{revise}) may uncover still more limitations. Finally, we would be interested to explore the extent to which privacy and consent are respected in Africa.

\begin{acks}
We wish to thank Jerone Andrews and Dora Zhao in particular for their expertise and assistance with our work pertaining to crowdsourcing. We also express gratitude to the whole SONY AI Ethics team, especially William Thong for great discussions and informative feedback on this research. Lastly, we would like to thank the anonymous reviewers for the insightful feedback that helped improve our paper.
\end{acks}

\bibliographystyle{ACM-Reference-Format}
\bibliography{arxiv_main}

\onecolumn
\newpage
\appendix

\section{Supplemental Data}\label{sec:app-expe-results}

\setcounter{table}{0}
\renewcommand{\thetable}{A\arabic{table}}
\setcounter{figure}{0}
\renewcommand{\thefigure}{A\arabic{figure}}

\subsection{Geotagged image distribution}


\begin{figure*}[thbp]
\centering
\subfloat[]{\includegraphics[width=0.48\textwidth]{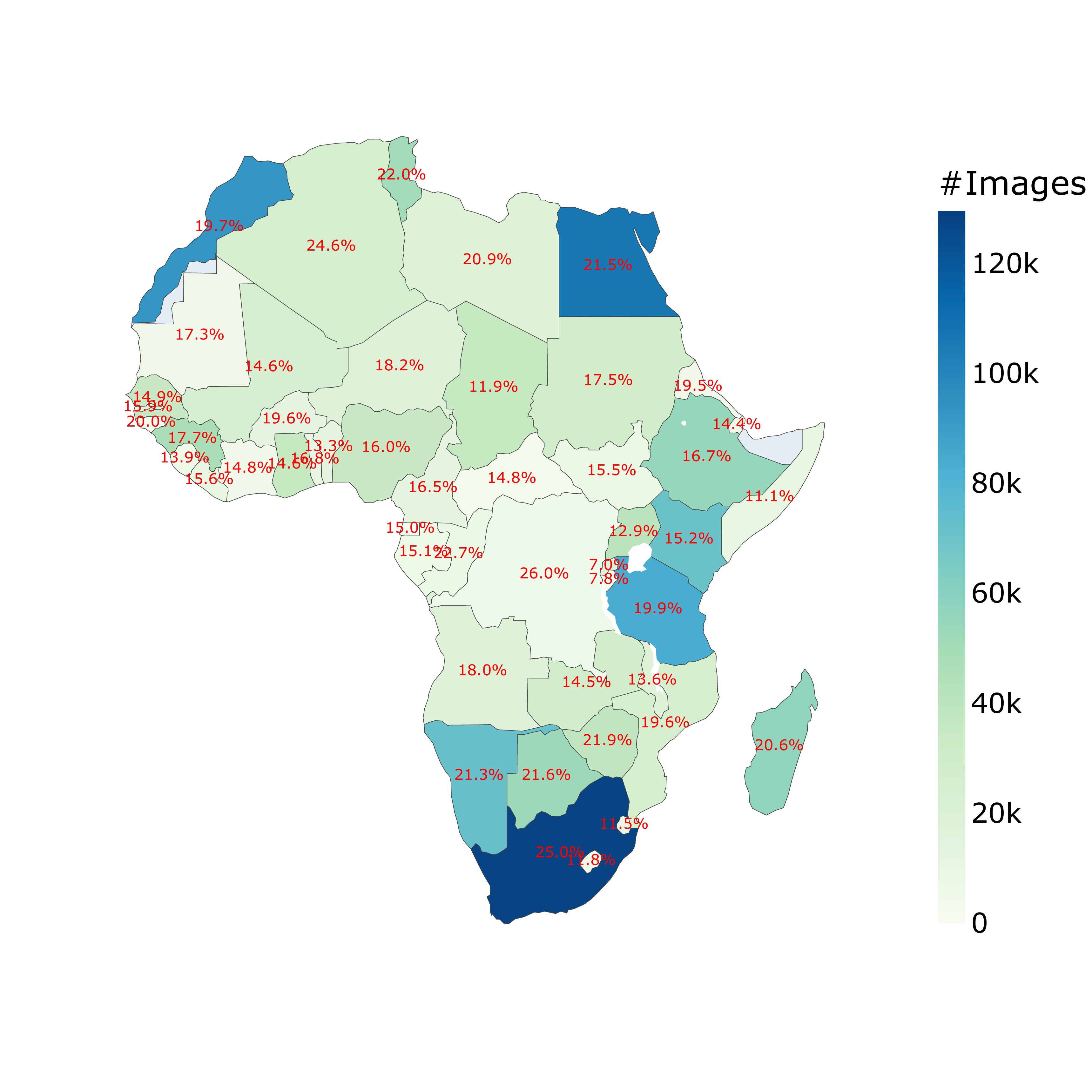} \label{fig:name_geo_cts}}
\subfloat[]{\includegraphics[width=0.48\textwidth]{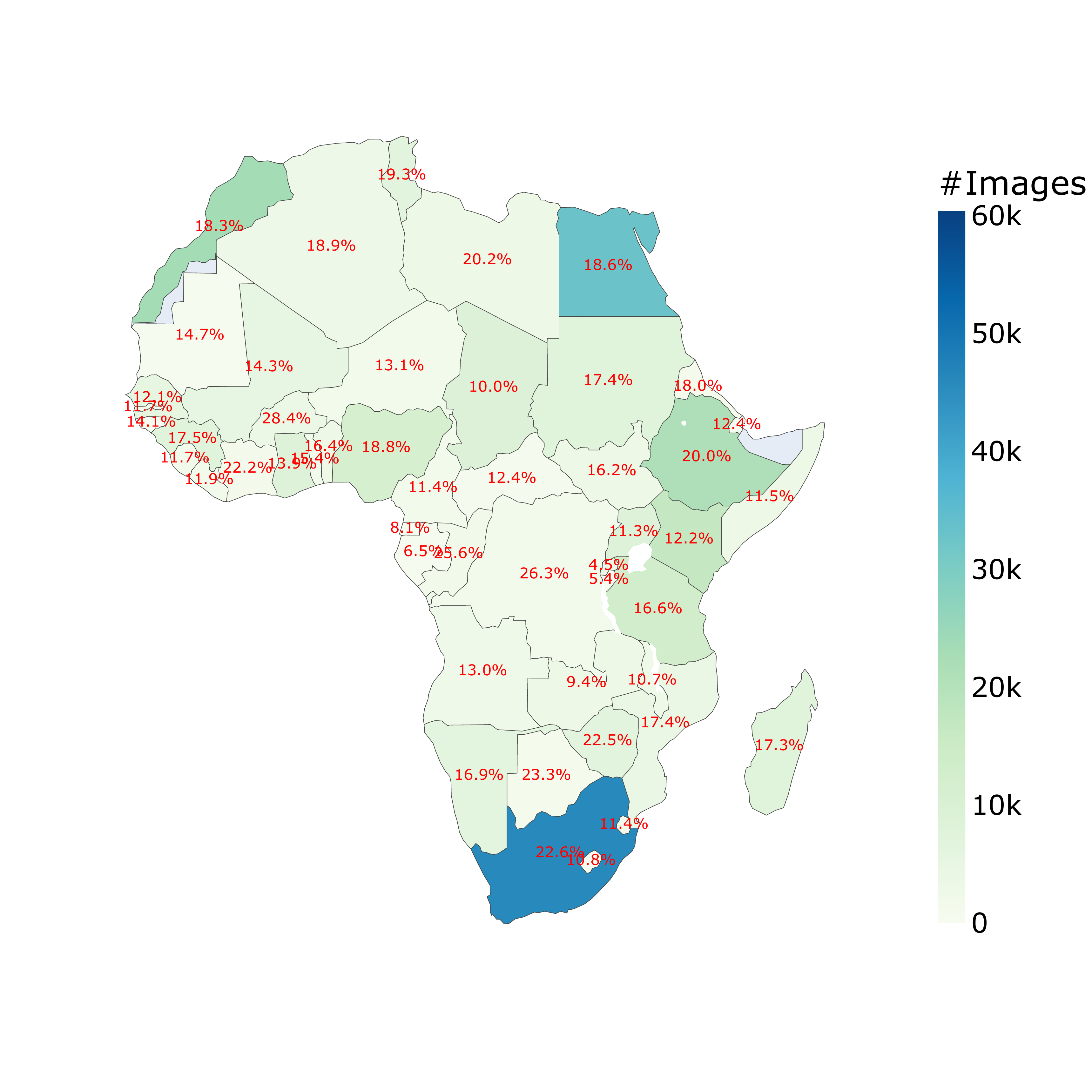} \label{fig:people_geo_cts}}
\caption{Percentage geotagged images out of total image count by nation for ``query-by-name'' and ``query-by-name+people'' and associated geotagged images count, (\textbf{a}) and (\textbf{b}) respectively. The numbers denote the percentage of geotagged images out of the total number of images. Nations are colored according to total number of geotagged images. South Africa had the highest number of geotagged images from both ``query-by-name'' and ``query-by-name+people'' ($128,294$ and $46,021$, respectively). While Cape Verde had the highest percentage of geotagged images ($26.49\%$) from query-by-name and  Burkina Faso ($28.41\%$) from query-by-name+people. By contrast, Sao Tome and Principe had the smallest number of geotagged images ($776$, $116$) and Rwanda had the lowest percentage of geotagged images ($6.96\%$, $4.52\%$) from ``query-by-name'' and ``query-by-name+people''.}
\label{fig:name_people_geoimages_cts}
\end{figure*}
\raggedbottom

\begin{figure*}[ht!]
\centering
\subfloat[Algeria]{\includegraphics[width=0.49\textwidth]{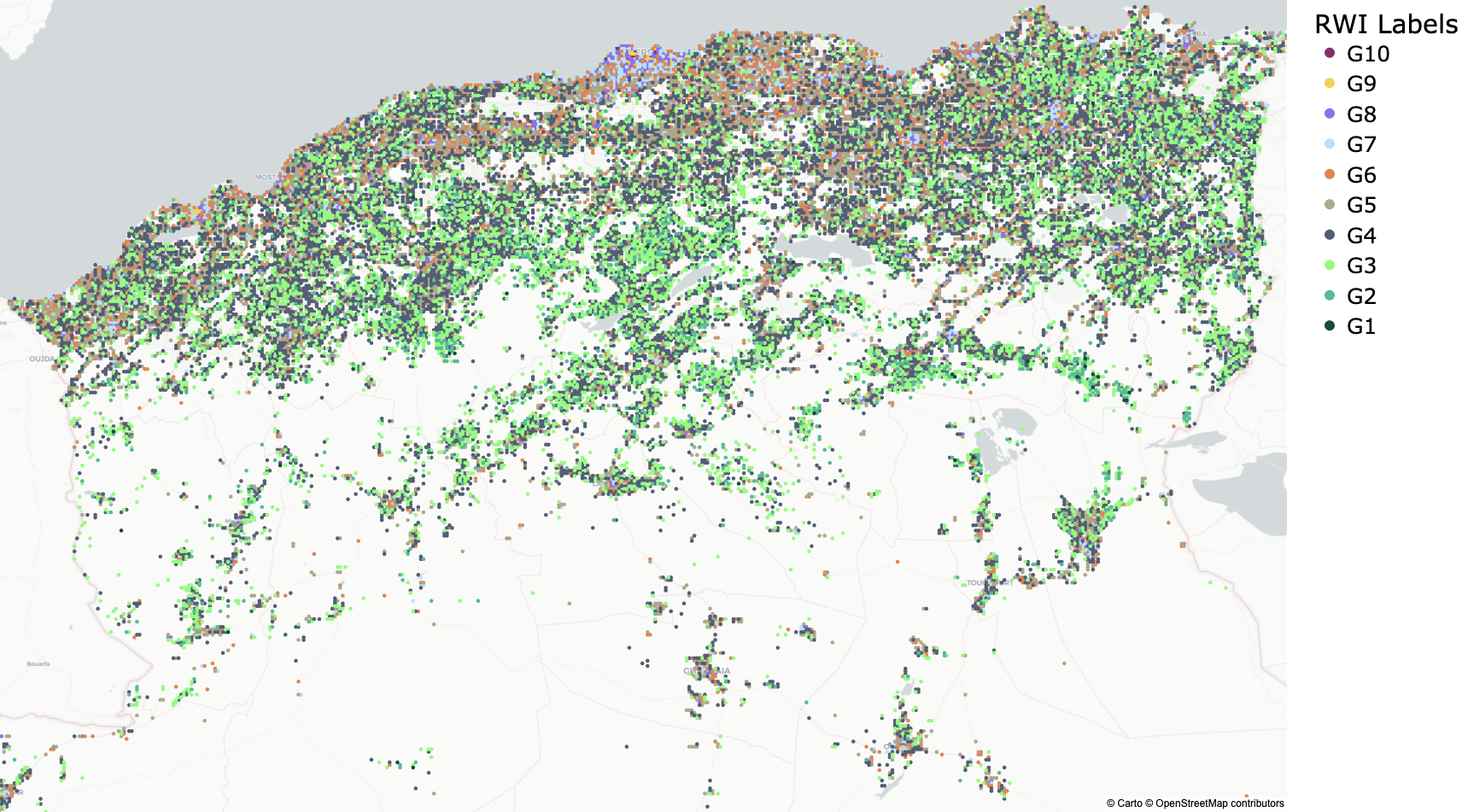} \label{fig:Algeria-rwi}}
\subfloat[Algeria.]{\includegraphics[width=0.48\textwidth]{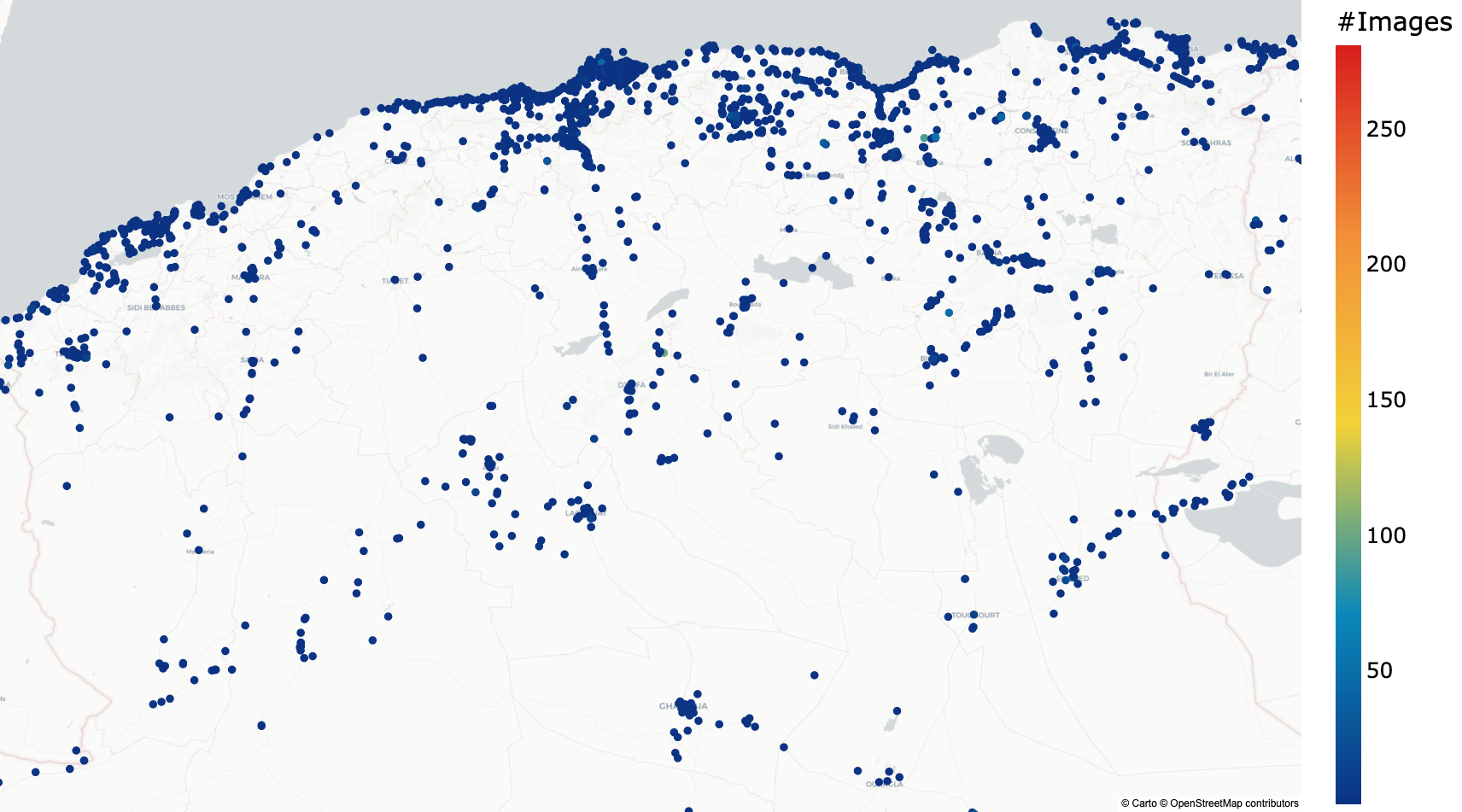} \label{fig:Algeria-images}}
\hfill
\subfloat[Madagascar: RWI groups data distribution.]{\includegraphics[width=0.495\textwidth]{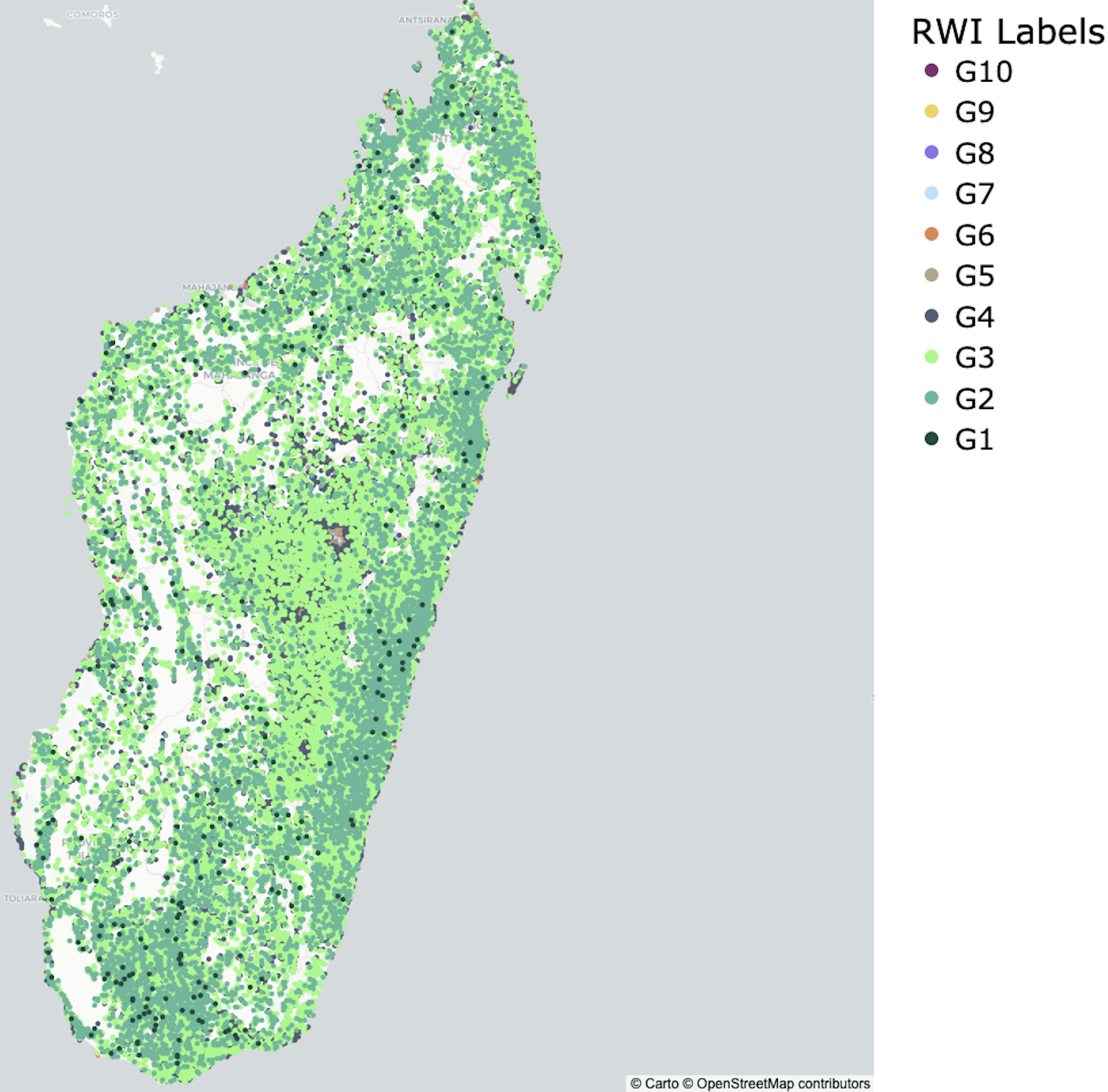} \label{fig:madagascar_rwi}}
\subfloat[Madagascar: Geoimages distribution.]{\includegraphics[width=0.47\textwidth]{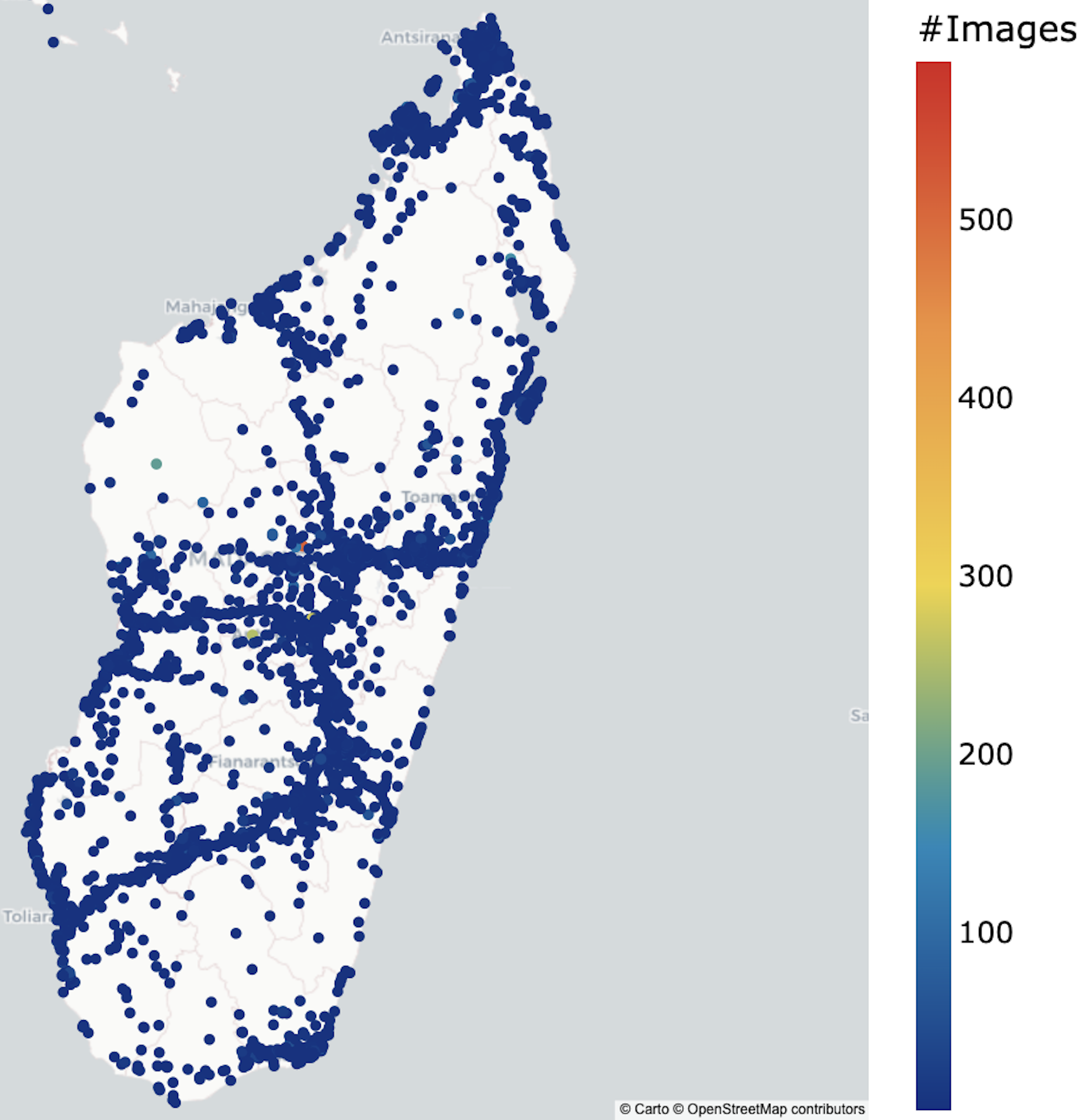} \label{fig:madagascar_images}}
\caption{Maps displaying relative wealth index (RWI) and geolocation of Flickr Africa images via ``query-by-name'' for (\textbf{a, b}) Algeria and (\textbf{c, d}) Madagascar.   (\textbf{a, c}) display the distribution of RWI regions as colored by RWI group. Here, RWI groups are equally binned into G1-G10 by nation, with G1 having the lowest RWI values.
(\textbf{b, d}) display the number of images sourced from each RWI region as colored by number of images per region. Tolerance distances from geotag to nearest RWI-labeled point are $\leq 10km$.}
\label{fig:rwi-images-algeria-madagascar}
\end{figure*}
\raggedbottom

\begin{table*}[ht!]
\footnotesize
\caption{Number of geotagged images, number of total images, and percentage of geotagged images for for each African nation as ordered by number of geotagged images.}
\begin{tabularx}{\textwidth}{@{} *{2}{C} @{}}
    \label{table:geo_images}
     \begin{tabular}{@{} llll @{}}
        \toprule
        \multirow{2.3}{*}{}
        & \mcc[3]{``query-by-name''}          \\
         \midrule
                 Country &  \#geotagged &  \#all &  \%geotagged \\
        \midrule
            South Africa &       128294 &       513082 &       25.004580 \\
                   Egypt &       105996 &       491913 &       21.547713 \\
                 Morocco &        93483 &       473673 &       19.735767 \\
                Tanzania &        83571 &       420362 &       19.880722 \\
                 Namibia &        71697 &       337390 &       21.250482 \\
                   Kenya &        70450 &       464611 &       15.163223 \\
              Madagascar &        57427 &       278207 &       20.641824 \\
                Ethiopia &        55099 &       329288 &       16.732769 \\
                Botswana &        53041 &       245312 &       21.621853 \\
                 Tunisia &        50899 &       231289 &       22.006667 \\
                  Guinea &        47602 &       268428 &       17.733619 \\
               Mauritius &        43953 &       210857 &       20.844933 \\
                  Uganda &        40828 &       317371 &       12.864439 \\
                Zimbabwe &        37752 &       172689 &       21.861265 \\
                    Chad &        35256 &       295114 &       11.946570 \\
                   Ghana &        35190 &       241065 &       14.597723 \\
                 Nigeria &        34166 &       213339 &       16.014887 \\
                 Senegal &        34123 &       228327 &       14.944794 \\
                   Sudan &        28627 &       163649 &       17.492927 \\
                  Zambia &        28173 &       194686 &       14.470994 \\
                 Algeria &        25849 &       105254 &       24.558687 \\
              Mozambique &        25139 &       128100 &       19.624512 \\
                    Mali &        24355 &       167186 &       14.567607 \\
              Seychelles &        21213 &       100623 &       21.081661 \\
                   Libya &        18156 &        86833 &       20.909101 \\
                  Malawi &        18038 &       133032 &       13.559144 \\
                  Angola &        17881 &        99342 &       17.999436 \\
                   Niger &        17506 &        96030 &       18.229720 \\
                  Rwanda &        17443 &       250469 &        6.964135 \\
                  Gambia &        15405 &        96972 &       15.886029 \\
                   Benin &        12884 &        97214 &       13.253235 \\
                Cameroon &        12023 &        72954 &       16.480248 \\
            Burkina Faso &        11939 &        60983 &       19.577587 \\
              Cape Verde &        11465 &        43283 &       26.488460 \\
                 Somalia &        11296 &       101989 &       11.075704 \\
                    Togo &        11054 &        65710 &       16.822401 \\
                 Liberia &        10724 &        68774 &       15.593102 \\
             South Sudan &         9380 &        60345 &       15.543956 \\
               Swaziland &         8699 &        75631 &       11.501897 \\
       Republic of Congo &         7581 &        33438 &       22.671811 \\
            Sierra Leone &         7303 &        52530 &       13.902532 \\
                   Gabon &         5963 &        39510 &       15.092382 \\
                     DRC &         5614 &        21560 &       26.038961 \\
                 Eritrea &         5221 &        26731 &       19.531630 \\
                Djibouti &         5179 &        36029 &       14.374532 \\
              Mauritania &         5126 &        29616 &       17.308212 \\
                 Lesotho &         5121 &        43282 &       11.831708 \\
             Ivory Coast &         4549 &        30743 &       14.796864 \\
                 Burundi &         3113 &        39888 &        7.804352 \\
                     CAF &         2954 &        19901 &       14.843475 \\
           Guinea-Bissau &         1614 &         8089 &       19.953023 \\
                 Comoros &         1470 &         6961 &       21.117656 \\
       Equatorial Guinea &         1293 &         8614 &       15.010448 \\
   Sao Tome and Principe &          776 &         3904 &       19.877049 \\
        \bottomrule
     \end{tabular}    &  \label{table:geo_images_people}
          \begin{tabular}{@{} llll @{}}
        \toprule
        \multirow{2.3}{*}{}
        & \mcc[3]{``query-by-name+people''}          \\
         \midrule
                 Country &  \#geotagged &  \#all &  \%geotagged \\
        \midrule
            South Africa &        46021 &       203871 &       22.573588 \\
                   Egypt &        32869 &       176992 &       18.570896 \\
                 Morocco &        23276 &       127263 &       18.289684 \\
                Ethiopia &        21421 &       107035 &       20.013080 \\
                   Kenya &        16511 &       135069 &       12.224122 \\
                Tanzania &        12829 &        77061 &       16.647850 \\
                 Nigeria &        11757 &        62525 &       18.803679 \\
                    Chad &         9081 &        90915 &        9.988451 \\
                   Ghana &         8708 &        62570 &       13.917213 \\
                  Uganda &         8687 &        77206 &       11.251716 \\
                  Guinea &         7576 &        43333 &       17.483211 \\
                   Sudan &         7545 &        43339 &       17.409262 \\
              Madagascar &         7412 &        42962 &       17.252456 \\
                 Tunisia &         6498 &        33732 &       19.263607 \\
                Zimbabwe &         6435 &        28614 &       22.488991 \\
               Mauritius &         6412 &        28053 &       22.856735 \\
                 Namibia &         6143 &        36366 &       16.892152 \\
                 Senegal &         5237 &        43136 &       12.140671 \\
                    Mali &         5142 &        36059 &       14.259963 \\
              Mozambique &         4231 &        24365 &       17.365073 \\
            Burkina Faso &         3691 &        12990 &       28.414165 \\
                  Rwanda &         3461 &        76521 &        4.522941 \\
                   Benin &         3311 &        20235 &       16.362738 \\
             South Sudan &         3269 &        20153 &       16.220910 \\
                   Libya &         3176 &        15737 &       20.181737 \\
                  Zambia &         3103 &        33063 &        9.385113 \\
                 Somalia &         3091 &        26796 &       11.535304 \\
                 Algeria &         2951 &        15587 &       18.932444 \\
                  Malawi &         2791 &        26139 &       10.677532 \\
                  Angola &         2508 &        19257 &       13.023835 \\
       Republic of Congo &         2110 &         8249 &       25.578858 \\
                    Togo &         1966 &        12806 &       15.352179 \\
                   Niger &         1859 &        14139 &       13.148030 \\
                     DRC &         1850 &         7028 &       26.323278 \\
                 Liberia &         1718 &        14388 &       11.940506 \\
                  Gambia &         1612 &        13806 &       11.676083 \\
                Cameroon &         1609 &        14100 &       11.411348 \\
            Sierra Leone &         1564 &        13410 &       11.662938 \\
               Swaziland &         1420 &        12425 &       11.428571 \\
             Ivory Coast &         1353 &         6095 &       22.198523 \\
              Cape Verde &         1333 &         5390 &       24.730983 \\
              Seychelles &         1225 &         8138 &       15.052839 \\
                Botswana &         1085 &         4664 &       23.263293 \\
                 Eritrea &         1074 &         5981 &       17.956863 \\
                     CAF &          882 &         7108 &       12.408554 \\
                Djibouti &          784 &         6328 &       12.389381 \\
                 Lesotho &          642 &         5971 &       10.751968 \\
              Mauritania &          590 &         4019 &       14.680269 \\
                 Burundi &          533 &         9862 &        5.404583 \\
                   Gabon &          476 &         7286 &        6.533077 \\
                 Comoros &          214 &         1173 &       18.243819 \\
           Guinea-Bissau &          196 &         1391 &       14.090582 \\
       Equatorial Guinea &          168 &         2081 &        8.073042 \\
   Sao Tome and Principe &          116 &          912 &       12.719298 \\
        \bottomrule
     \end{tabular}
\end{tabularx}
\label{table:all_geotag_countries}
\end{table*}
\raggedbottom

\clearpage
\subsection{Analysis of Flickr user-defined tags for geotagged images}


\begin{figure*}[ht!]
\centering
\begin{subfigure}{0.44\textwidth}
    \includegraphics[width=\textwidth]{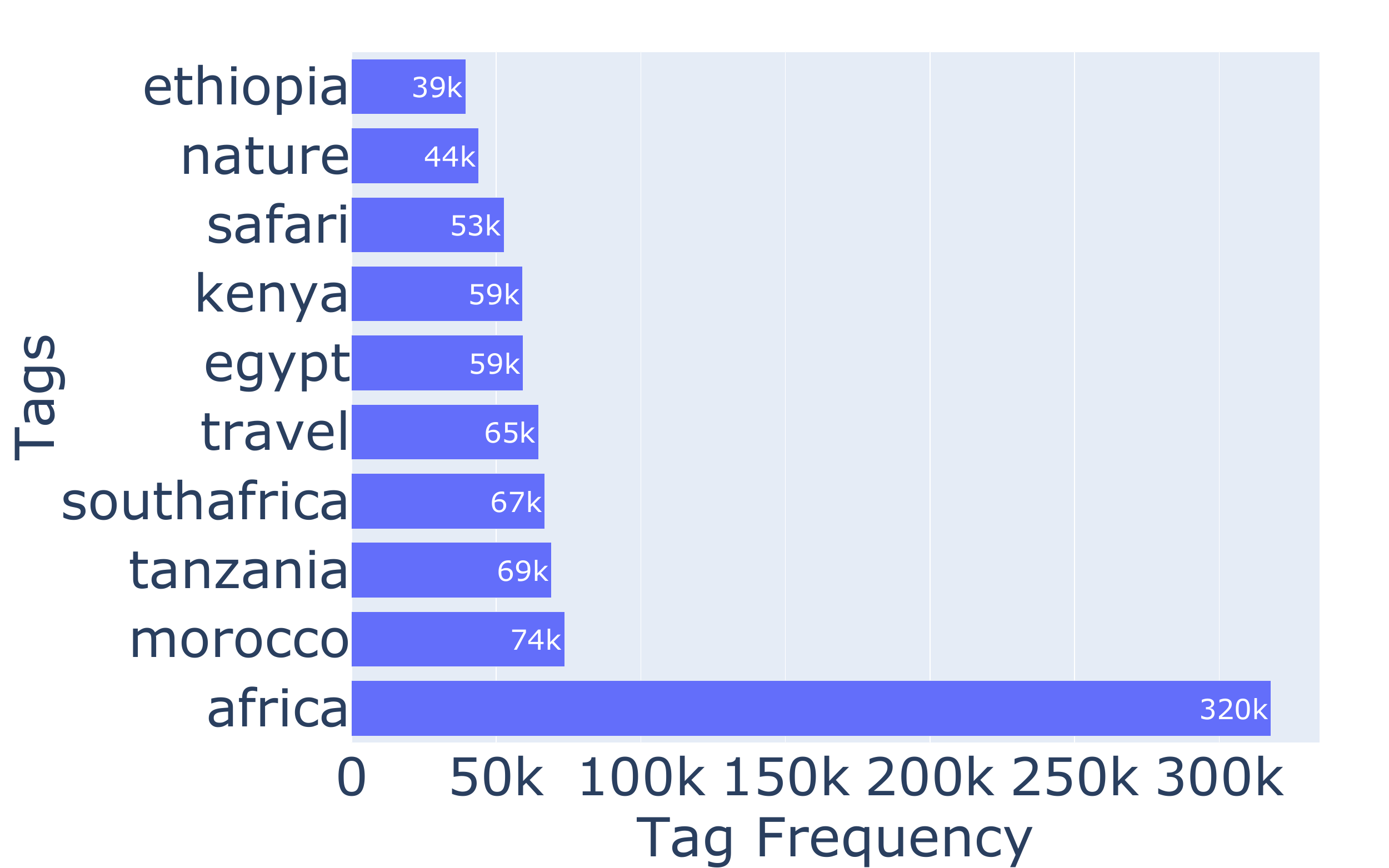}
    \caption{Most common tags;  \\     
    ``query-by-name''.}
    \label{fig:toptags_bigdata_name}
\end{subfigure}
\hfill
\begin{subfigure}{0.44\textwidth}
    \includegraphics[width=\textwidth]{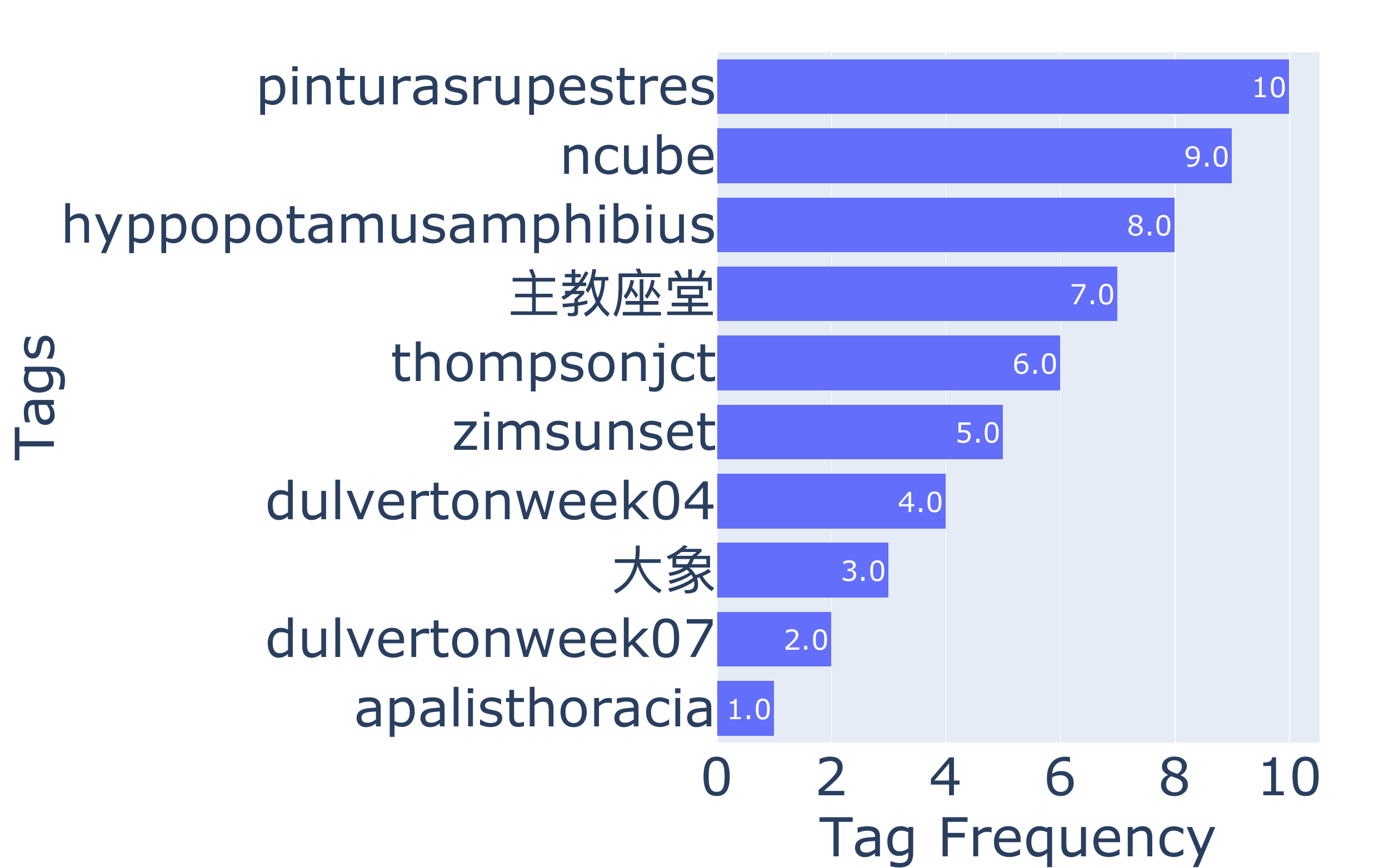}
    \caption{Least common tags;   \\   
    ``query-by-name''.}
    \label{fig:bottags_bigdata_name}
\end{subfigure}
\hfill
\begin{subfigure}{0.44\textwidth}
    \includegraphics[width=\textwidth]{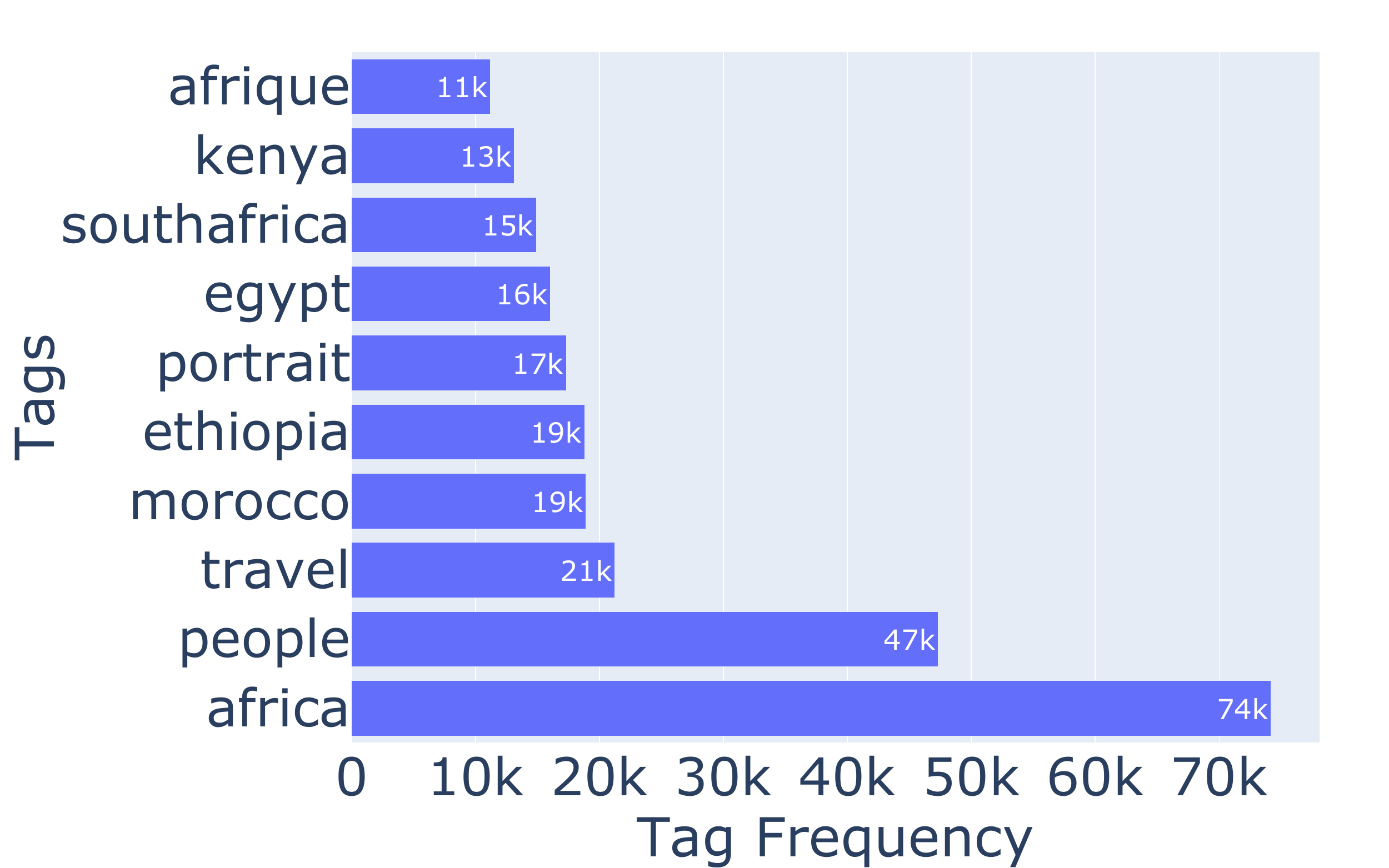}
    \caption{Most common tags;   \\   
    ``query-by-name+people''.}
    \label{fig:toptags_bigdata_people}
\end{subfigure}
\hfill
\begin{subfigure}{0.44\textwidth}
    \includegraphics[width=\textwidth]{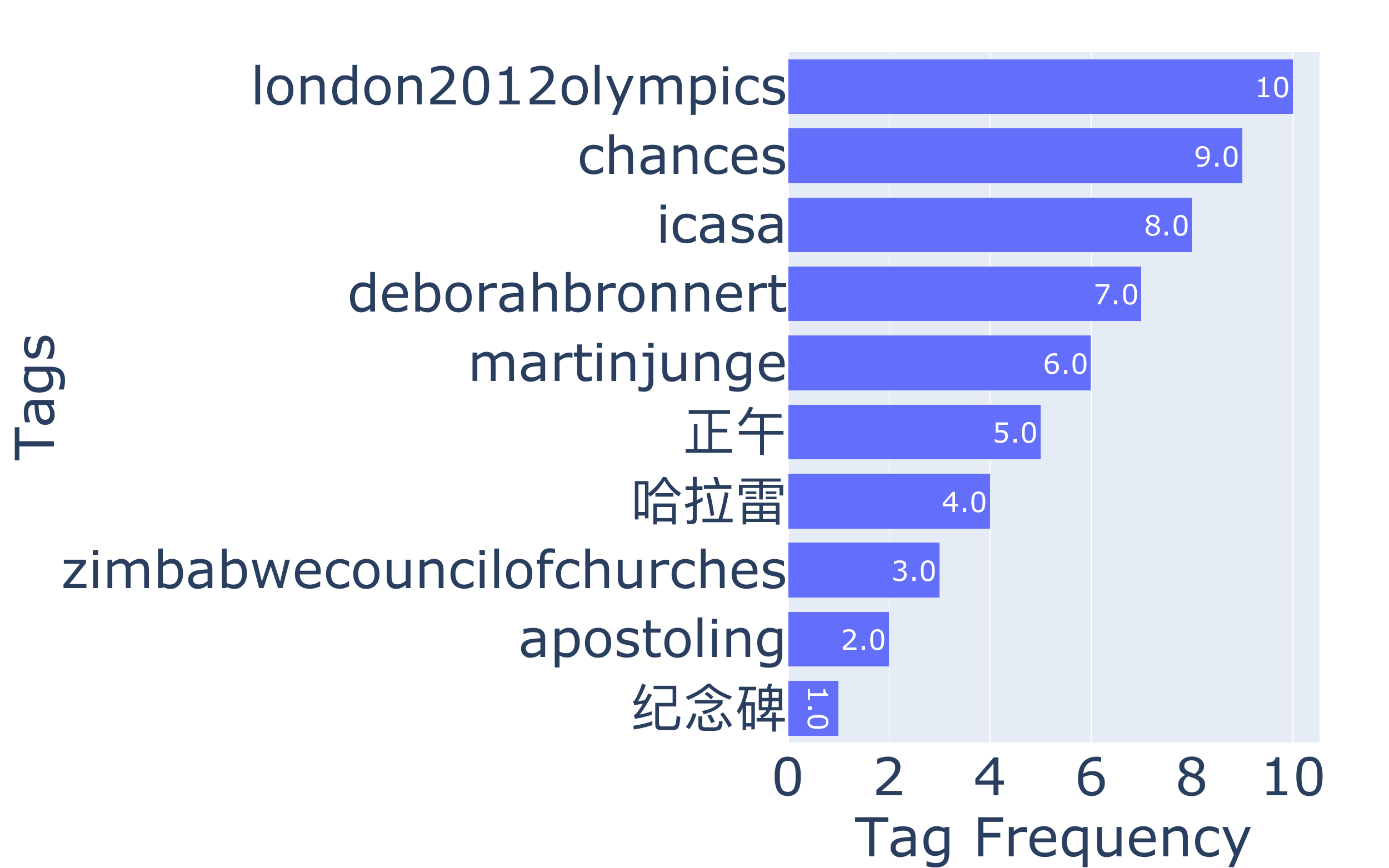}
    \caption{Least common tags; \\    
    ``query-by-name+people''.}
    \label{fig:bottags_bigdata_people}
\end{subfigure}
\caption{{For images sourced from Africa, bar plots show tag frequency of the most and least commonly returned tags. Only those images sufficiently close to the nearest RWI-labeled address (within $10km$) are analyzed for each query type.}}
\label{fig:tags_bigdata}
\end{figure*}
\raggedbottom

\clearpage
\subsection{Country and city tag distribution for geotagged images}

\begin{table*}[ht!]
\footnotesize
\captionof{table}{Number of different country locations (by geotag) associated with images queried by each country name, plus the top country location associated with each query-by-name, and the associated count/percentage of top result by geotag. For queries like ``Chad'', the most common geotag location does not match the country name query; in this case, because ``Chad'' is also a male name in the United States and elsewhere.}
\label{table:top_countries_name}
\begin{tabular}{lllll}
\toprule
Query country & \#Countries returned &                        Top country & Frequency & Percentage \\
\midrule
    Tanzania &        104 &          Tanzania, United Republic of &                 77613 &                92.8707 \\
  Seychelles &         86 &                            Seychelles &                 19077 &                89.9307 \\
  Cape Verde &         74 &                            Cape Verde &                 10293 &                89.7776 \\
     Tunisia &        103 &                               Tunisia &                 45688 &                89.7621 \\
     Morocco &        120 &                               Morocco &                 82983 &                 88.768 \\
   Mauritius &         89 &                             Mauritius &                 38851 &                88.3921 \\
     Namibia &         96 &                               Namibia &                 63312 &                88.3049 \\
      Gambia &         80 &                                Gambia &                 13596 &                88.2571 \\
Sao Tome and Principe &         22 &                 Sao Tome and Principe &                   674 &                86.8557 \\
      Uganda &        122 &                                Uganda &                 34326 &                 84.087 \\
    Botswana &         74 &                              Botswana &                 43371 &                81.7688 \\
  Madagascar &        155 &                            Madagascar &                 45127 &                78.5815 \\
    Djibouti &         64 &                              Djibouti &                  4021 &                77.6405 \\
      Rwanda &         84 &                                Rwanda &                 13459 &                77.1599 \\
      Malawi &         74 &                                Malawi &                 13913 &                77.1316 \\
       Kenya &        128 &                                 Kenya &                 53006 &                75.2402 \\
  Mauritania &         73 &                            Mauritania &                  3830 &                74.7171 \\
   Swaziland &         49 &                             Swaziland &                  6495 &                74.6638 \\
    Ethiopia &        134 &                              Ethiopia &                 41070 &                74.5386 \\
Burkina Faso &         74 &                          Burkina Faso &                  8881 &                74.3865 \\
     Senegal &        127 &                               Senegal &                 24800 &                72.6783 \\
       Ghana &        122 &                                 Ghana &                 25316 &                71.9429 \\
Guinea-Bissau &         40 &                         Guinea-Bissau &                  1159 &                71.8092 \\
  Mozambique &         93 &                            Mozambique &                 17489 &                69.5692 \\
South Africa &        177 &                          South Africa &                 88677 &                69.1201 \\
Sierra Leone &         64 &                          Sierra Leone &                  4934 &                67.5613 \\
    Cameroon &        104 &                              Cameroon &                  8112 &                67.4707 \\
     Comoros &         47 &                               Comoros &                   966 &                65.7143 \\
     Algeria &        102 &                               Algeria &                 16744 &                64.7762 \\
       Egypt &        151 &                                 Egypt &                 68311 &                64.4492 \\
     Burundi &         65 &                               Burundi &                  1927 &                61.9017 \\
       Gabon &         86 &                                 Gabon &                  3688 &                61.8481 \\
    Zimbabwe &         96 &                              Zimbabwe &                 22461 &                59.4962 \\
       Benin &         90 &                                 Benin &                  7631 &                59.2285 \\
     Lesotho &         42 &                               Lesotho &                  3021 &                58.9924 \\
     Nigeria &        133 &                               Nigeria &                 20149 &                58.9738 \\
      Zambia &         87 &                                Zambia &                 16612 &                58.9643 \\
       Libya &        100 &                Libyan Arab Jamahiriya &                 10553 &                58.1272 \\
     Eritrea &         82 &                               Eritrea &                  2920 &                 55.928 \\
        Togo &         94 &                                  Togo &                  6083 &                55.0299 \\
        Mali &        129 &                                  Mali &                 13352 &                54.8562 \\
        \hl{Chad} &        129 &                        \hl{ United States} &                 19260 &                 54.629 \\
 Ivory Coast &         78 &                         Cote d'Ivoire &                  2450 &                 53.858 \\
 South Sudan &         83 &                           South Sudan &                  4864 &                 51.855 \\
       Sudan &        124 &                                 Sudan &                 14010 &                48.9398 \\
     Liberia &        110 &                               Liberia &                  5086 &                47.4396 \\
Democratic Republic of the Congo &         86 & Congo, The Democratic Republic of the &                  2412 &                 42.964 \\
Equatorial Guinea &         54 &                     Equatorial Guinea &                   553 &                42.7688 \\
\hl{Republic of Congo }&         93 & \hl{Congo, The Democratic Republic of the} &                  3195 &                42.1448 \\
      Angola &        117 &                                Angola &                  7487 &                41.8713 \\
Central African Republic &         86 &              Central African Republic &                  1125 &                 38.084 \\
     Somalia &        113 &                               Somalia &                  3840 &                33.9943 \\
      \hl{Guinea} &        190 &                      \hl{Papua New Guinea} &                 14267 &                29.9714 \\
       Niger &        147 &                                 Niger &                  3517 &                20.0903 \\
\bottomrule
\end{tabular}
\label{table:number_query_country_freq_per}
\end{table*}


\begin{figure*}[!ht]
\centering
\begin{subfigure}{0.495\textwidth}
    \includegraphics[width=1.1\textwidth]{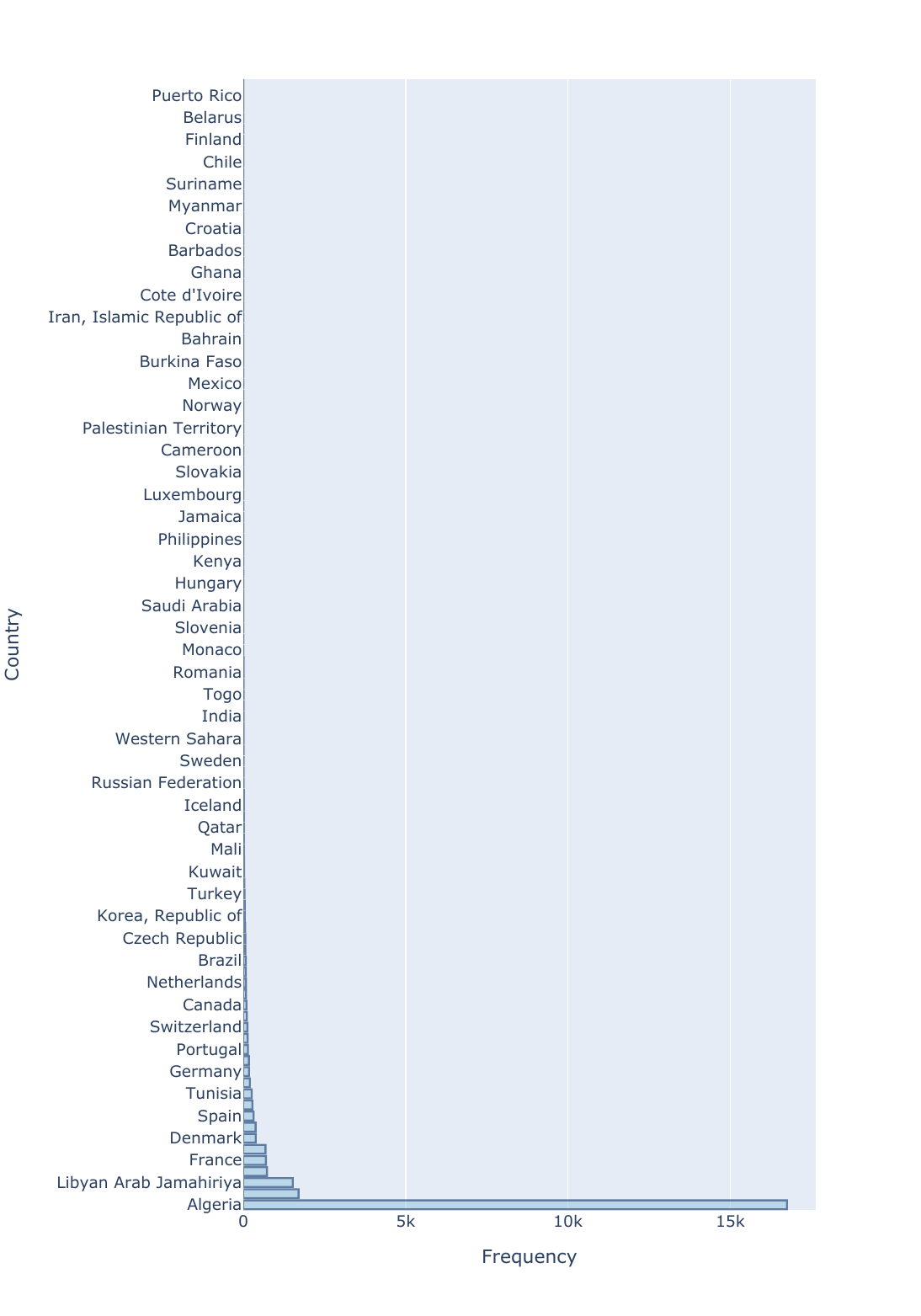}
    \caption{query by Algeria}
    \label{fig:algeria_countries_freq}
\end{subfigure}
\hfill
\begin{subfigure}{0.4955\textwidth}
    \includegraphics[width=1.1\textwidth]{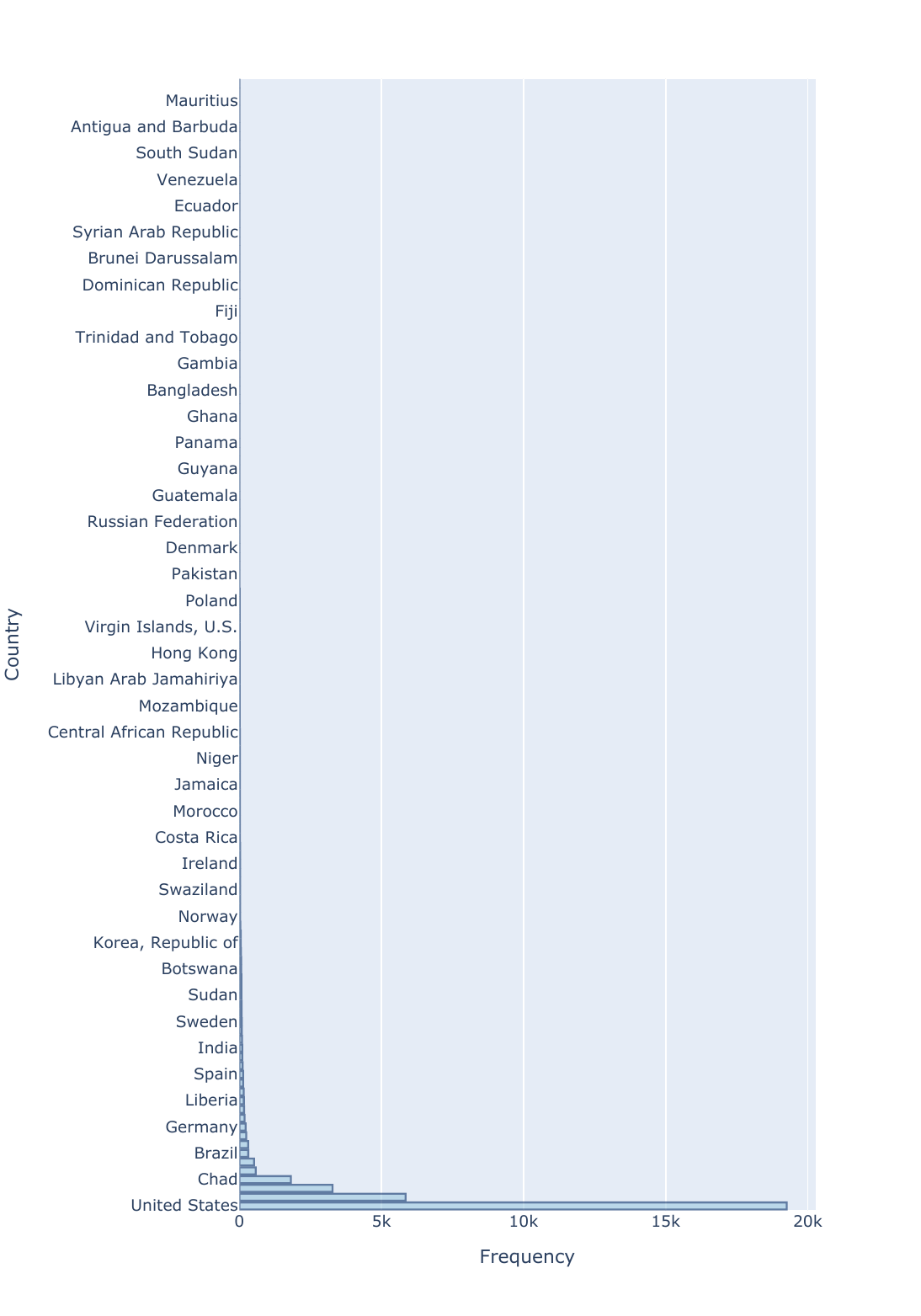}
    \caption{query by Chad}
    \label{fig:chad_countries_freq}
\end{subfigure}
\caption{The count for each geotag location for image data resulting from Flickr queries of (\textbf{a}) Algeria and (\textbf{b}) Madagascar, without exclusion of images with nearest RWI-labeled address beyond threshold. Queries by country names such as Algeria and Chad return more image data obtained from mainly Algeria and the United States respectively, according to geotag location.}
\label{fig:countries_dist}
\end{figure*}




\begin{table*}[ht!]
\footnotesize
\captionof{table}{Number of cities returned by the  ``query-by-name+people'' according to geotag location,  and the top city, its frequencies and percentage with respect to other returned cities in the geotags for each nation.}
\label{table:top_cities_name}
\begin{tabular}{llllll}
\toprule
Query country & \#Cities returned &                        Top city & Frequency & Percentage & Capital\\
\midrule
Sao Tome and Principe &      22 &        São Tomé &                475 &             70.4748 &                 São Tomé \\
                         Eritrea &      82 &          Asmara &               1750 &             59.9315 &                   Asmara \\
                        Djibouti &      64 &        Djibouti &               2326 &             57.8463 &                 Djibouti \\
                         Liberia &     110 &        Monrovia &               2873 &             56.4884 &                 Monrovia \\
                      Seychelles &      86 &        \hl{La Passe} &              10688 &             56.0256 &                 \hl{Victoria} \\
                         Burundi &      65 &       Bujumbura &                914 &             47.4312 &                Bujumbura \\
                        Zimbabwe &      96 &  \hl{Victoria Falls} &               9859 &             43.8939 &                   \hl{Harare} \\
               Equatorial Guinea &      54 &          Malabo &                238 &              43.038 &                   Malabo \\
                          Angola &     117 &          Luanda &               3059 &             40.8575 &                   Luanda \\
                           Gabon &      86 &      Libreville &               1480 &             40.1302 &               Libreville \\
        Central African Republic &      86 &          Bangui &                427 &             37.9556 &                   Bangui \\
                          Rwanda &      84 &         \hl{Musanze} &               4993 &             37.0979 &                   \hl{Kigali} \\
                      Mozambique &      93 &          Maputo &               6481 &             37.0576 &                   Maputo \\
               Republic of Congo &      93 &          \hl{Makoua} &                180 &             36.2903 &              \hl{Brazzaville} \\
                         Lesotho &      42 &            \hl{Nako} &               1045 &             34.5912 &                   \hl{Maseru} \\
                            Togo &      94 &            Lomé &               2007 &             32.9936 &                     Lomé \\
                        Botswana &      74 &          \hl{Kasane} &              14262 &             32.8837 &                 \hl{Gaborone} \\
                           Niger &     147 &          Niamey &               1127 &             32.0444 &                   Niamey \\
                    Burkina Faso &      74 &     Ouagadougou &               2619 &             29.4899 &              Ouagadougou \\
                           Libya &     100 &         Tripoli &               2924 &             27.7078 &                  Tripoli \\
                         Morocco &     120 &       \hl{Marrakesh} &              21581 &             26.0065 &                    \hl{Rabat} \\
                         Somalia &     113 &       Mogadishu &                983 &              25.599 &                Mogadishu \\
                       Swaziland &      49 &         Lobamba &               1655 &             25.4811 &      Mbabane and Lobamba \\
                   Guinea-Bissau &      40 &         \hl{Bubaqu}e &                291 &             25.1079 &                   \hl{Bissau} \\
                      Cape Verde &      74 & \hl{Vila de Sal Rei} &               2356 &             22.8893 &                    \hl{Praia} \\
                          Guinea &     190 &         Conakry &                621 &                21.6 &                  Conakry \\
                           Egypt &     151 &           Cairo &              14531 &             21.2718 &                    Cairo \\
                          Gambia &      80 &          \hl{Sukuta} &               2887 &             21.2342 &                   \hl{Banjul} \\
                     Ivory Coast &      78 &         Abidjan &                503 &             20.5306 & Abidjan and Yamoussoukro \\
                           Ghana &     122 &           Accra &               5114 &             20.2007 &                    Accra \\
                       Mauritius &      89 &   \hl{Quatre Bornes} &               7657 &             19.7086 &               \hl{Port Louis} \\
                          Zambia &      87 &         \hl{Chipata} &               3270 &             19.6846 &                   \hl{Lusaka} \\
                      Mauritania &      73 &      Nouadhibou &                707 &             18.4595 &               Nouakchott \\
                            Mali &     129 &          Bamako &               2392 &             17.9149 &                   Bamako \\
                         Nigeria &     133 &           \hl{Ikoyi} &               3583 &             17.7825 &                    \hl{Abuja} \\
                            Chad &     129 &    \hl{Faya-Largeau} &                321 &              17.725 &                \hl{N'Djamena} \\
                           Benin &      90 &         \hl{Cotonou} &               1334 &             17.4813 &               \hl{Porto-Novo} \\
                         Algeria &     102 &         Algiers &               2909 &             17.3734 &                  Algiers \\
                        Tanzania &     104 &      \hl{Ngorongoro} &              13458 &             17.3399 &            \hl{Dar es Salaam} \\
                         Senegal &     127 &           Dakar &               4291 &             17.3024 &                    Dakar \\
                         Namibia &      96 &       \hl{Maltah\"{\o}he} &               9938 &             15.6969 &                 \hl{Windhoek} \\
Democratic Republic of the Congo &      86 &        \hl{Yangambi} &                378 &             15.6716 &                 \hl{Kinshasa} \\
                        Ethiopia &     134 &     Addis Ababa &               6274 &             15.2764 &              Addis Ababa \\
                           Kenya &     128 &         Nairobi &               7893 &             14.8908 &                  Nairobi \\
                         Tunisia &     103 &           Tunis &               6465 &             14.1503 &                    Tunis \\
                          Malawi &      74 &        Lilongwe &               1942 &             13.9582 &                 Lilongwe \\
                         Comoros &      47 &          Moroni &                124 &             12.8364 &                   Moroni \\
                    South Africa &     177 &       Cape Town &               9451 &             10.6578 &   Cape Town and Pretoria \\
                        Cameroon &     104 &         Yaoundé &                766 &              9.4428 &                  Yaoundé \\
                      Madagascar &     155 &      \hl{Hell-Ville} &               4136 &              9.1652 &             \hl{Antananarivo} \\
                          Uganda &     122 &         \hl{Entebbe} &               2501 &               7.286 &                  \hl{Kampala} \\
\bottomrule
\end{tabular}
\label{table:num_cities_freq_per_capital}
\end{table*}


\begin{figure*}[!ht]
\centering
\begin{subfigure}{0.495\textwidth}
    \includegraphics[width=1.1\textwidth]{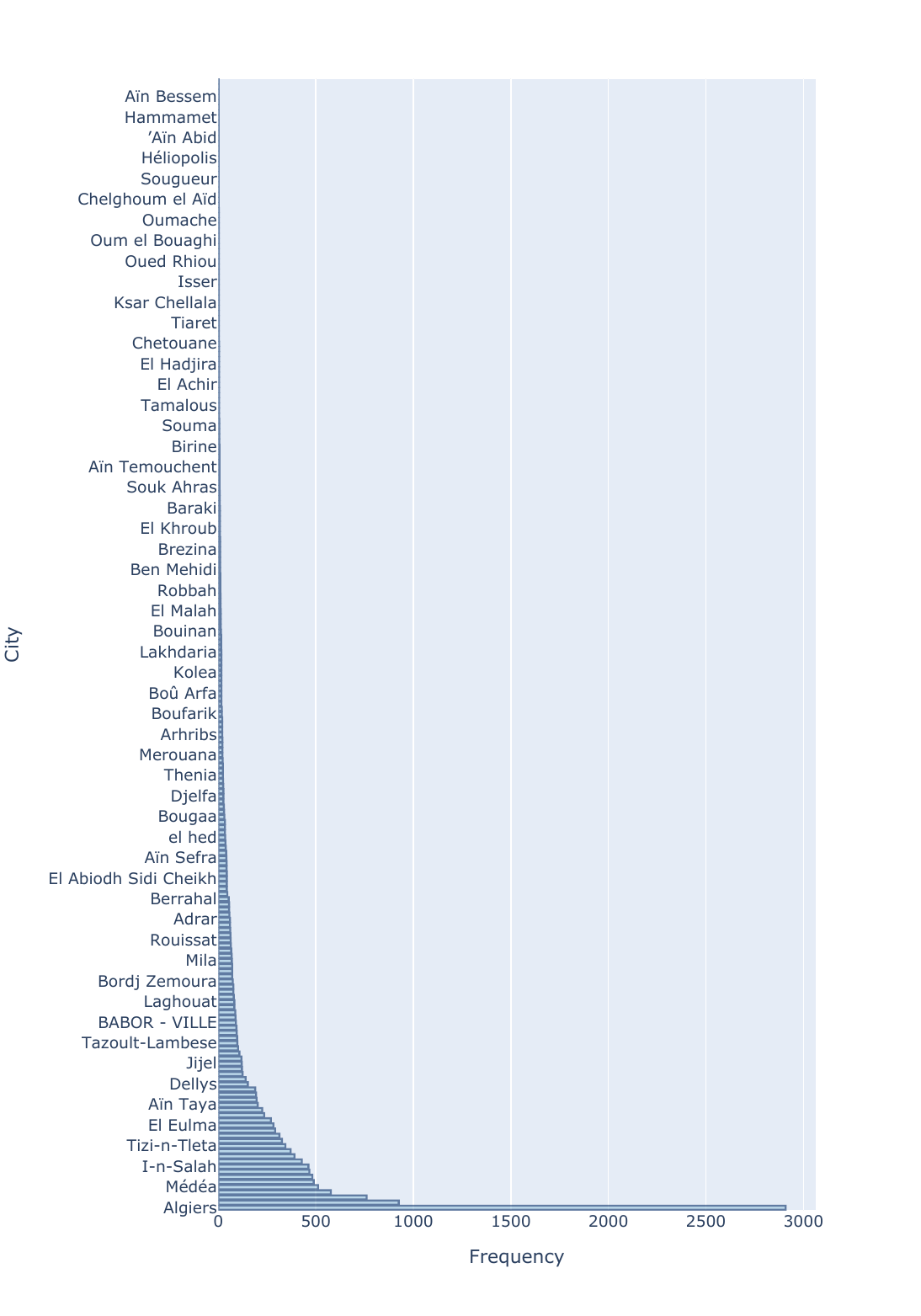}
    \caption{query by Algeria }
    \label{fig:algeria_cities_freq}
\end{subfigure}
\hfill
\begin{subfigure}{0.4955\textwidth}
    \includegraphics[width=1.1\textwidth]{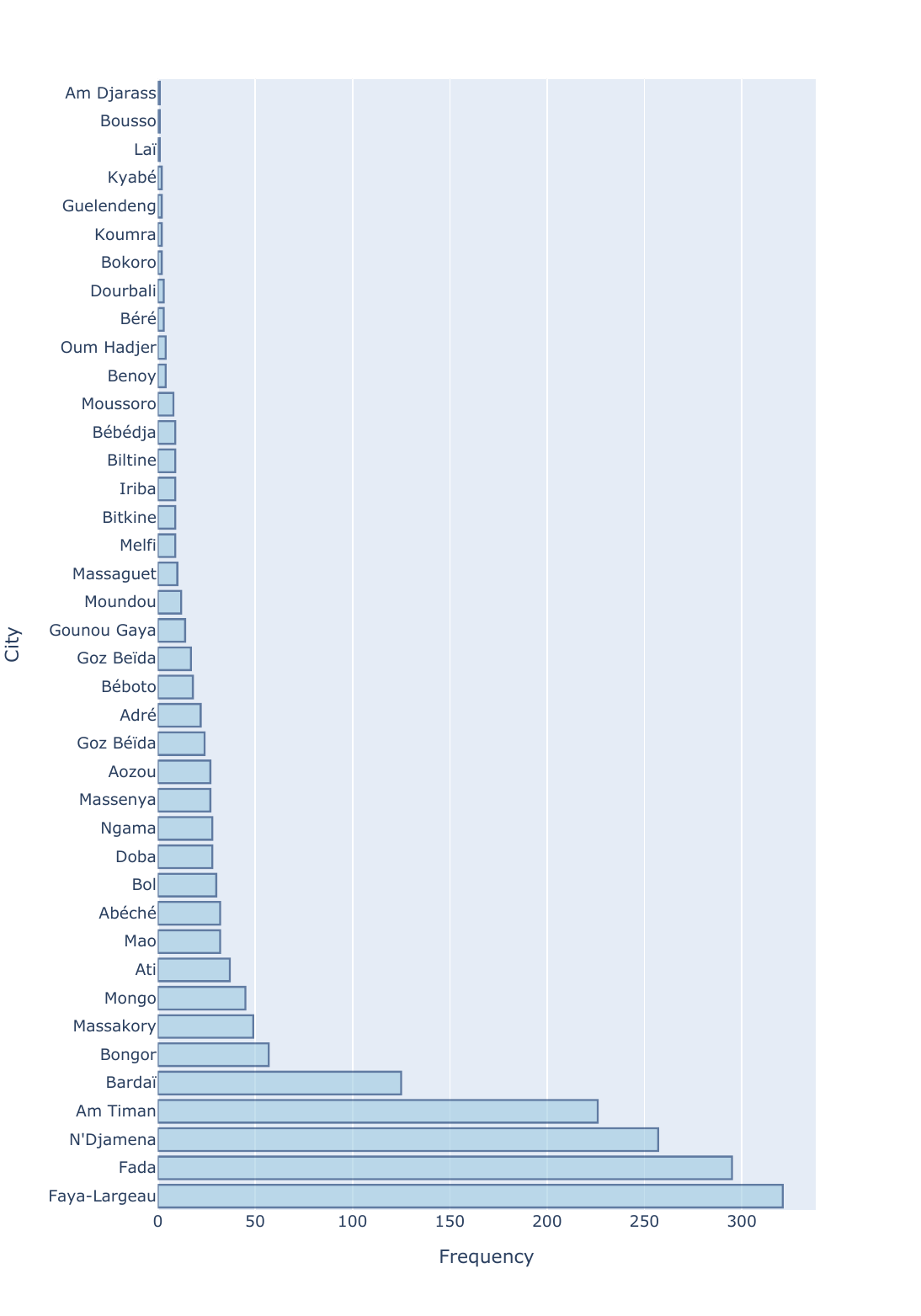}
    \caption{query by Chad}
    \label{fig:chad_cities_freq}
\end{subfigure}
\caption{The count for each city-based geotag location for image data resulting from Flickr queries of (\textbf{a}) Algeria and (\textbf{b}) Chad, without exclusion of images with nearest RWI-labeled address beyond threshold and with selection of only those images obtained from the respective countries according to geotag. Queries by Algeria and Chad return more image data geotagged to have been obtained from mainly the Algiers (capital city of Algeria) and Faya-Largeau (not capital city of Chad).}
\label{fig:cities_dist}
\end{figure*}


\subsection{Geotagged image distribution by RWI groups}

\begin{table*}[ht!]
\footnotesize
\caption{RWI lower, middle and upper grouping of RWI-labeled data for 49 African countries with search query ``query-by-name'', with data limited within $10km$. The RWI data is grouped in $10$ equal sized bins where, G1$<$G2$<$G3$<$G4$<$G5$<$G6$<$G7$<$G8$<$G9$<$G10.} 

\label{table:query_by_name_rwigroups}
\end{table*}


\begin{table*}[ht!]
\footnotesize
\caption{RWI lower, middle and upper grouping of RWI-labeled data  with search ``query-by-name+people'' for 49 African countries, with data limited within $10km$. The RWI data is grouped in $10$ equal sized bins where, G1$<$G2$<$G3$<$G4$<$G5$<$G6$<$G7$<$G8$<$G9$<$G10.} 
%
\label{table:query_by_name+people_rwigroups}
\end{table*}


\clearpage
\subsection{Temporal analysis of geotagged images}\label{sssec:temporal_a}

Geotagged images were assessed in approximately 2-year spans as labeled from A-I, from earlier to later date-time values. 
\begin{table}[!ht]
\centering
\begin{tabular}{cl}
    \textbf{A}: & $2004/2/10$, $00$::$00$::$00$ \textbf{---} $2006/2/9$, $23$::$59$::$59$ \\
    \textbf{B}: & $2006/2/10$, $00$::$00$::$00$ \textbf{---} $2008/2/10$, $23$::$59$::$59$ \\
    \textbf{C}: & $2008/2/11$, $00$::$00$::$00$ \textbf{---} $2010/2/9$, $23$::$59$::$59$ \\
    \textbf{D}: & $2010/2/10$, $00$::$00$::$00$ \textbf{---} $2012/2/10$, $23$::$59$::$59$ \\
    \textbf{E}: & $2012/2/11$, $00$::$00$::$00$ \textbf{---} $2014/2/9$, $23$::$59$::$59$ \\
    \textbf{F}: & $2014/2/10$, $00$::$00$::$00$ \textbf{---} $2016/2/10$, $23$::$59$::$59$ \\
    \textbf{G}: & $2016/2/11$, $00$::$00$::$00$ \textbf{---} $2018/2/9$, $23$::$59$::$59$ \\
    \textbf{H}: & $2018/2/10$, $00$::$00$::$00$ \textbf{---} $2020/2/10$, $23$::$59$::$59$ \\
    \textbf{I}: & $2020/2/11$, $00$::$00$::$00$ \textbf{---} $2022/2/10$, $23$::$59$::$59$ \\
\end{tabular}
\end{table}

For brevity, we report license type according to the numerical license IDs used by Flickr: 

\begin{table}[!ht]
\centering
\begin{tabular}{cl}
$\mathbf{0}$: & All Rights Reserved\\
$\mathbf{1}$: & Attribution-NonCommercial-ShareAlike License\\
$\mathbf{2}$: & Attribution-NonCommercial License\\
$\mathbf{3}$: & Attribution-NonCommercial-NoDerivs License\\
$\mathbf{4}$: & Attribution License\\
$\mathbf{5}$: & Attribution-ShareAlike License\\
$\mathbf{6}$: & Attribution-NoDerivs License\\
$\mathbf{7}$: & No known copyright restrictions\\
$\mathbf{8}$: & United States Government Work\\
$\mathbf{9}$: & Public Domain Dedication (CC0)\\
$\mathbf{10}$: & Public Domain Mark\\
\end{tabular}
\end{table}

\begin{table*}[ht!]
\footnotesize
\caption{Distribution of geotagged images by date group and nation with ``query-by-name''. Bold text indicates time interval with highest geotagged image count. } 
    \begin{tabular}{llllllllll}
    \toprule
    Country &    2004 - 2006   &   2006 - 2008   &   2008 - 2010   &   2010 - 2012   &   2012 - 2014   &   2014 - 2016   &   2016 - 2018   &   2018 - 2020   &   2020 - 2022\\   
    \midrule
         Algeria &   54 & 1258 & 1829 &  3417 &  2606 &  1376 &  \textbf{3639} &  2872 &   471 \\
          Angola &    2 &  793 & 1066 &  1526 &  1465 &  \textbf{2068} &  1087 &   940 &   688 \\
           Benin &  205 & 1231 & 1241 &  1212 &  1462 &  \textbf{1760} &   444 &   889 &   494 \\
        Botswana &  101 & 2523 & 1881 &  4680 &  4169 &  5244 &  4765 &  \textbf{5395} &  2043 \\
    Burkina Faso &   26 &  492 & \textbf{2514} &  1950 &  1886 &   661 &  1072 &   625 &   183 \\
         Burundi &    3 &  496 &  412 &   \textbf{567} &   485 &   293 &    99 &    67 &    25 \\
        Cameroon &   27 & 1216 & 1347 &  \textbf{1753} &  2688 &   595 &   625 &   213 &   342 \\
\end{tabular}
\end{table*}


\begin{table*}[ht!]
\footnotesize
\caption{Distribution of geotagged images by date group and nation with ``query-by-name+people''. Bold text indicates time interval with highest geotagged image count.} 
    \begin{tabular}{llllllllll}
    \toprule
     Country &    2004 - 2006   &   2006 - 2008   &   2008 - 2010   &   2010 - 2012   &   2012 - 2014   &   2014 - 2016   &   2016 - 2018   &   2018 - 2020   &   2020 - 2022\\   
    \midrule
             Algeria &   0 &   36 &   65 &  314 &  \textbf{508} &  156 &  116 &  185 &  17 \\
              Angola &   0 &   49 &   53 &  \textbf{278} &  186 &  209 &   44 &  168 &   9 \\
               Benin &  79 &  104 &  205 &  243 &  323 &  \textbf{706} &  113 &  128 &   3 \\
            Botswana &   2 &   15 &   94 &   52 &   51 &   87 &   54 &  \textbf{192} & 120 \\
        Burkina Faso &   0 &   44 &  318 &  514 &  \textbf{631} &  199 &  402 &  153 &   9 \\
             Burundi &   1 &   42 &   53 &  \textbf{104} &   71 &   57 &    8 &    9 &   2 \\
            Cameroon &   4 &  100 &  147 &  \textbf{285} &  198 &   96 &  245 &   35 &   9 \\
\end{tabular}
\end{table*}


\begin{table*}[ht!]
\footnotesize
\caption{Distribution of geotagged images by date group per nation and RWI group with ``query-by-name''. Undefined implies that there was no data in that time-range, therefore, no RWI group would be dominant. When two RWI groups are listed, this indicates that data from those groups was equal; i.e., no group dominates.} 
\begin{tabular}{llllllllll}
    \toprule
    Country &    2004 - 2006   &   2006 - 2008   &   2008 - 2010   &   2010 - 2012   &   2012 - 2014   &   2014 - 2016   &   2016 - 2018   &   2018 - 2020   &   2020 - 2022\\   
    \midrule
    Algeria &       upper &  upper & middle & middle &  upper &        upper &       middle &       middle &  upper \\
    Angola & lower,upper & middle & middle & middle & middle &        lower &       middle &       middle & middle \\
    Benin &       upper & middle & middle &  upper & middle &       middle &        upper &        upper & middle \\
    \hl{Botswana} &      \hl{middle} & \hl{middle} & \hl{middle} & \hl{middle} & \hl{middle} &      \hl{middle} &       \hl{middle} &       \hl{middle} & \hl{middle} \\
    Burkina Faso &      middle &  upper &  upper &  upper & middle &        lower &        upper &        upper & middle \\
    Burundi &       upper & middle & middle & middle & middle &       middle &       middle &       middle & middle \\
    Cameroon &       lower & middle &  upper & middle & middle &       middle &       middle &       middle & middle \\
\end{tabular}
\end{table*}


\begin{table*}[ht!]
\footnotesize
\caption{Distribution of geotagged images by date group per nation and RWI group with query by ``query-by-name+people''. Undefined implies that there was no data in that time-range, therefore, no RWI group would be dominant. When two RWI groups are listed, this indicates that data from those groups was equal; i.e., no group dominates.} 
\begin{tabular}{llllllllll}
    \toprule
    Country &    2004 - 2006   &   2006 - 2008   &   2008 - 2010   &   2010 - 2012   &   2012 - 2014   &   2014 - 2016   &   2016 - 2018   &   2018 - 2020   &   2020 - 2022\\   
    \midrule
        Algeria &    \textbf{undefined} &        upper &        upper & middle &       upper &       upper &        upper &       upper & \textbf{lower,middle} \\
        Angola &    \textbf{undefined} &       middle &        upper &  upper &       upper &      middle &       middle &      middle &       middle \\
        Benin &        upper &        upper &        upper &  lower &       upper &      middle &        upper & \textbf{lower,upper} &       middle \\
        Botswana &        upper &        lower &        upper & middle &       lower &       lower &        upper &      middle &        lower \\
        Burkina Faso &    \textbf{undefined} &        upper &        lower &  upper &       upper &       lower &       middle &       upper &       middle \\
        Burundi &        lower &        lower &       middle &  upper &      middle &      middle &       middle &      middle &       middle \\
        Cameroon &        lower &       middle &        lower &  lower &      middle &      middle &        lower &      middle &       middle \\
\end{tabular}
\end{table*}


\begin{table*}[ht!]
\footnotesize
\caption{Distribution of most common license group and number of geotagged images within that majority license group by date group per nation with ``query-by-name''. Refer to \Cref{sssec:temporal_a} for definition of license IDs.} 
\begin{tabular}{llllllllll}
    \toprule
    Country &    2004 - 2006   &   2006 - 2008   &   2008 - 2010   &   2010 - 2012   &   2012 - 2014   &   2014 - 2016   &   2016 - 2018   &   2018 - 2020   &   2020 - 2022\\   
    \midrule
Algeria &                               [0]:38 &  [0]:1077 &                             [0]:1617 &  [0]:2910 &  [0]:2220 & [0]:1270 & [0]:2679 & [0]:2441 &   [0]:401 \\
  Angola &                                [0]:2 &   [0]:724 &                              [0]:891 &  [0]:1351 &  [0]:1178 &  \textbf{[2]}:986 & [0]:1007 &  [0]:793 &   [0]:632 \\
   Benin &                              [0]:199 &  [0]:1087 &                              [0]:797 &  [0]:1055 &  [0]:1076 & [0]:1524 &  [0]:404 &  [0]:675 &   \textbf{[1]}:200 \\
Botswana &                               [0]:92 &  [0]:1451 &                             [0]:1648 &  [0]:3597 &  [0]:3471 & [0]:4245 & [0]:4042 & [0]:4587 &  [0]:1914 \\
Burkina Faso &                               \textbf{[1]}:24 &   [0]:383 &                             [0]:1656 &  [0]:1313 &  [0]:1709 &  [0]:579 &  [0]:934 &  [0]:534 &   [0]:136 \\
 Burundi &                                \textbf{[1]}:3 &   [0]:296 &                              [0]:380 &   [0]:446 &   [0]:477 &  [0]:217 &   [0]:95 &   [0]:60 &    [0]:25 \\
Cameroon &                               \textbf{[2]}:21 &   [0]:646 &                              [0]:813 &  [0]:1003 &  [0]:1482 &  [0]:374 &  [0]:373 &  [0]:151 &   [0]:158 \\
\end{tabular}
\end{table*}



\begin{table*}[ht!]
\footnotesize
\caption{Distribution of most common license group and number of geotagged images within that majority license group by date group per nation with ``query-by-name+people''. Refer to \Cref{sssec:temporal_a} for definition of license IDs. } 
\begin{tabular}{llllllllll}
    \toprule
    Country &    2004 - 2006   &   2006 - 2008   &   2008 - 2010   &   2010 - 2012   &   2012 - 2014   &   2014 - 2016   &   2016 - 2018   &   2018 - 2020   &   2020 - 2022\\   
    \midrule
Algeria & \textbf{[all]}:0 &                               [0]:29 &                               [0]:46 &  [0]:259 &                              [0]:459 &                              [0]:147 &                              [0]:103 &                              [0]:157 &                               [0]:16 \\
Angola & \textbf{[all]}:0 &                               [0]:32 &                               [0]:50 &  [0]:267 &                              [0]:156 &                              [0]:127 &                               [0]:41 &                              [0]:162 &                                [0]:9 \\
Benin &                               [0]:78 &                               [0]:88 &                              [0]:189 &  [0]:170 &                              [0]:306 &                              [0]:689 &                              [0]:108 &                               [0]:88 &                                [0]:3 \\
Botswana &                                [0]:2 &                               [0]:15 &                               \textbf{[2]}:83 &   [0]:34 &                               [0]:34 &                               [0]:52 &                               [0]:50 &                              [0]:191 &                              [0]:120 \\
Burkina Faso & \textbf{[all]}:0 &                               [0]:35 &                              [0]:253 &  [0]:367 &                              [0]:590 &                              [0]:186 &                              [0]:299 &                              [0]:115 &                                [0]:9 \\
\end{tabular}
\end{table*}


\begin{figure*}[!thbp]
\centering
\includegraphics[width=1\textwidth]{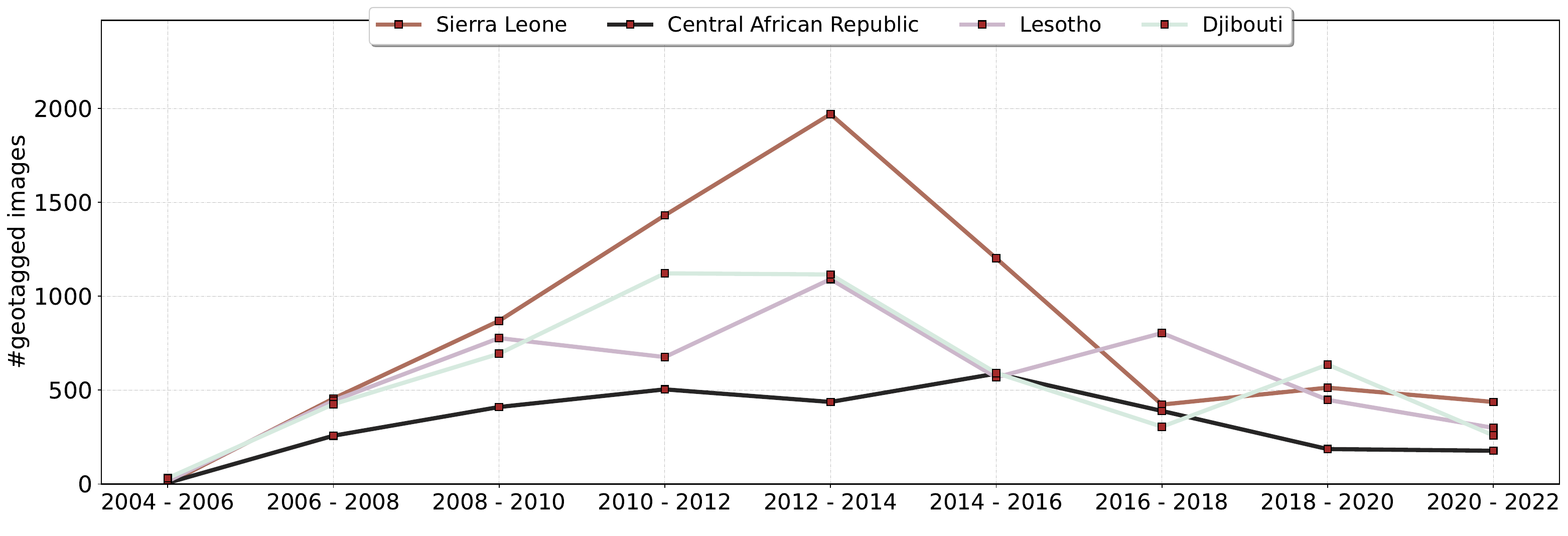}
\caption{The number of geotagged images from select African nations in approximately 2-year time ranges, as queried by country name. In general, a high number of images were uploaded in the dates in range 2012-2014, while fewer images were uploaded between dates in ranges 2004-2006 and 2020-2022.}
\label{fig:africa_images_temp}
\end{figure*}
\raggedbottom

\begin{figure*}[!thbp]
\centering
\includegraphics[width=0.9\textwidth]{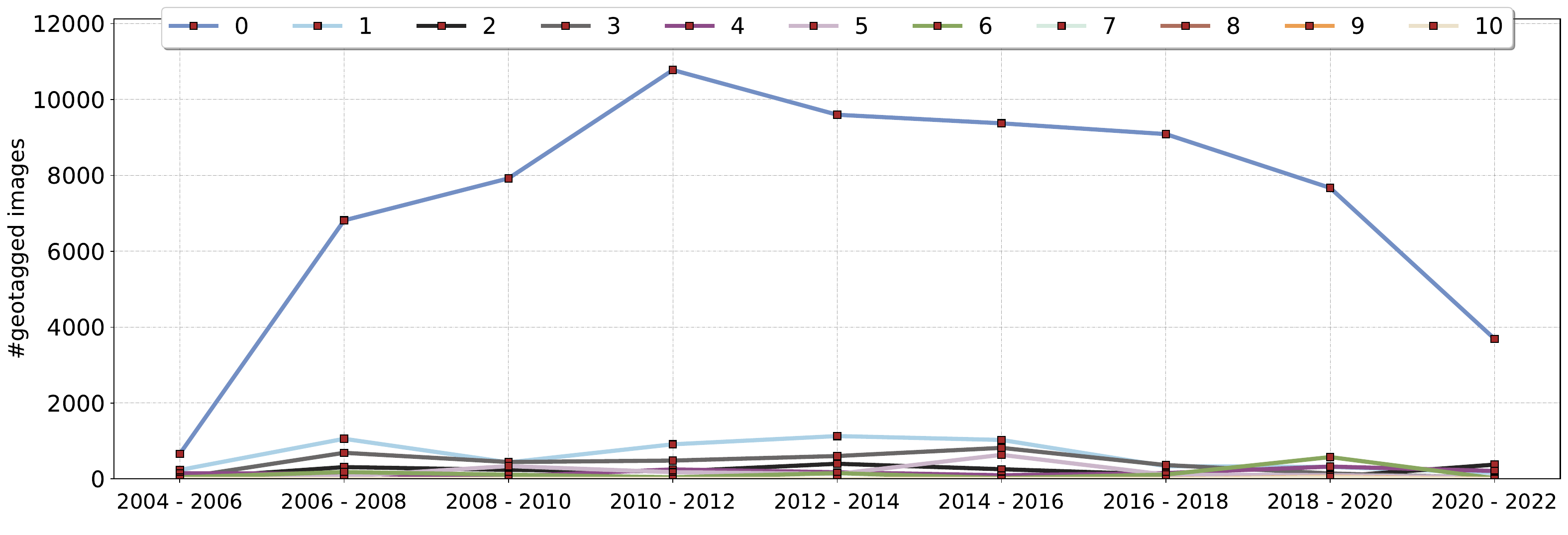}
\caption{Quantity of geotagged images over time with various Flickr license types, by Flickr numerical license IDs, for Morocco via query-by-name. The vast majority of queried images from Morocco were uploaded under the ``All Rights Reserved'' (ID: 0) license.}
\label{fig:morrocco_name_half_line_lic}
\end{figure*}
\raggedbottom

\begin{figure*}[!thbp]
\centering
\includegraphics[width=1\textwidth]{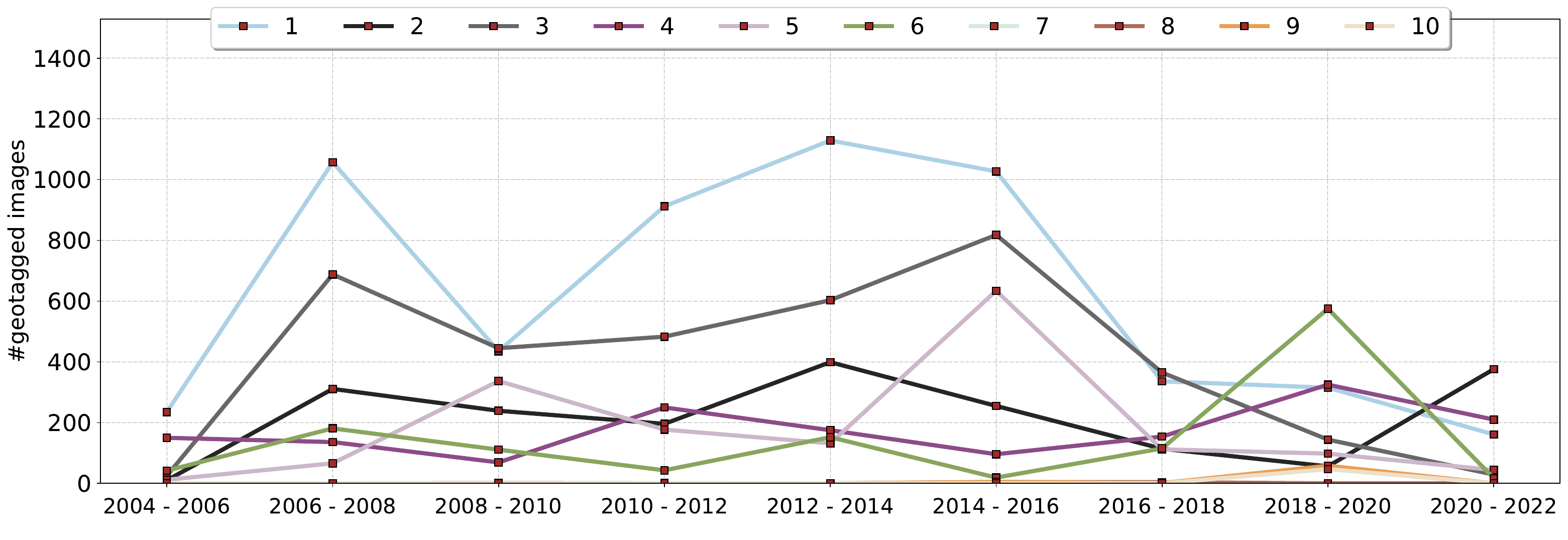}
\caption{Excluding license ID: 0, quantity of geotagged images over time with various Flickr license types, by Flickr numerical license IDs, for Morocco via query-by-name. By excluding the ``All Rights Reserved'' (ID: 0) license, we observe that more images were uploaded under the ``Attribution Non Commercial Share Alike License'' (1) and ``Attribution Non Commercial No-Derivs License'' (3) licenses, and least under ``United States Government Work'' (8), `Public Domain Dedication (CC0)'' (9), and ``Public Domain Mark'' (10) licenses.}
\label{fig:morrocco_people_half_line_lic}
\end{figure*}
\raggedbottom


\clearpage
\subsection{Local vs. non-local representation by self-reported photographer's locations}

\begin{figure*}[thbp]
\centering
\includegraphics[width=1\textwidth]{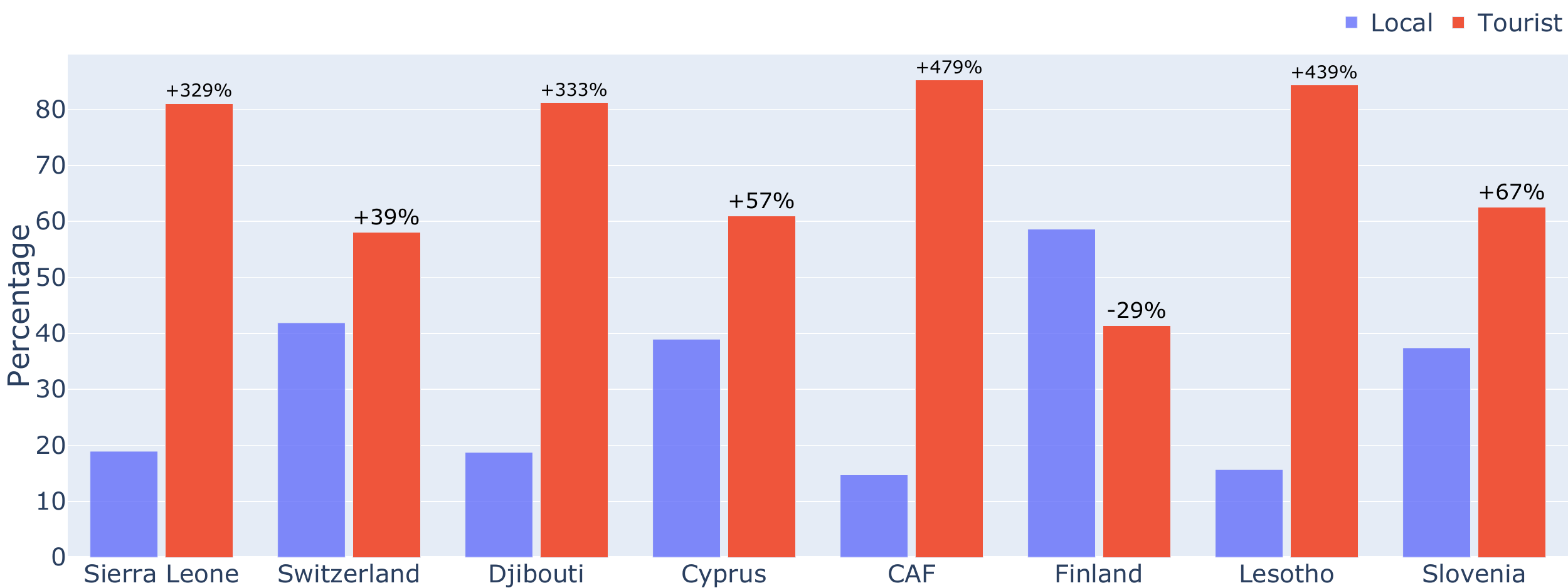}
\caption{Bar chart showing the percentage out of all images from each matched African/European nation pair via query-by-name as taken by locals (blue) and tourists (red), according to photographer's locations. The percent change from local to tourist percentage value is additionally indicated for each nation. In general, Images were predominantly taken by foreigners with higher percentages for African countries than the higher-GDP European countries, e.g., (+$329\%$) for Sierra Leone vs (+$39\%$) for Switzerland. ``CAF'' indicates the country ``Central African Republic''. }
\label{fig:all-data-tours}
\end{figure*}
\raggedbottom

\begin{figure}[!thbp]
\centering
\includegraphics[width=1\textwidth]{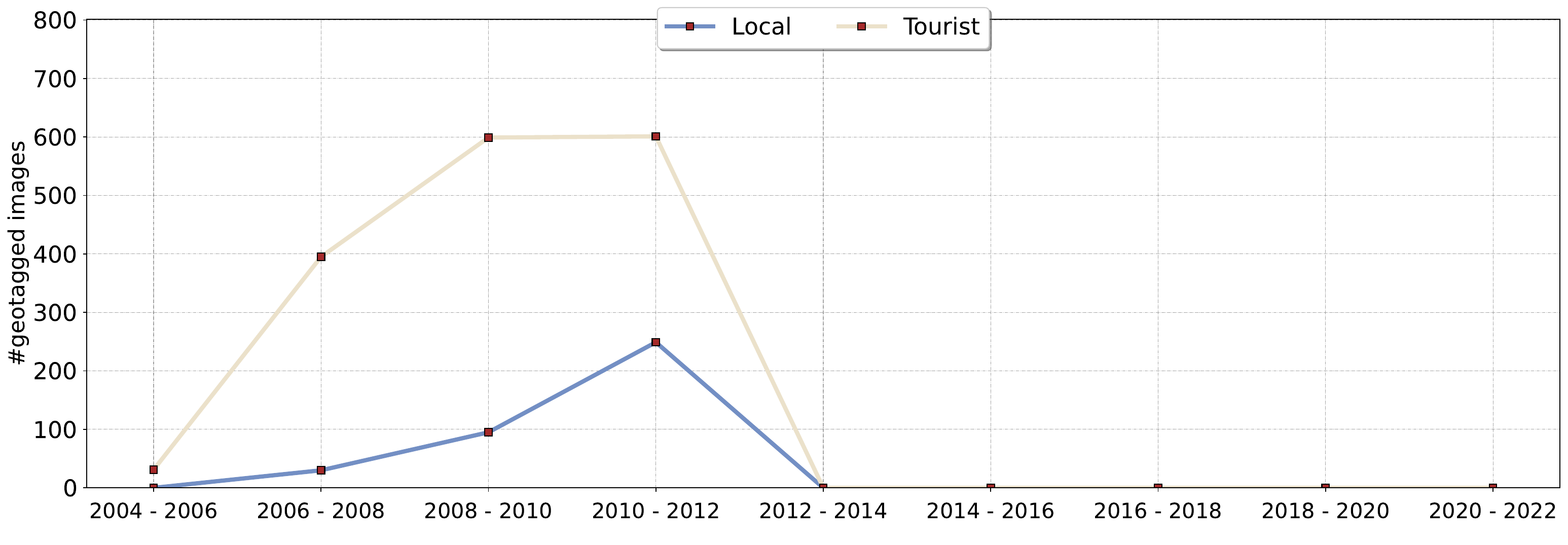}
\caption{Quantity of $2,000$ geotagged images over time for Djibouti via query-by-name according to Flickr user origin: ``Local'' (blue) or non-local/``Tourist'' (beige). In general, far more geotagged images sourced from African nations are uploaded by non-local users.}
\label{fig:survey-Djibouti-Temporal}
\end{figure}
\raggedbottom


\begin{figure*}[thbp]
\centering
\includegraphics[width=1\textwidth]{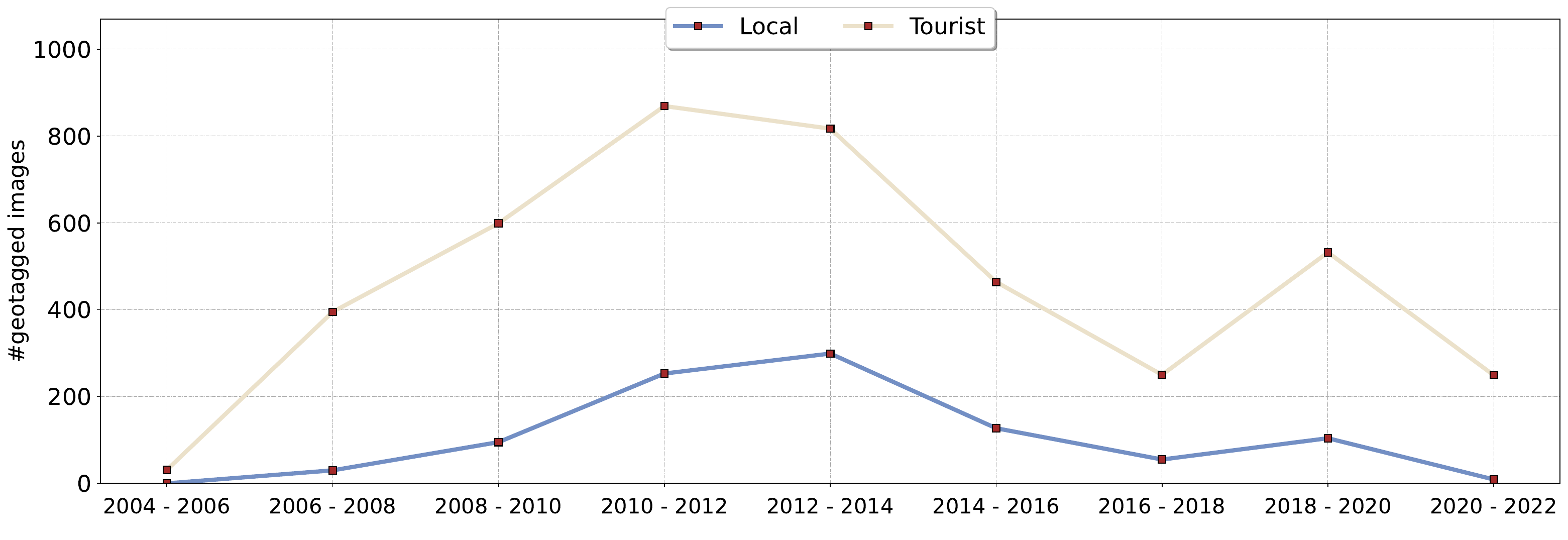}
\caption{Quantity of all geotagged images over time for Djibouti via query-by-name according to Flickr user origin: ``Local'' (blue) or non-local/``Tourist'' (beige). In general, far more geotagged images sourced from African nations are uploaded by non-local users.}
\label{fig:alldata-Djibouti-Temporal}
\end{figure*}
\raggedbottom

\clearpage
\subsection{Selected samples of geotagged images}


\begin{figure*}[!thbp]
\centering
\includegraphics[width=0.79\textwidth]{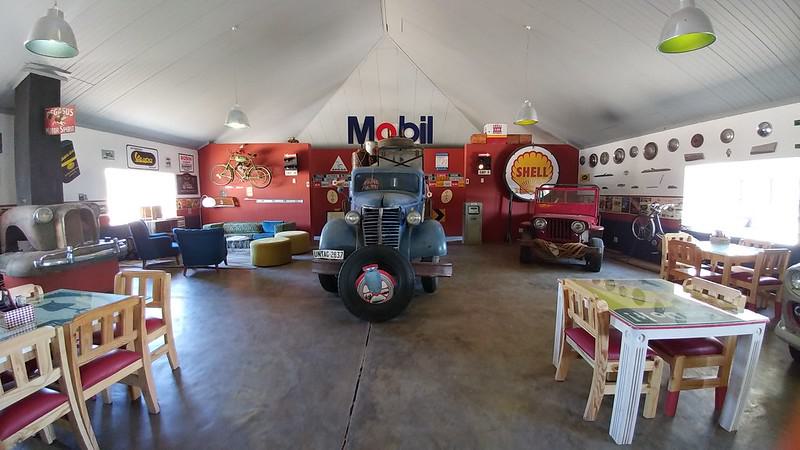} \label{fig:v7_img}
\caption{Indoor image. Image credit: Keetmanshoop, via Flickr.com (CC PDM-BY 1.0).}
\label{fig:v7}
\end{figure*}
\raggedbottom

\begin{figure*}[!thbp]
\centering
\includegraphics[width=0.79\textwidth]{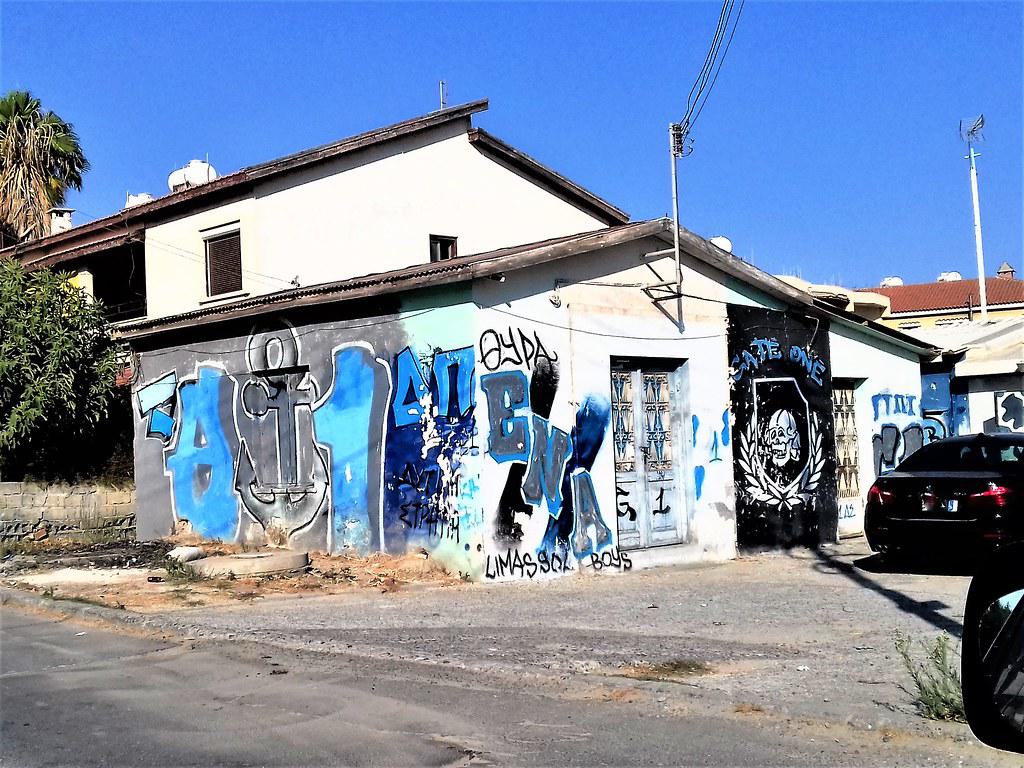} \label{fig:v5_img}
\caption{Outdoor image. Image credit: Jeremiah West, via Flickr.com (CC PDM-BY 1.0).}
\label{fig:v5}
\end{figure*}
\raggedbottom

\begin{figure*}[!thbp]
\centering
\includegraphics[width=0.78\textwidth]{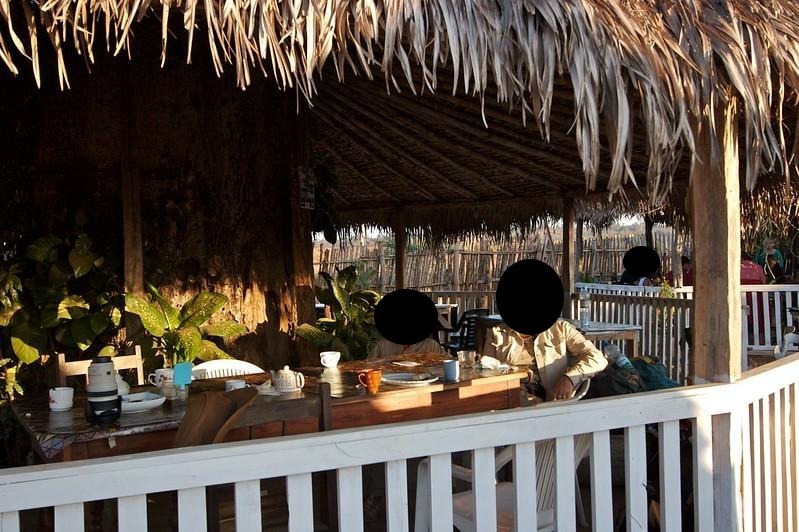} \label{fig:v3_img}
\caption{Public image scene. Image credit: Doganowscy, via Flickr.com (CC NC-BY 2.0).}
\label{fig:v3}
\end{figure*}
\raggedbottom

\begin{figure*}[!thbp]
\centering
\includegraphics[width=0.78\textwidth]{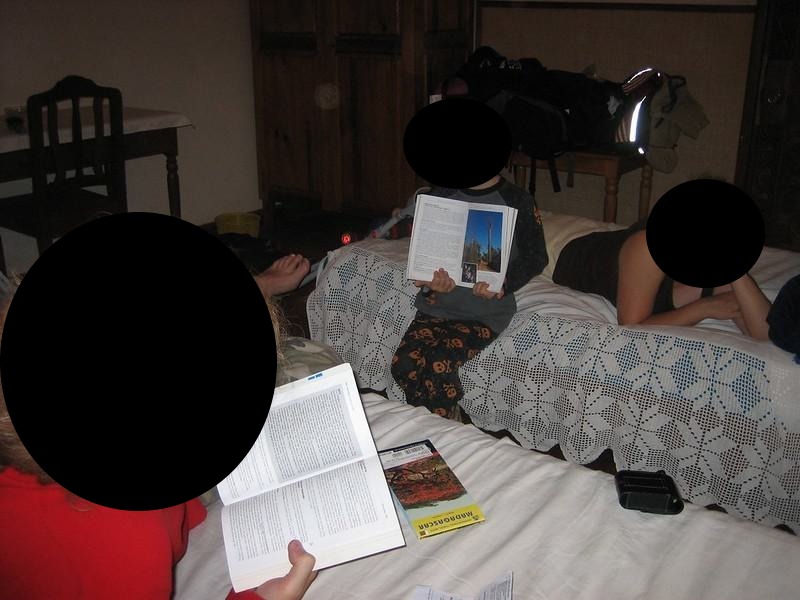} \label{fig:v1_img}
\caption{Private scene image. Image credit: Doganowscy, via Flickr.com (CC NC-BY 2.0).}
\label{fig:v1}
\end{figure*}
\raggedbottom

\begin{figure*}[!thbp]
\centering
\includegraphics[width=0.8\textwidth]{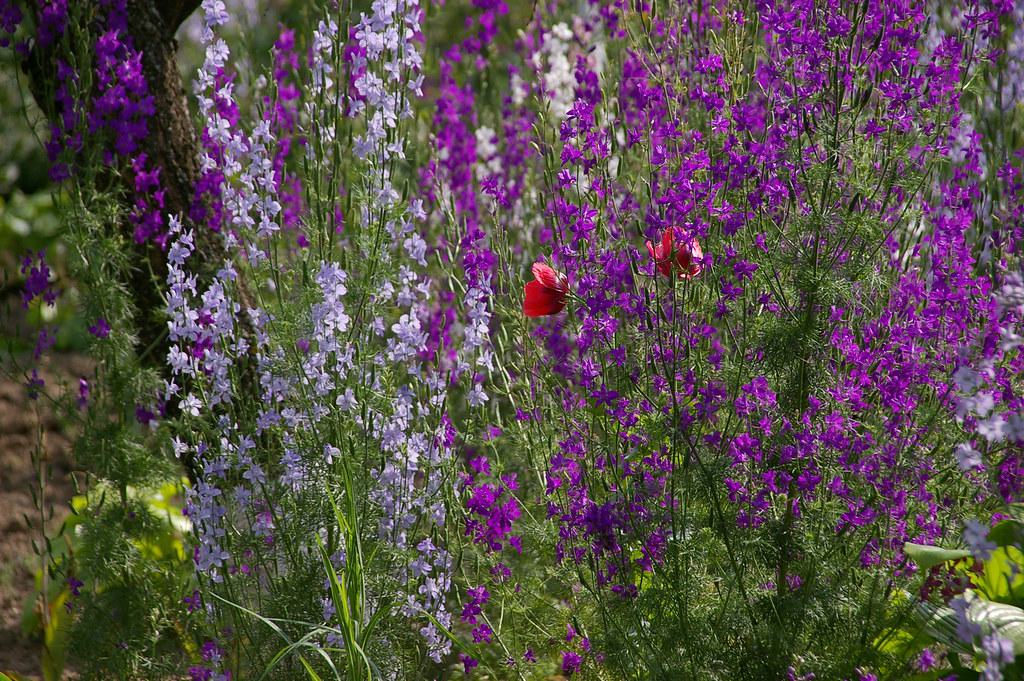} \label{fig:v6_img}
\caption{Nature image scene. Image credit: vlitvinov, via Flickr.com (CC PDM-BY 1.0).}
\label{fig:v6}
\end{figure*}
\raggedbottom

\begin{figure*}[!thbp]
\centering
\includegraphics[width=0.8\textwidth]{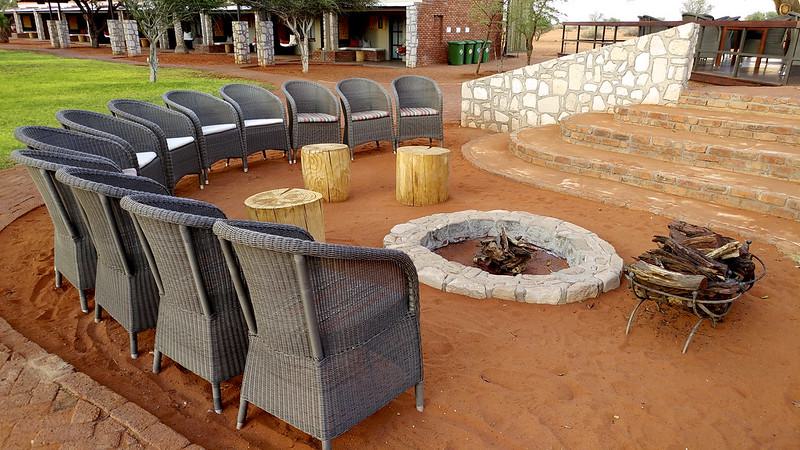} \label{fig:v4_img}
\caption{Man-made (non-``nature'') image scene. Image credit: bayer.com, via Flickr.com (CC PDM-BY 1.0).}
\label{fig:v4}
\end{figure*}
\raggedbottom

\begin{figure*}[!thbp]
\centering
\includegraphics[width=0.8\textwidth]{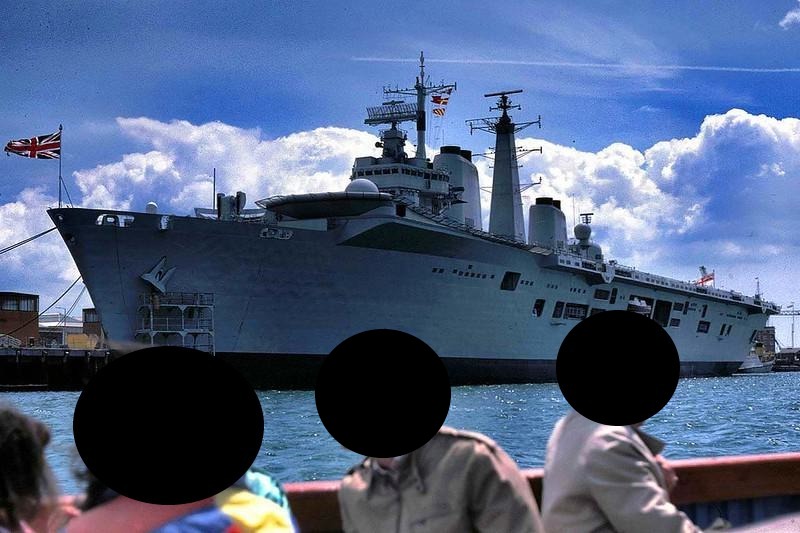} \label{fig:v2_img}
\caption{Image containing people. Image credit: beareye2010, via Flickr.com (CC PDM-BY 1.0).}
\label{fig:v2}
\end{figure*}
\raggedbottom

\begin{figure*}[!thbp]
\centering
\includegraphics[width=0.8\textwidth]{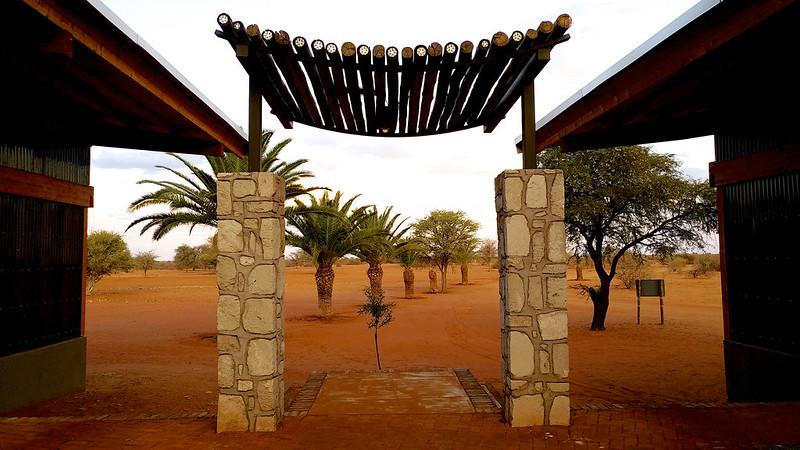} \label{fig:v8_img}
\caption{Image without people. Image credit: bayer.com, via Flickr.com (CC PDM-BY 1.0).}
\label{fig:v8}
\end{figure*}
\raggedbottom


\end{document}